\documentclass[twoside]{article}

\usepackage{microtype}
\usepackage{graphicx}
\usepackage{subfigure}
\usepackage{booktabs} 

\usepackage{hyperref}

\usepackage[algo2e]{algorithm2e} 
\usepackage{algorithm}
\usepackage{algorithmic}

\usepackage[percent]{overpic}
\usepackage{amsthm}
\usepackage{mathtools}
\usepackage{amssymb}

\usepackage{wrapfig}

\usepackage{placeins}

\usepackage[utf8]{inputenc}
\usepackage[english]{babel}
\usepackage{amsthm}
\newtheorem{theorem}{Theorem}

\theoremstyle{definition}
\newtheorem{definition}{Definition}
\theoremstyle{theorem}
\newtheorem*{remark}{Remark}
\theoremstyle{lemma}
\newtheorem{lemma}{Lemma}[theorem]
\theoremstyle{corollary}
\newtheorem{corollary}{Corollary}[theorem]

\newcommand{\tphat}[1]{\expandafter\hat#1} 
\newcommand{\tpbar}[1]{\expandafter\bar#1} 
\newcommand{\tptilde}[1]{\expandafter\tilde#1} 
\newcommand{\bftab}{\fontseries{b}\selectfont}

\usepackage{multirow}
\usepackage{url}

\usepackage{breakurl}
\usepackage{appendix}
\usepackage{makecell}
\usepackage{booktabs}

\usepackage{xcolor}
\usepackage{natbib}

\usepackage{xcolor}

\usepackage{makeidx}
\makeindex
\printindex

\usepackage[accepted]{aistats2020}
\begin{document}

\runningtitle{Uncertainty in Neural Networks: Approximately Bayesian Ensembling}

\runningauthor{Tim Pearce, Felix Leibfried, Alexandra Brintrup, Mohamed Zaki, Andy Neely}

\twocolumn[

\aistatstitle{Uncertainty in Neural Networks:\\ Approximately Bayesian Ensembling}

\aistatsauthor{ Tim Pearce$^{1}$, Felix Leibfried$^2$, Alexandra Brintrup$^1$, Mohamed Zaki$^1$, Andy Neely$^1$}
\aistatsaddress{$^1$University of Cambridge, $^2$PROWLER.io}


]

\begin{abstract}
Understanding the uncertainty of a neural network's (NN) predictions is essential for many purposes. The Bayesian framework provides a principled approach to this, however applying it to NNs is challenging due to large numbers of parameters and data. Ensembling NNs provides an easily implementable, scalable method for uncertainty quantification, however, it has been criticised for not being Bayesian. This work proposes one modification to the usual process that we argue does result in approximate Bayesian inference; regularising parameters about values drawn from a distribution which can be set equal to the prior. A theoretical analysis of the procedure in a simplified setting suggests the recovered posterior is centred correctly but tends to have underestimated marginal variance, and overestimated correlation. However, two conditions can lead to exact recovery. We argue that these conditions are partially present in NNs.  Empirical evaluations demonstrate it has an advantage over standard ensembling, and is competitive with variational methods.




\footnotesize{Interactive demo: \href{https://teapearce.github.io/portfolio/github_io_1_ens/}{\textcolor{blue}{teapearce.github.io}}.}
\end{abstract}

\section{Introduction}
\label{sec_intro}
Neural networks (NNs) are the current dominant force within machine learning, however, quantifying the uncertainty of their predictions is a challenge. This is important for many real-world applications \citep{Bishop1994} as well as in auxiliary ways; to drive exploration in reinforcement learning (RL), for active learning, and to guard against adversarial examples \citep{Smith2018, Sunderhauf2018}

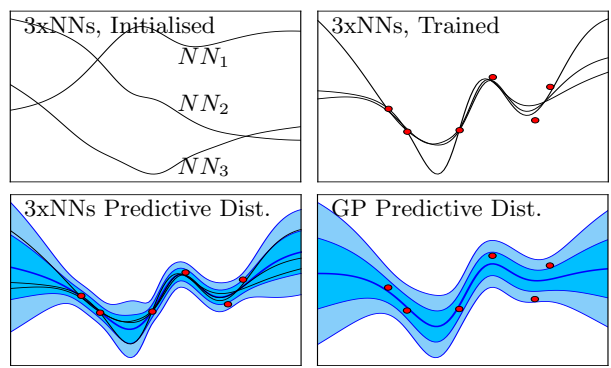
\begin{figure}[b!]
\vskip -0.18in
\begin{center}
    \begin{minipage}{0.243\textwidth}
        \centering
        \begin{overpic}[width=1.\textwidth, height=0.6\textwidth]{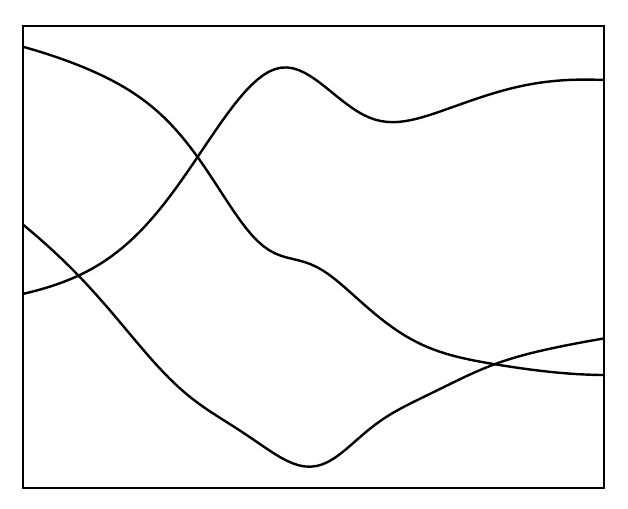}
        \put (8,50) {\small {3xNNs, Initialised}}
        \put (57,40) {\small $NN_1$}
         \put (57,25) {\small $NN_2$}
         \put (57,5) {\small $NN_3$}
        \end{overpic}
    \end{minipage}\hfill
    \hskip -0.12in
    \begin{minipage}{0.243\textwidth}
        \centering
        \begin{overpic}[width=1.\textwidth, height=0.6\textwidth]{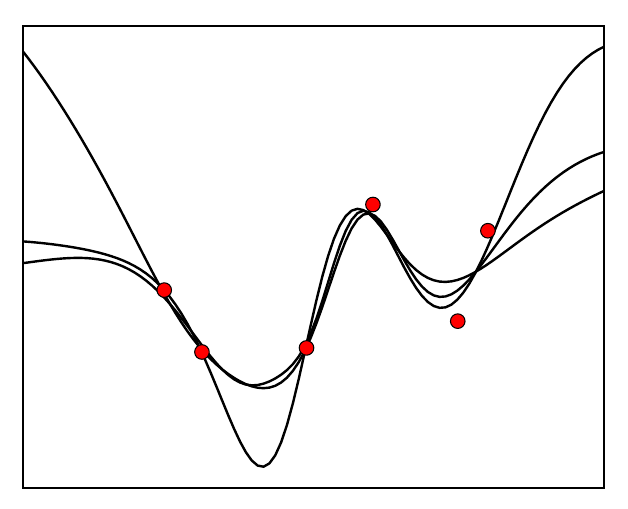}
        \put (8,50) {\small {3xNNs, Trained}}
        \end{overpic}
    \end{minipage}
    \vskip -0.02in
    \begin{minipage}{0.243\textwidth}
        \centering
        \begin{overpic}[width=1.\textwidth, height=0.6\textwidth]{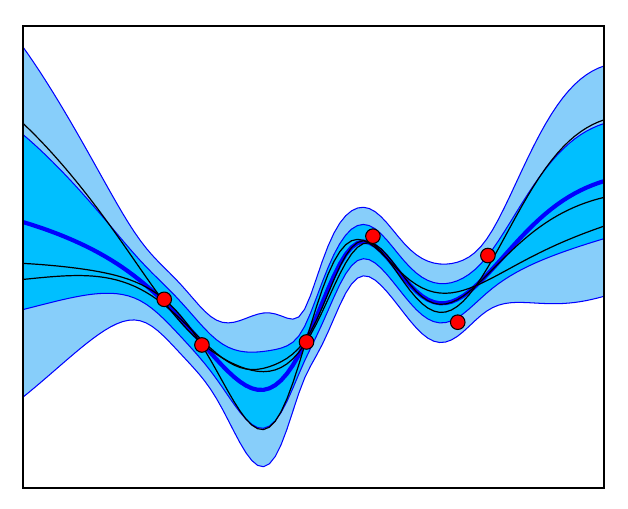}  
        \put (8,50) {\small {3xNNs Predictive Dist.}}
        \end{overpic}
    \end{minipage}
    \hskip -0.08in
    \begin{minipage}{0.243\textwidth}
        \centering
        \begin{overpic}[width=1.\textwidth, height=0.6\textwidth]{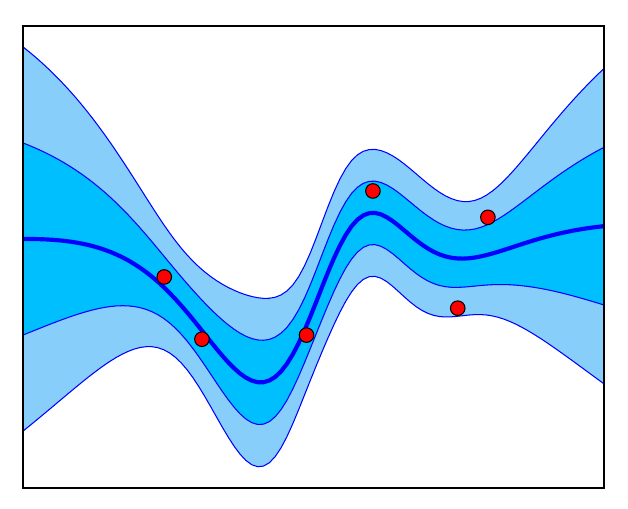}
        \put (8,50) {\small {GP Predictive Dist.}}
        \end{overpic}
    \end{minipage}
\vskip -0.12in
\caption{An ensemble of NNs, starting from different initialisations and trained with the proposed modification, produce a predictive distribution approximating that of a GP. This improves with number of NNs.}
\label{fig_anch_action}
\end{center}
\vskip -0.2in
\end{figure}

A principled approach to modelling uncertainty is provided by the Bayesian framework. Bayesian Neural Networks (BNNs) model the parameters of a NN as probability distributions computed via Bayes rule \citep{MacKay1992}. Whilst appealing, the large number of parameters and data points used with modern NNs renders many Bayesian inference techniques that work well in small-scale settings infeasible, e.g. MCMC methods.





Ensembling provides an alternative way to estimate uncertainty: it aggregates the estimates of multiple individual NNs, trained from different initialisations and sometimes on noisy versions of the training data. The variance of the ensemble's predictions may be interpreted as its uncertainty. The intuition is simple: predictions converge to similar results where data has been observed, and will be diverse elsewhere. The chief attraction is that the method scales well to large parameter and data settings, with each individual NN implemented in precisely the usual way.

Whilst ensembling has proven empirically successful \citep{Tibshirani1996, Lakshminarayanan2016, Osband2016a}, the absence of connection to Bayesian methodology has drawn critics and inhibited uptake, e.g. Gal (\citeyear{Gal2016}) [p. 27].



\subsection{Contribution}

This paper proposes, analyses and tests one modification to the usual NN ensembling process, with the purpose of examining how closely the resulting procedure aligns with Bayesian inference. {The modification regularises parameters about values drawn from an anchor distribution, which can be set to be equal to the prior distribution}. We name this procedure \textit{anchored ensembling} - see figure \ref{fig_anch_action} for an illustration. This falls into a family of little known Bayesian inference methods, \textit{randomised MAP sampling} (RMS) (section \ref{sec_bg_rand_map}).

Our first contributions do not specifically consider NNs; we derive an abstracted version of RMS in parameter space rather than output space. (This abstraction later allows us to propose RMS for classification tasks for the first time.) Under the assumption that the joint parameter likelihood and prior obey a multivariate normal distribution, we show that it is always possible to design an RMS procedure to recover the true posterior.

This design requires knowing the parameter likelihood covariance a priori, which is infeasible except in the simplest models. We propose a workaround that results in an approximation of the posterior. In general this approximation has correct mean but underestimated variance and overestimated correlation. However, two conditions lead to an exact recovery: 1) perfectly correlated parameters, 2) parameters whose marginal likelihood variance is infinite (`extrapolation parameters'). 



We proceed by considering the applicability of RMS to NNs. We discuss the appropriateness of assumptions used in the theoretical analysis, and argue that the two conditions leading to exact recovery of the posterior are partially present in NNs. We postulate this as the reason that predictive posteriors produced by anchored ensembling appear very similar to those by exact Bayesian methods in figures \ref{fig_methods}, \ref{fig_reg_or_not}, \ref{fig_KL_converge_plain} \& \ref{fig_convergence_toy1d}. 

The performance of anchored ensembling is assessed experimentally on regression, image classification, sentiment analysis and RL tasks. It provides an advantage over standard ensembling procedures, and is competitive with variational methods.

\begin{figure*}[t]
\vskip -0.1in
\begin{center}
    \hskip -0.1in
    \begin{minipage}{0.247\textwidth}
        \centering
        \vskip -0.01in 
        \includegraphics[width=0.99\textwidth, height=0.9\textwidth]{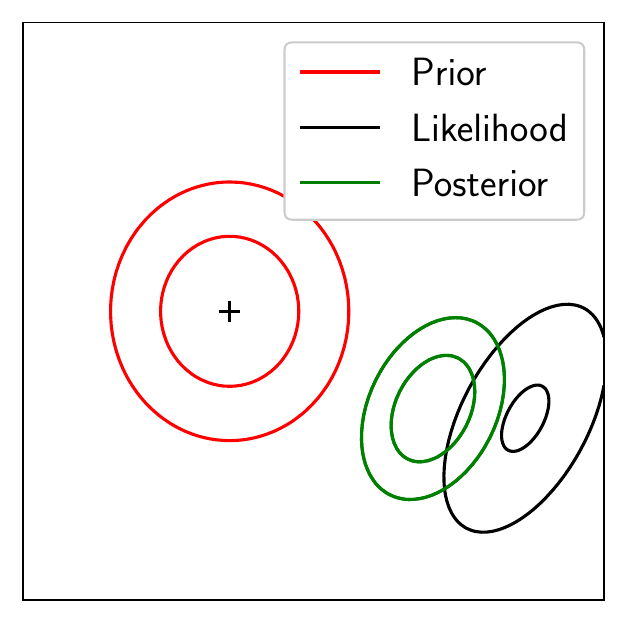}
        \put (-73,50) {\small $\pmb{\mu}_{prior}$}
        \put (-125,50) {\small ${\theta}_{2}$}
        \put (-60,-5) {\small ${\theta}_{1}$}
        \put (-73,50) {\small $\pmb{\mu}_{prior}$}
        \put (-115,-10) {\small Ground truth:}
        \put (-115,-20) {\small Analytical Bayesian}
        \put (-115,-30) {\small  Inference}
    \end{minipage}
    \begin{minipage}{0.247\textwidth}
        \centering
        \includegraphics[width=0.99\textwidth, height=0.9\textwidth]{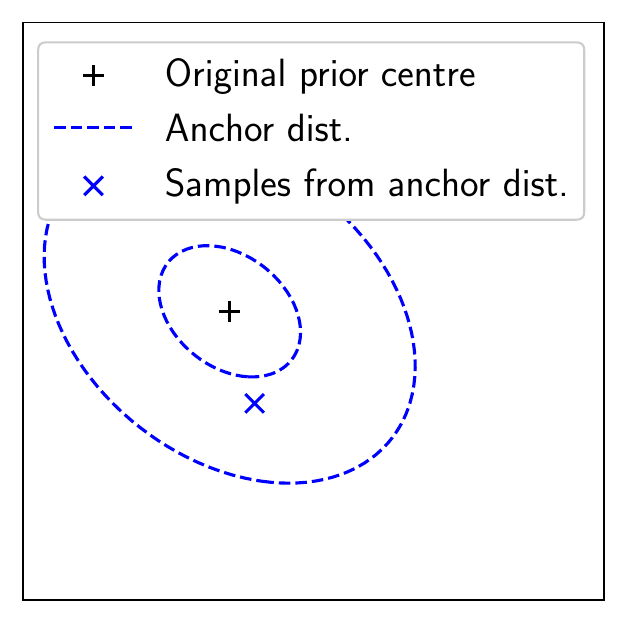}
        \put (-68,32) {\small $\pmb{\theta}_{anc}$}
        \put (-115,-10) {\small \textbf{Step 1:} Set anchor dist.}
        \put (-115,-20) {\small  as $\mathcal{N}(\pmb{\mu}_{anc},\pmb{\Sigma}_{anc})$ from eq.}
        \put (-115,-30) {\small \ref{eq_anch_full_mu} \& \ref{eq_anch_full}. Sample $\pmb{\theta}_{anc}$ once.}
    \end{minipage}
    \begin{minipage}{0.247\textwidth}
        \centering
        \includegraphics[width=0.99\textwidth, height=0.9\textwidth]{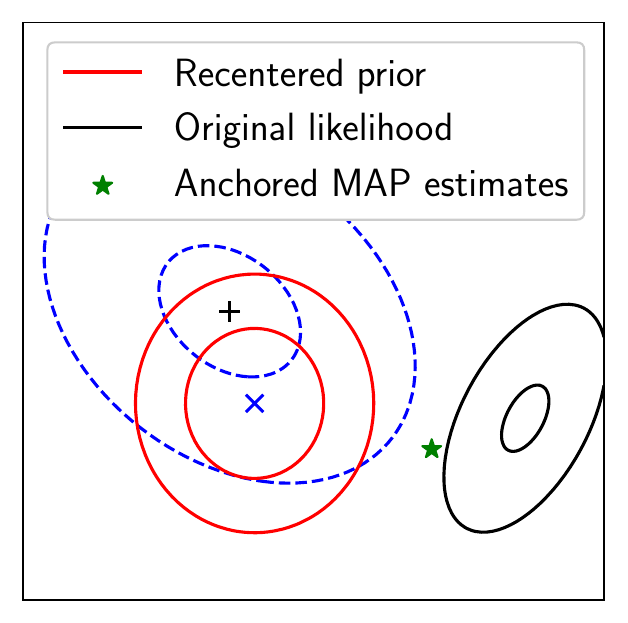}
        \put (-52,21) {\small $\pmb{f}_{MAP}(\pmb{\theta}_{anc})$ }
        \put (-115,-10) {\small \textbf{Step 2:} Return $\pmb{f}_{MAP} (\pmb{\theta}_{anc})$ }
        \put (-115,-20) {\small from eq. \ref{eq_mu_post_text_anch} with $\pmb{\theta}_{anc}$ sampled }
        \put (-115,-30) {\small from step one.}
    \end{minipage}
    \begin{minipage}{0.247\textwidth}
        \centering
        \vskip -0.02in 
        \includegraphics[width=0.99\textwidth, height=0.9\textwidth]{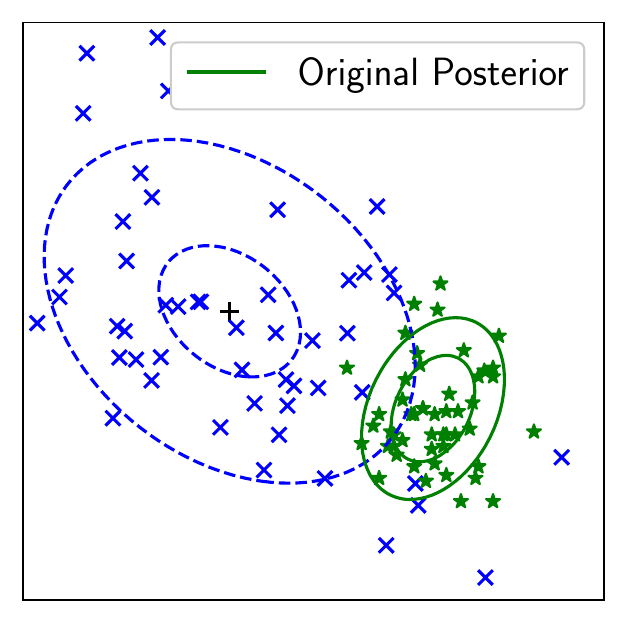}
        \put (-115,-10) {\small \textbf{Step 3:} Repeat steps 1 \& 2.}
        \put (-115,-20) {\small The distribution of $\pmb{f}_{MAP}(\pmb{\theta}_{anc})$  }
        \put (-115,-30) {\small  recovers original posterior.}
    \end{minipage}
\caption{Demonstration of (exact) RMS in a 2D parameter space.}
\label{fig_randomised_anch_inference}
\end{center}
\vskip -0.2in
\end{figure*}
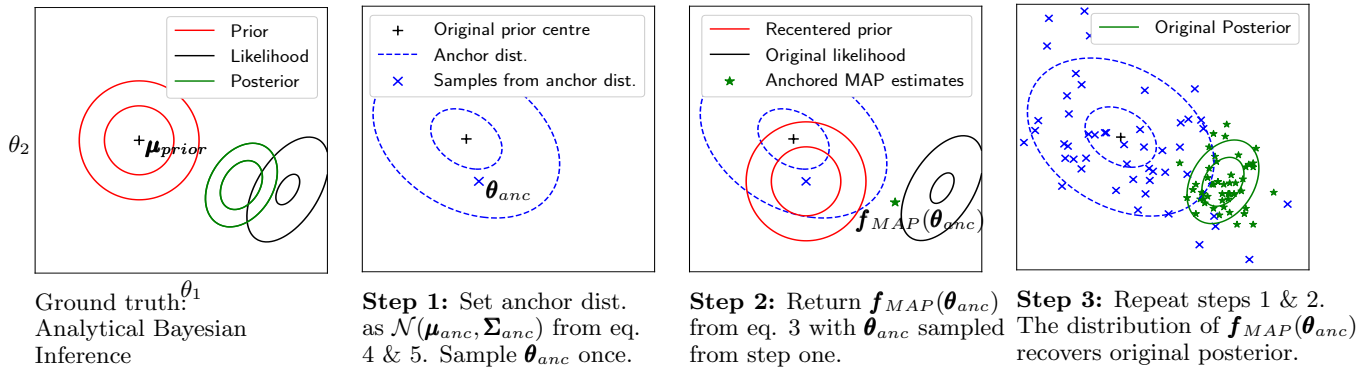

\section{Background}
\subsection{Bayesian Neural Networks}
\label{sec_bg_bnns}

A variety of methods have been developed to perform Bayesian inference in NNs. Variational inference (VI) has received much attention both explicitly modelling parameters with distributions \citep{Graves2011, Hernandez-Lobato2015} and also implicitly through noisy optimisation procedures - MC Dropout \citep{Gal2015}, Vadam \citep{Khan2018}. Correlations between parameters are often ignored - mean-field VI (MFVI).


Other inference methods include: Hamiltonian Monte Carlo (HMC), a MCMC variant which provides `gold standard' inference but at limited scalability \citep{Neal1997}; The Laplace method fits a multivariate normal distribution to the posterior \citep{Ritter2018}. Whilst ensembling is generally seen as a non-Bayesian alternative, \cite{Duvenaud2016} interpreted it, with early stopping, as approximate inference. Aside from \textit{doing} Bayesian inference, recent works have begun exploring prior design in BNNs, e.g. \cite{Pearce2019}.


BNNs of infinite width converge to GPs \citep{Neal1997}. Analytical kernels exist for NNs with certain activation functions, including sigmoidal (Error Function, ERF) \citep{Williams1996}, Rectified Linear Unit (ReLU) \citep{Cho2009}, and leaky ReLU \citep{Tsuchida2018}. Whilst GPs scale superlinearly with data (though see \citep{Wang2019}), they provide a convenient method for doing exact inference on small problems. In this paper we use these GPs as `ground truth' predictive distributions to compare to wide NNs. In section \ref{sec_results}, we benchmark the ReLU GP on UCI datasets.




\subsection{Randomised MAP Sampling}
\label{sec_bg_rand_map}

Recent work in the Bayesian community, and independently in the RL community, has begun to explore a novel approach to Bayesian inference. Roughly speaking, it exploits the fact that adding a regularisation term to a loss function returns a maximum a posteriori (MAP) parameter estimate - a point estimate of the Bayesian posterior. Injecting noise into this loss, either to targets or regularisation term, and sampling repeatedly (i.e. ensembling), produces a \textit{distribution} of MAP solutions mimicking that of the true posterior. This can be an efficient method to sample from high-dimensional posteriors \citep{Gu2007, Chen2012, Bardsley2018}.

Whilst it is possible to specify a noise injection that produces exact inference in linear regression, there is difficulty in transferring this idea to more complex settings, such as NNs or classification. Directly applying the noise distribution from linear regression to NNs has had some empirical success, despite not reproducing the true posterior \citep{Lu2017, Osband2018} (section \ref{sec_comp_work}). A more accurate, though more computationally demanding solution, is to wrap the optimisation step in an MCMC procedure \citep{Bardsley2012, Bardsley2018}. 

These works have been proposed under several names including randomise-then-optimise, randomised prior functions, and ensemble sampling. We refer to this family of procedures \textit{randomised MAP sampling} (RMS).

\section{RMS Theoretical Results}

This section presents several novel results. We first derive a general form of RMS by analysing the procedure in parameter space, using the simplifying assumption that both prior and parameter likelihood are multivariate normal distributions. This is an abstraction compared to previous works. Appendix \ref{app_proof} contains definitions and proofs in full.

If the parameter likelihood covariance is known a priori, we show how RMS can be designed to recover the true posterior. In general, this will not be known, and we propose a practical workaround requiring knowledge only of the prior distribution.

This workaround no longer guarantees exact recovery of the posterior. We derive results specifying in what ways the estimated RMS posterior is in general biased, including underestimated marginal variance, and overestimated correlation coefficient. We discover two special conditions that lead to an exact recovery.


The appropriateness of the normal assumption in non-linear models for general data likelihoods will be discussed in section \ref{sec_anch_ens_method}, when we consider applying this RMS scheme with workaround to NNs. 


\subsection{Parameter-Space Derivation}
\label{sec_rand_anch_samp}

Consider multivariate normal prior and parameter likelihood distributions, $P(\pmb{\theta}) = \mathcal{N}(\pmb{\mu}_{prior},\pmb{\Sigma}_{prior})$, $P_{\pmb{\theta}}(\mathcal{D} | \pmb{\theta})  \propto \mathcal{N}(\pmb{\mu}_{like},\pmb{\Sigma}_{like})$. We make a distinction between two forms of likelihood: \textit{data likelihood}, which is defined on the domain of the target variable, and \textit{parameter likelihood}, which is specified in parameter space. (See definition \ref{def_likelihoods}.)

From Bayes rule the posterior, also normal, is, 
\begin{equation}
\mathcal{N}(\pmb{\mu}_{post},\pmb{\Sigma}_{post}) \propto \mathcal{N}(\pmb{\mu}_{prior},\pmb{\Sigma}_{prior})  \mathcal{N}(\pmb{\mu}_{like},\pmb{\Sigma}_{like})
\end{equation}
The MAP solution is simply $\pmb{\theta}_\text{MAP} = \pmb{\mu}_{post}$, 
\begin{equation}
\label{eq_mu_post_text}
\pmb{\theta}_\text{MAP} = \pmb{\Sigma}_{post}\pmb{\Sigma}^{-1}_{like} \pmb{\mu}_{like} + \pmb{\Sigma}_{post}\pmb{\Sigma}^{-1}_{prior} \pmb{\mu}_{prior},
\end{equation}
where $\pmb{\Sigma}_{post} = (\pmb{\Sigma}^{-1}_{like} +\pmb{\Sigma}^{-1}_{prior})^{-1}$.
In RMS we assume availability of a mechanism for returning $\pmb{\theta}_\text{MAP}$, and are interested in injecting noise into eq. \ref{eq_mu_post_text} so that a \textit{distribution} of $\pmb{\theta}_\text{MAP}$ solutions are produced, matching the true posterior distribution.


A practical choice of noise source is the mean of the prior, $\pmb{\mu}_{prior}$, since a modeller has full control over this value. Let us replace $\pmb{\mu}_{prior}$ with some noisy random variable, $\pmb{\theta}_{anc}$. This is the same place as a hyperprior over $\pmb{\mu}_{prior}$, though with a subtly different role. Denote $\pmb{f}_\text{MAP}(\pmb{\theta}_{anc})$ a function that takes as input $\pmb{\theta}_{anc}$ and returns the resulting MAP estimate,
\begin{equation}
\label{eq_mu_post_text_anch}
\pmb{f}_\text{MAP}(\pmb{\theta}_{anc}) = \pmb{\Sigma}_{post}\pmb{\Sigma}^{-1}_{like} \pmb{\mu}_{like} + \pmb{\Sigma}_{post}\pmb{\Sigma}^{-1}_{prior} \pmb{\theta}_{anc}.
\end{equation}
Accuracy of this procedure hinges on selection of an appropriate distribution for $\pmb{\theta}_{anc}$, which we term the \textit{anchor distribution}. The distribution that will produce the true posterior can be found by setting  $\mathbb{E}[\pmb{f}_{MAP}(\pmb{\theta}_{anc}) ]= \pmb{\mu}_{post}$ and $\mathbb{V}ar[\pmb{f}_{MAP}(\pmb{\theta}_{anc}) ]= \pmb{\Sigma}_{post}$. 

\textbf{Theorem \ref{theorem_anchored_cov}.} \textit{In order that, $P(\pmb{f}_{\text{MAP}}(\pmb{\theta}_{anc})) = P(\pmb{\theta}|\mathcal{D})$, the required distribution of $\pmb{\theta}_{anc}$ is also multivariate normal, $P(\pmb{\theta}_{anc}) =  \mathcal{N}(\pmb{\mu}_{anc},\pmb{\Sigma}_{anc})$, where, }
\vspace{-0.05in}
\begin{align}
\pmb{\mu}_{anc} &= \pmb{\mu}_{prior} \label{eq_anch_full_mu} \\
\pmb{\Sigma}_{anc} &= \pmb{\Sigma}_{prior} +  \pmb{\Sigma}_{prior} \pmb{\Sigma}_{like}^{-1}  \pmb{\Sigma}_{prior}. \label{eq_anch_full}
\end{align}
Figure \ref{fig_randomised_anch_inference} provides a demonstration of the RMS algorithm in 2D parameter space.

\subsection{Comparison to Prior Work}
\label{sec_comp_work}

Previous work on RMS \citep{Lu2017, Osband2017} was motivated via linear regression. Noting that the MAP solution is given by,
\begin{equation}
\label{eq_mu_post_linear}
\pmb{\theta}_\text{MAP} = (\frac{1}{\sigma^2_\epsilon} \mathbf{X}^T\mathbf{X} + \pmb{\Sigma}^{-1}_{prior})^{-1} (\frac{1}{\sigma^2_\epsilon} \mathbf{X}^T \mathbf{y} + \pmb{\Sigma}^{-1}_{prior} \pmb{\mu}_{prior} ),
\end{equation}
these works added Gaussian noise to $\pmb{\mu}_{prior}$, \textit{in addition to} adding noise to $\mathbf{y}$, either by additive Gaussian noise or bootstrapping. Eq. \ref{eq_mu_post_linear} is a special case of our own derivation, substituting $ \pmb{\Sigma}^{-1}_{like} =  1 / \sigma^2_\epsilon \mathbf{X}^T\mathbf{X}$ into eq. \ref{eq_mu_post_text}.


\begin{figure}[b!]
\begin{flushright}
        \begin{minipage}{0.115\textwidth}
        \centering
        \begin{overpic}[width=1. \textwidth]{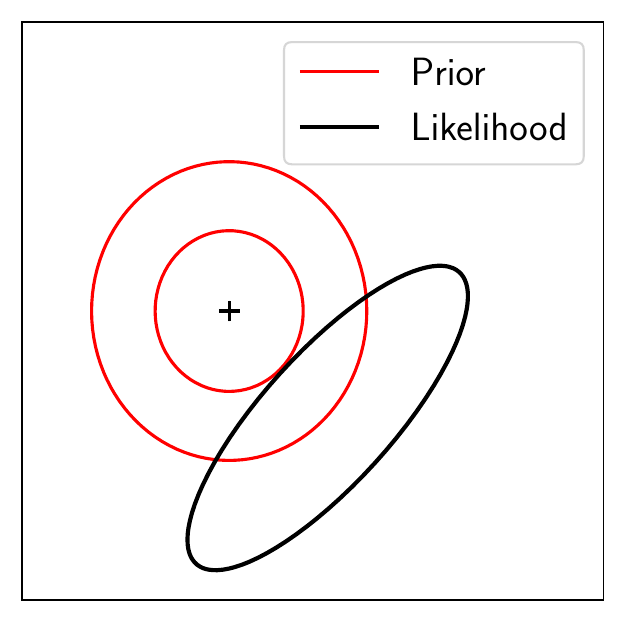}
        \put (5,100) {{\scriptsize Likelihood}}
        \put (-10,5) {\rotatebox{90}{\scriptsize A. General case}}
        \end{overpic}
    \end{minipage}
    \hskip -0.04in
    \begin{minipage}{0.115\textwidth}
        \begin{overpic}[width=1. \textwidth]{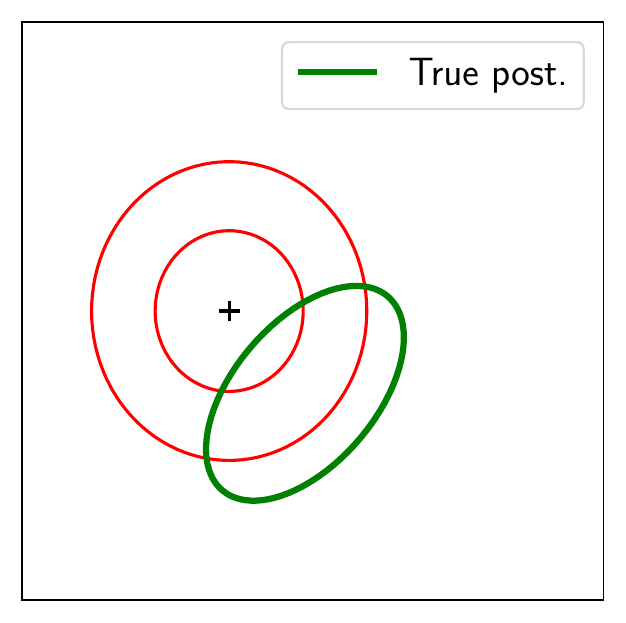}
        \put (5,100) {{\scriptsize True posterior}}
        \end{overpic}
    \end{minipage}
     \hskip -0.04in
    \begin{minipage}{0.115\textwidth}
        \begin{overpic}[width=1. \textwidth]{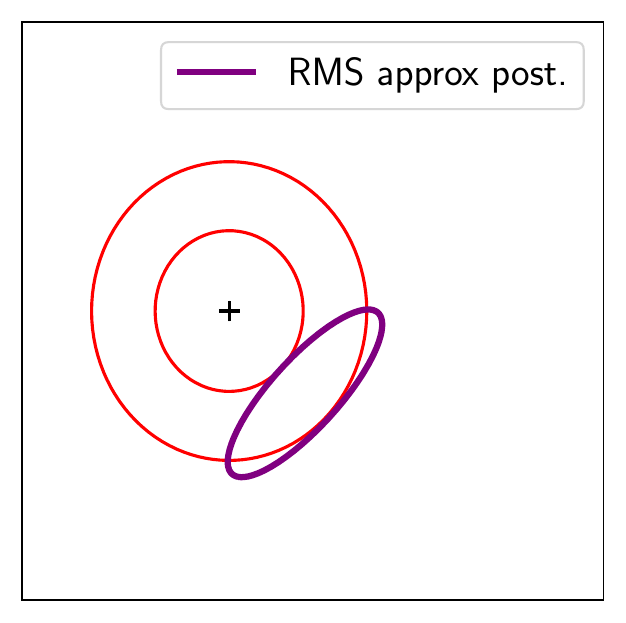}
        \put (5,100) {{\scriptsize RMS approx.}}
        \end{overpic}
    \end{minipage}
    \hskip -0.04in
    \begin{minipage}{0.115\textwidth}
        \begin{overpic}[width=1. \textwidth]{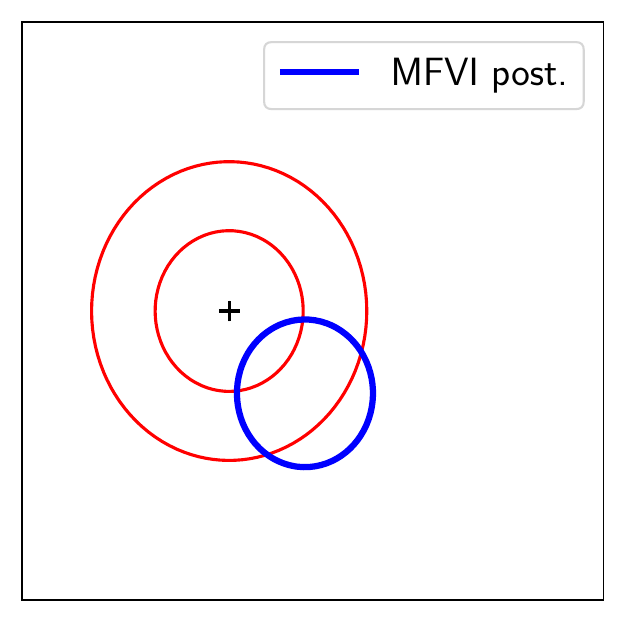}
        \put (5,100) {{\scriptsize MFVI approx.}}
        \end{overpic}
    \end{minipage}
    
        \begin{minipage}{0.115\textwidth}
        \centering
        \begin{overpic}[width=1. \textwidth]{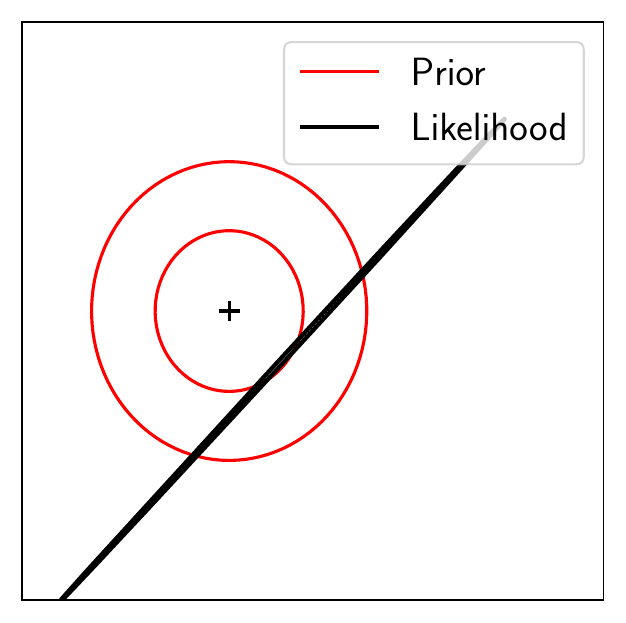}
        \put (-23,5) {\rotatebox{90}{\scriptsize B. Perfect}}
        \put (-10,5) {\rotatebox{90}{\scriptsize correlations}}
        \end{overpic}
    \end{minipage}
    \hskip -0.04in
    \begin{minipage}{0.115\textwidth}
        \begin{overpic}[width=1. \textwidth]{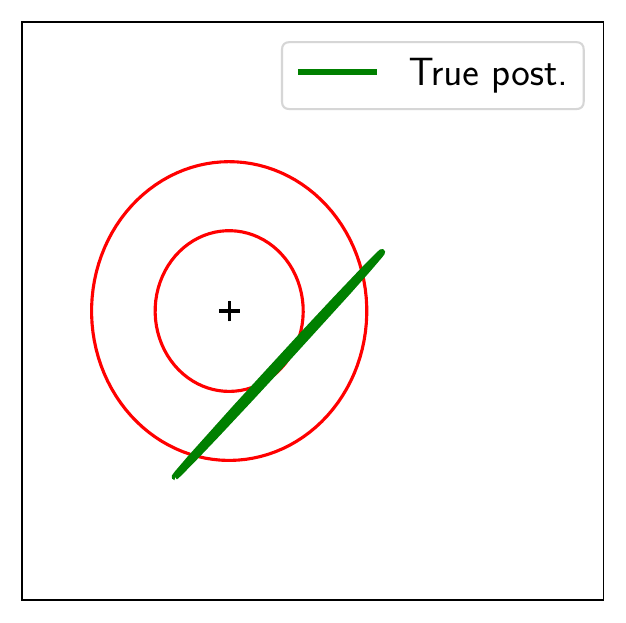}
        \end{overpic}
    \end{minipage}
     \hskip -0.04in
    \begin{minipage}{0.115\textwidth}
        \begin{overpic}[width=1. \textwidth]{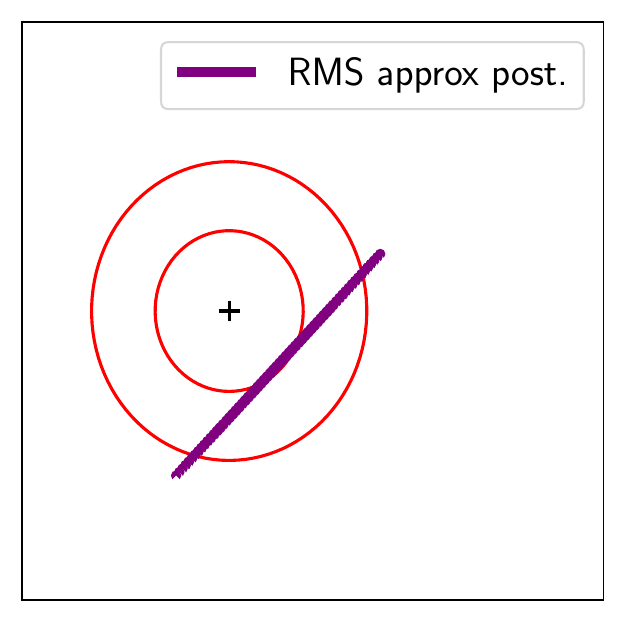}
        \end{overpic}
    \end{minipage}
    \hskip -0.04in
    \begin{minipage}{0.115\textwidth}
        \begin{overpic}[width=1. \textwidth]{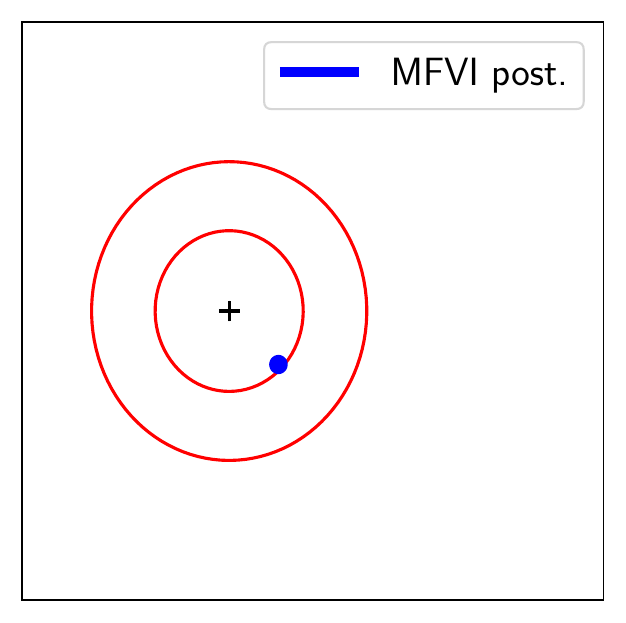}
        \end{overpic}
    \end{minipage}
    
    \begin{minipage}{0.115\textwidth}
        \centering
        \begin{overpic}[width=1. \textwidth]{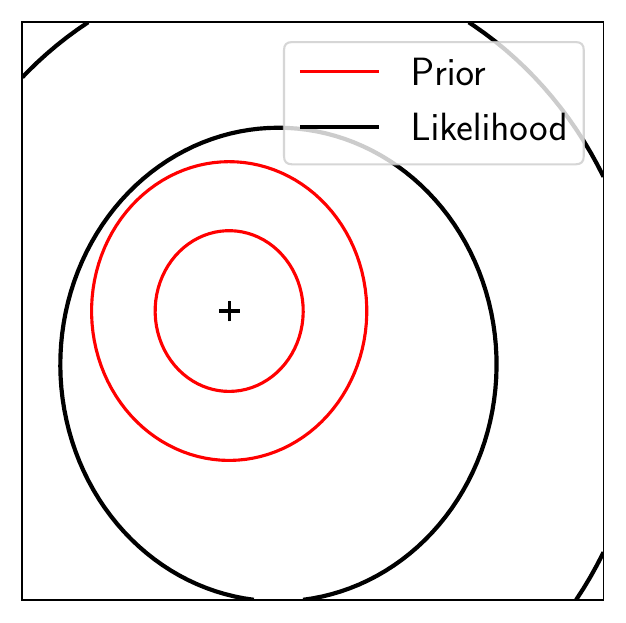}
        \put (-23,5) {\rotatebox{90}{\scriptsize C. Extrapola-}}
        \put (-10,5) {\rotatebox{90}{\scriptsize -tion params}}
        \end{overpic}
    \end{minipage}
    \hskip -0.04in
    \begin{minipage}{0.115\textwidth}
        \begin{overpic}[width=1. \textwidth]{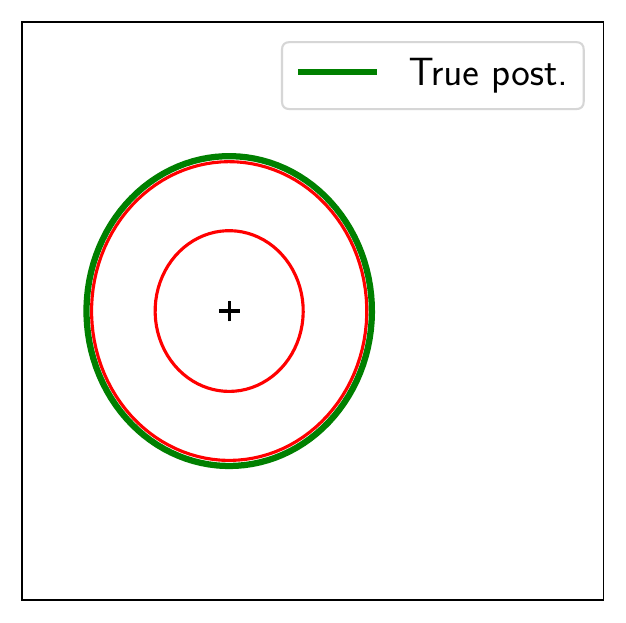}
        \end{overpic}
    \end{minipage}
     \hskip -0.04in
    \begin{minipage}{0.115\textwidth}
        \begin{overpic}[width=1. \textwidth]{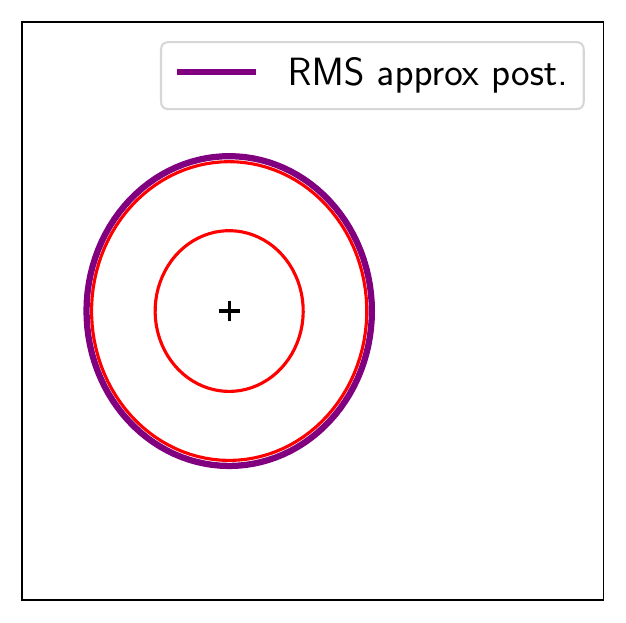}
        \end{overpic}
    \end{minipage}
    \hskip -0.04in
    \begin{minipage}{0.115\textwidth}
        \begin{overpic}[width=1. \textwidth]{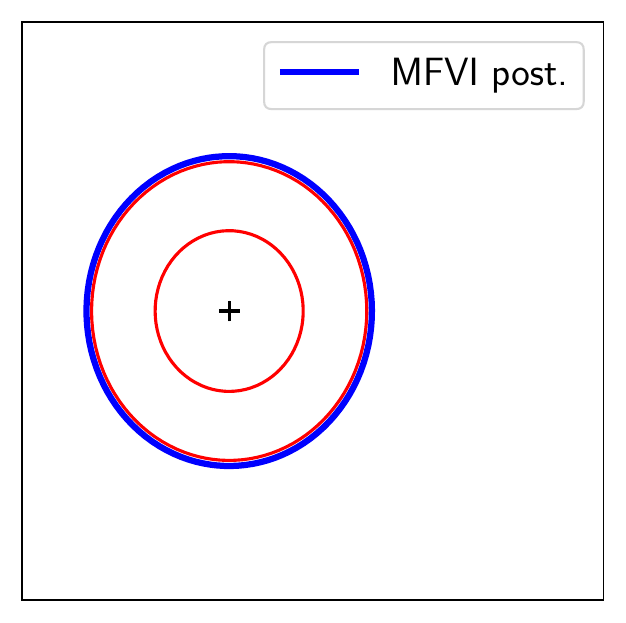}
        \end{overpic}
    \end{minipage}

\caption{Examples of the RMS approximate posterior when $\pmb{\mu}_{anc} \coloneqq \pmb{\mu}_{prior}, \pmb{\Sigma}_{anc} \coloneqq \pmb{\Sigma}_{prior} $. MFVI also shown.}
\vskip -0.2in
\label{fig_RMS_general_special_egs}
\end{flushright}
\end{figure}

\subsection{Practical Workaround: General Case}
The previous section showed how to design an RMS procedure that will precisely recover the true Bayesian posterior. Unfortunately, in eq. \ref{eq_anch_full} one must specify the parameter likelihood covariance in order to set the anchor distribution. For most models, this is infeasible.

A practical workaround is to simply ignore the second term in eq. \ref{eq_anch_full} and set $\pmb{\Sigma}_{anc} \coloneqq \pmb{\Sigma}_{prior}$. Using RMS with this anchor distribution will not generally lead to an exact recovery of the true posterior, however the resulting distribution can be considered an approximation of it. Corollary \ref{cor_anc_eq_prior} derives this RMS approximate posterior in terms of the true posterior.

\textbf{Corollary \ref{cor_anc_eq_prior}.} \textit{Set $\pmb{\mu}_{anc} \coloneqq \pmb{\mu}_{prior}$ and $\pmb{\Sigma}_{anc} \coloneqq \pmb{\Sigma}_{prior}$. The RMS approximate posterior is $P(\pmb{f}_{\text{MAP}}(\pmb{\theta}_{anc})) = \mathcal{N}(\pmb{\mu}_{post}, \pmb{\Sigma}_{post} \pmb{\Sigma}_{prior}^{-1} \pmb{\Sigma}_{post})$.}

\textit{Proof sketch.} This follows similar working to theorem \ref{theorem_anchored_cov}, but instead of enforcing $\mathbb{E}[\pmb{f}_{MAP}(\pmb{\theta}_{anc}) ]= \pmb{\mu}_{post}, \mathbb{V}ar[\pmb{f}_{MAP}(\pmb{\theta}_{anc}) ]= \pmb{\Sigma}_{post}$ and solving for $\pmb{\mu}_{anc}, \pmb{\Sigma}_{anc}$, we now enforce $\pmb{\mu}_{anc} \coloneqq \pmb{\mu}_{prior}$, $\pmb{\Sigma}_{anc} \coloneqq \pmb{\Sigma}_{prior}$ and solve for $\mathbb{E}[\pmb{f}_{MAP}(\pmb{\theta}_{anc})], \mathbb{V}ar[\pmb{f}_{MAP}(\pmb{\theta}_{anc}) ]$.

This corollary shows that the means of the two distributions are aligned, although the covariances are not. Next we state several results quantifying how the RMS approximate posterior covariance differs compared to the true posterior covariance. All results assume multivariate normal prior and parameter likelihood. They can be observed in figure \ref{fig_RMS_general_special_egs} (A).

\textbf{Lemma \ref{lem_anc_eq_prior_underpredicts_var}.} \textit{When $\pmb{\mu}_{anc} \coloneqq \pmb{\mu}_{prior}$, $\pmb{\Sigma}_{anc} \coloneqq \pmb{\Sigma}_{prior}$ the RMS approximate posterior will in general underestimate the marginal variance compared to the true posterior, $\mathbb{V}ar[\pmb{f}_{\text{MAP}}({\theta}_{anc})] < \mathbb{V}ar[ {\theta} | \mathcal{D}]$.}

\textit{Proof sketch.} $\pmb{\Sigma}_{post} \pmb{\Sigma}_{prior}^{-1} \pmb{\Sigma}_{post}$ can be rearranged as $\pmb{\Sigma}_{post} - \pmb{\Sigma}_{post}\pmb{\Sigma}_{like}^{-1}\pmb{\Sigma}_{post}$. The second term will be positive definite, so the diagonal entry is positive, and hence,  $\text{diag}(\pmb{\Sigma}_{post} - \pmb{\Sigma}_{post}\pmb{\Sigma}_{like}^{-1}\pmb{\Sigma}_{post})_i < \text{diag}(\pmb{\Sigma}_{post})_i$.


\textbf{Lemma \ref{lem_anc_eq_prior_isotropic_orientation}.} \textit{Additionally assume the prior is isotropic. When $\pmb{\mu}_{anc} \coloneqq \pmb{\mu}_{prior}$, $\pmb{\Sigma}_{anc} \coloneqq \pmb{\Sigma}_{prior}$ the eigenvectors (or `orientation') of the RMS approximate posterior equal those of the true posterior.}

\textit{Proof sketch.} $\pmb{\Sigma}_{post} \pmb{\Sigma}_{prior}^{-1} \pmb{\Sigma}_{post} = 1/\sigma^2_{prior }\pmb{\Sigma}_{post}^2$. Squaring a matrix only modifies eigenvalues not eigenvectors. As does multiplying by a constant.

\textbf{Theorem \ref{theorem_corr_overestimate}.} \textit{Additionally assume the prior is isotropic. For a two parameter model, when  $\pmb{\mu}_{anc} \coloneqq \pmb{\mu}_{prior}, \pmb{\Sigma}_{anc} \coloneqq \pmb{\Sigma}_{prior}$, the RMS approximate posterior will in general overestimate the magnitude of the true posterior parameter correlation coefficient, $\lvert \rho \rvert$. If $\lvert \rho \rvert = 1$, then it will recover it precisely.}

\textit{Proof sketch.} We compute the individual entries resulting from the required $2\times2$ matrix multiplications.

We were unable to generalise theorem \ref{theorem_corr_overestimate} beyond a two parameter model, but numerical examples (appendix \ref{app_numerical_egs_of_proofs}) suggest that it holds for higher dimensionality.

\begin{figure*}[!t]
\begin{center}
    \begin{minipage}{0.205\textwidth}
        \centering
        \includegraphics[width=0.95\textwidth, height=0.7\textwidth]{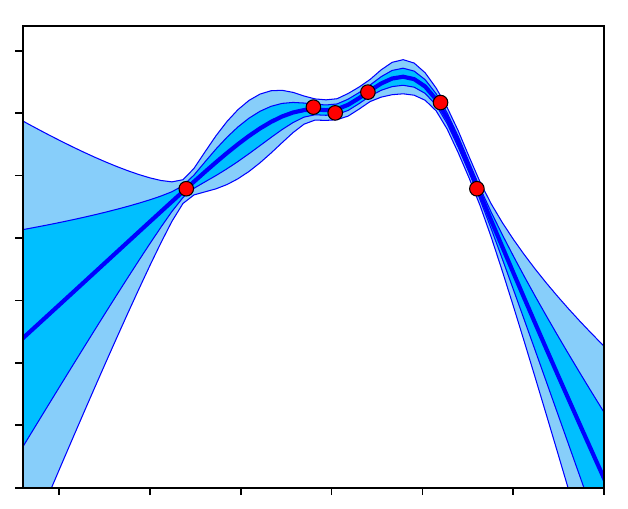}
        \put (-85,72) {\small Ground Truth - GP}
        \put(-105,23){\rotatebox{90}{\small ReLU}}
    \end{minipage}
     \hspace{-0.2in}
    \begin{minipage}{0.205\textwidth}
        \centering
        \includegraphics[width=0.95\textwidth, height=0.7\textwidth]{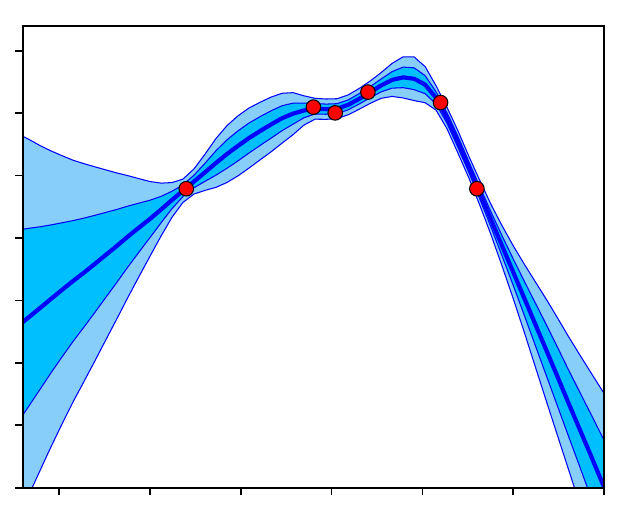}
        \put (-81,72) {\small Hamiltonian MC}
    \end{minipage}
    \hspace{-0.2in}
    \begin{minipage}{0.205\textwidth}
        \centering
        \includegraphics[width=0.95\textwidth, height=0.7\textwidth]{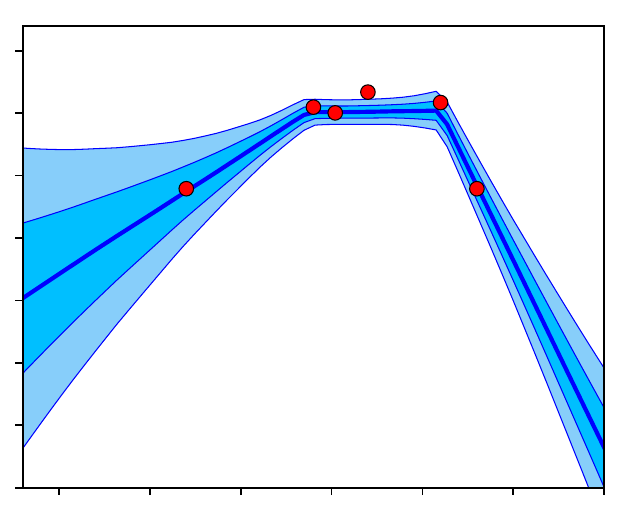}
        \put (-88,72) {\small Variational Inference}
    \end{minipage}
    \hspace{-0.2in}
    \begin{minipage}{0.205\textwidth}
        \centering
        \includegraphics[width=0.95\textwidth, height=0.7\textwidth]{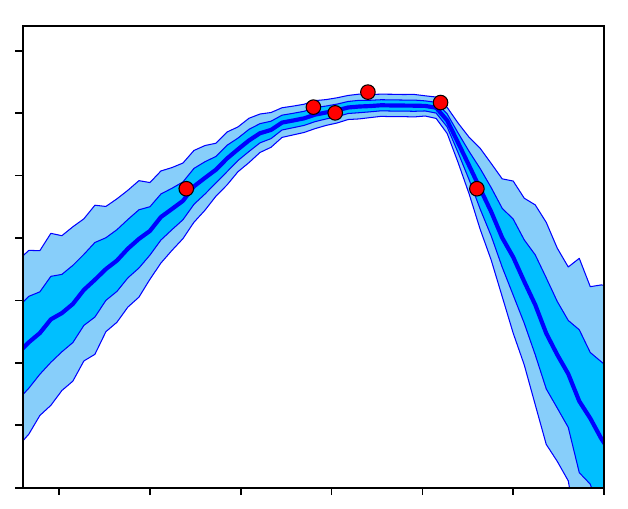}
        \put (-76,72) {\small MC Dropout}
    \end{minipage}
    \hspace{-0.2in}
    \begin{minipage}{0.205\textwidth}
        \centering
        \includegraphics[width=0.95\textwidth, height=0.7\textwidth]{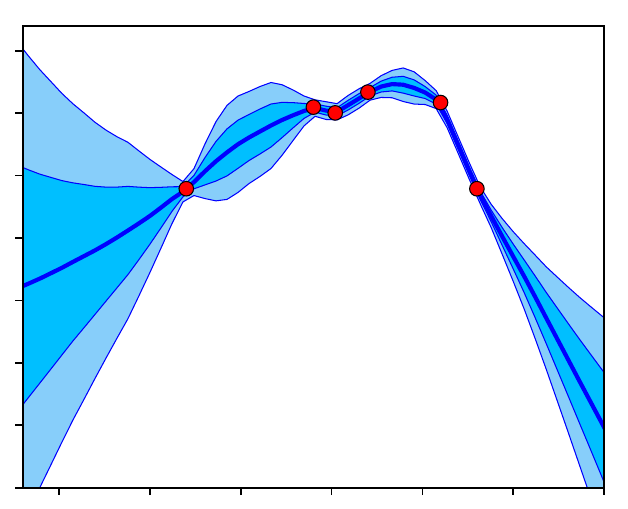}
        \put (-74,72) {\small Our Method}
    \end{minipage}
    
   \vspace{-0.05in}
   
    \begin{minipage}{0.205\textwidth}
        \centering
        \includegraphics[width=0.95\textwidth, height=0.7\textwidth]{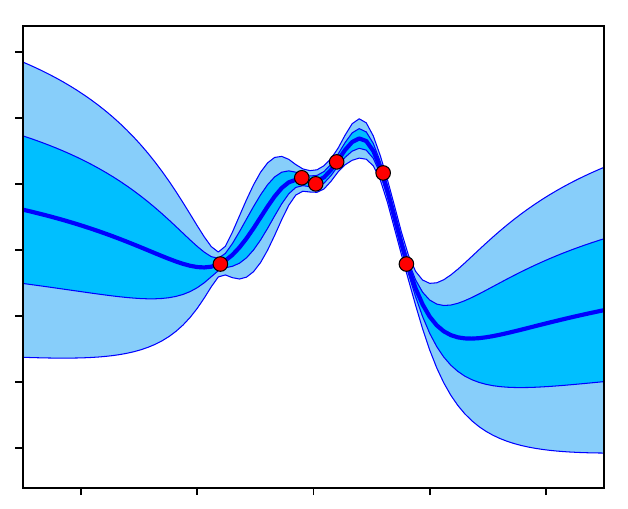}
        \put(-105,5){\rotatebox{90}{\small Sigmoid (ERF)}}
    \end{minipage}
    \hspace{-0.2in}
    \begin{minipage}{0.205\textwidth}
        \centering
        \includegraphics[width=0.95\textwidth, height=0.7\textwidth]{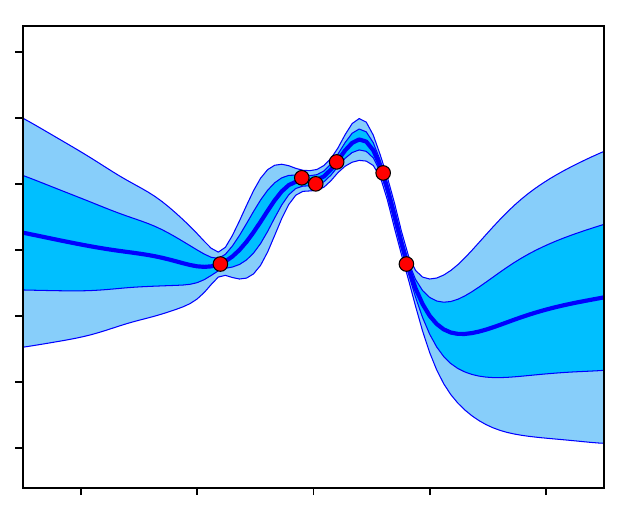}
    \end{minipage}
    \hspace{-0.2in}
    \begin{minipage}{0.205\textwidth}
        \centering
        \includegraphics[width=0.95\textwidth, height=0.7\textwidth]{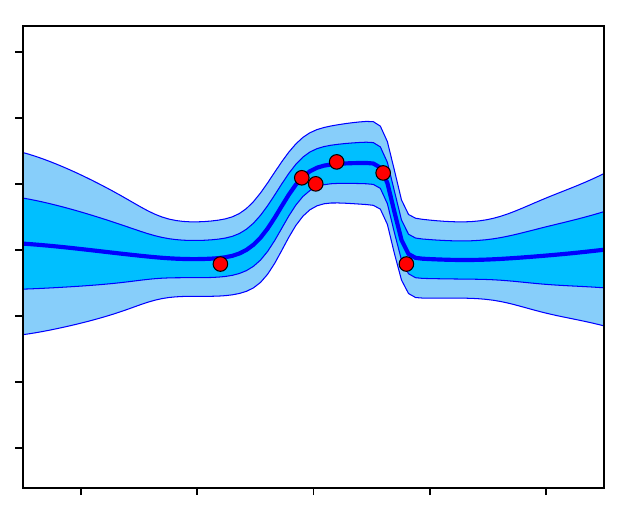}
    \end{minipage}
    \hspace{-0.2in}
    \begin{minipage}{0.205\textwidth}
        \centering
        \includegraphics[width=0.95\textwidth, height=0.7\textwidth]{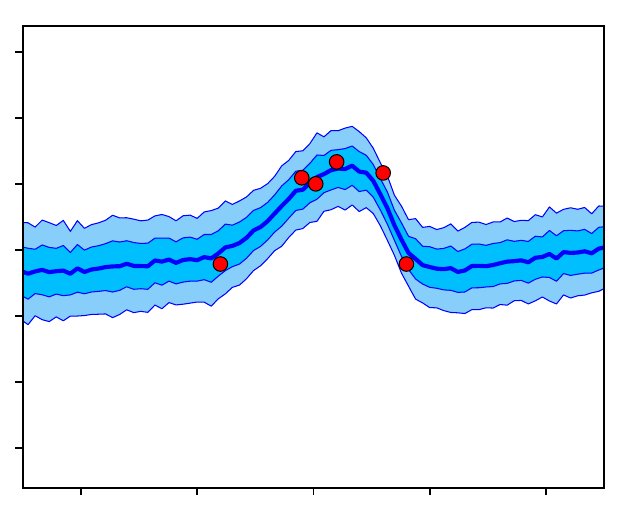}
    \end{minipage}
    \hspace{-0.2in}
    \begin{minipage}{0.205\textwidth}
        \centering
        \includegraphics[width=0.95\textwidth, height=0.7\textwidth]{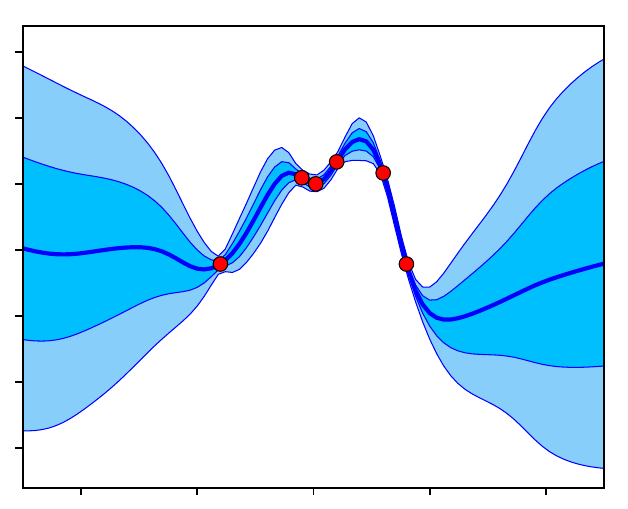}
    \end{minipage}
      
\caption{Predictive distributions produced by various inference methods (columns) with varying activation functions (rows) in single-layer NNs on a toy regression task.}

\label{fig_methods}
\end{center}
\vskip -0.2in
\end{figure*}

\subsection{Practical Workaround: Special Cases}
\label{sec_RMS_special_case}

Having described the covariance bias that in general will be present in the RMS approximate posterior, we now give two special conditions under which there is no bias, and the true posterior is exactly recovered. Illustrations of these cases are shown in figure \ref{fig_RMS_general_special_egs} (B, C).


\textbf{Theorem \ref{theorem_infinite_likelihood}.} \textit{For extrapolation parameters (def. \ref{def_extrapolation parameters} - parameters which do not affect data likelihood but may affect new predictions) of a model, setting $\pmb{\mu}_{anc} \coloneqq \pmb{\mu}_{prior}$, $\pmb{\Sigma}_{anc} \coloneqq \pmb{\Sigma}_{prior}$, means the marginal RMS approximate posterior equals that of the marginal true posterior.}

\textit{Proof sketch.} We show that the required matrix multiplications, $\pmb{\Sigma}_{post} \pmb{\Sigma}_{prior}^{-1} \pmb{\Sigma}_{post}$, do not affect rows corresponding to extrapolation parameters.

\textbf{Theorem \ref{theorem_correlations}.} \textit{Set $\pmb{\mu}_{anc} \coloneqq \pmb{\mu}_{prior}, \pmb{\Sigma}_{anc} \coloneqq \pmb{\Sigma}_{prior}$. The RMS approximate posterior will exactly equal the true posterior, $\pmb{\Sigma}_{post}$, when all eigenvalues of a scaled version of $\pmb{\Sigma}_{post}$ (scaled such that the prior equals the identity matrix) are equal to either $0$ or $1$. This corresponds to posteriors that are a mixture of perfectly correlated and perfectly uncorrelated parameters.}

\textit{Proof sketch.} We are searching for solutions to $\pmb{\Sigma}_{post} = \pmb{\Sigma}_{post} \pmb{\Sigma}_{prior}^{-1} \pmb{\Sigma}_{post}$. Applying a scaling, $\pmb{\Sigma}_{post}' = \pmb{\Sigma}_{prior}^{-1/2} \pmb{\Sigma}_{post} \pmb{\Sigma}_{prior}^{-1/2}$, results in a slightly simpler equation to find a solution to, $\pmb{\Sigma}_{post}' = \pmb{\Sigma}_{post}'^2$.  Results for idempotent matrices tell us that if $\pmb{\Sigma}'_{post}$ is singular with all eigenvalues equal to $0$ or $1$, this will be a solution.

To provide intuition behind theorem \ref{theorem_correlations} consider a two parameter model. If parameters are perfectly correlated, the effect on the data likelihood of an increase in the first can be exactly compensated for by a change in the second. If the region over which this applies is large relative to the prior, the likelihood is a line of negligible width. This leads to a posterior of negligible width spanning the prior. Examples in appendix \ref{app_numerical_egs_of_proofs} show what combinations of parameters this holds for.



This section's proofs show that if these two conditions exist, RMS makes a precise recovery.  In practise, one would expect to see an increasingly accurate RMS approximation as these conditions are approached. 



\section{RMS for Neural Networks}
\label{sec_anch_ens_method}

We now apply RMS with practical workaround to NNs. We will refer to this as `anchored ensembling'.





First, we define the NN loss function to be optimised that corresponds to RMS. We then discuss the validity of the RMS procedure in the context of NNs, given the assumptions made. Finally we consider some matters arising in implementation of the scheme. Appendix, algorithm \ref{alg_RL_ens} details the full procedure.

\subsection{Loss Function}
Consider a NN containing parameters, $\pmb{\theta}$, making predictions, $\mathbf{\hat{y}}$, with $H$ hidden nodes and $N$ data points. If the prior is given by $P(\pmb{\theta}) = \mathcal{N}(\pmb{\mu}_{prior}, \pmb{\Sigma}_{prior})$, maximising the following returns MAP parameter estimates. (See appendix \ref{app_map_solution_interp} for the standard derivation.)
\begin{equation}
\label{eq_MAP_loglike_anc}
\pmb{\theta}_{MAP}  = \text{argmax}_{\pmb{\theta}} \log( P_{\mathcal{D}}( \mathcal{D} | \pmb{\theta} ) ) - 
 \frac{1}{2} 
 \lVert \pmb{\Sigma}_{prior}^{-1/2} \cdot
 (\pmb{\theta} - \pmb{\mu}_{prior}) \rVert^2_2
\end{equation}
When $\pmb{\mu}_{prior} = \mathbf{0}$, this is standard L2 regularisation. In order to apply RMS we instead replace $\pmb{\mu}_{prior}$ with some random variable $ \pmb{\theta}_{anc}$.
To use the practical form of RMS, we will draw $\pmb{\theta}_{anc} \sim \mathcal{N}(\pmb{\mu}_{prior},\pmb{\Sigma}_{prior} )$. 

Conveniently, no parametric form of data likelihood has yet been specified. For a regression task assuming homoskedastic Gaussian noise of variance $\sigma^2_\epsilon$, MAP estimates are found by minimising,
\begin{equation}
\label{eqn_anch_loss_matrix}
Loss_{j} =  
\frac{1}{N} \lvert \lvert \mathbf{y} - \hat{\mathbf{y}}_j \rvert \rvert ^2_2
+ \frac{1}{N} \lvert \lvert \pmb{\Gamma}^{1/2} \cdot (\pmb{\theta}_j - \pmb{\theta}_{anc,j}) \rvert \rvert ^2_2.
\end{equation}
We have defined a diagonal regularisation matrix, $\pmb{\Gamma}$, where the $i^{th}$ diagonal element is the ratio of data noise of the target variable to prior variance for parameter $\theta_i$, $\text{diag}(\pmb{\Gamma})_i = \sigma^2_\epsilon / \sigma^2_{prior_i}$.
Note a subscript has been introduced, $j \in \{1 ... M\}$, with the view of an ensemble of $M$ NNs, each with a distinct draw of $\pmb{\theta}_{anc}$. 

For classification tasks, cross entropy is normally maximised, which assumes a multinomial data likelihood,
\vspace{-0.2in}
\begin{equation}
\label{eqn_anch_loss_matrix_class}
=-\frac{1}{N} \sum_{n=1}^{N}  \sum_{c=1}^{C} {y_{n,c} \log \hat{ y}_{n,c,j}  }
+ \frac{1}{N} \lvert \lvert \pmb{\Gamma}^{1/2} \cdot (\pmb{\theta}_j - \pmb{\theta}_{anc,j}) \rvert \rvert ^2_2,
\end{equation}
where $y_c$ is the label for class $c \in \{1 ... C \}$. Here, $\text{diag}(\pmb{\Gamma})_i = 1 / 2 \sigma^2_{prior_i}$.


\subsection{Validity of RMS in NNs}
Theory derived to motivate and analyse RMS assumed a simplified setting of multivariate normal parameter likelihoods. This section discusses this assumption, then considers the prevalence of special conditions (section \ref{sec_RMS_special_case}) that would lead to a close approximation of the true posterior.

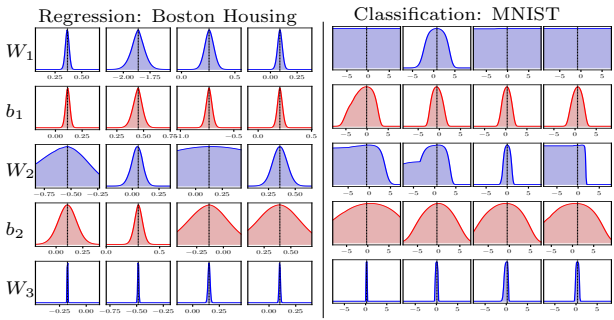
\begin{figure}[b!]
\vskip -0.05in
\begin{center}
    \begin{minipage}{0.23\textwidth}
        \centering
        \begin{overpic}[width=1.\textwidth, height=1.\textwidth]{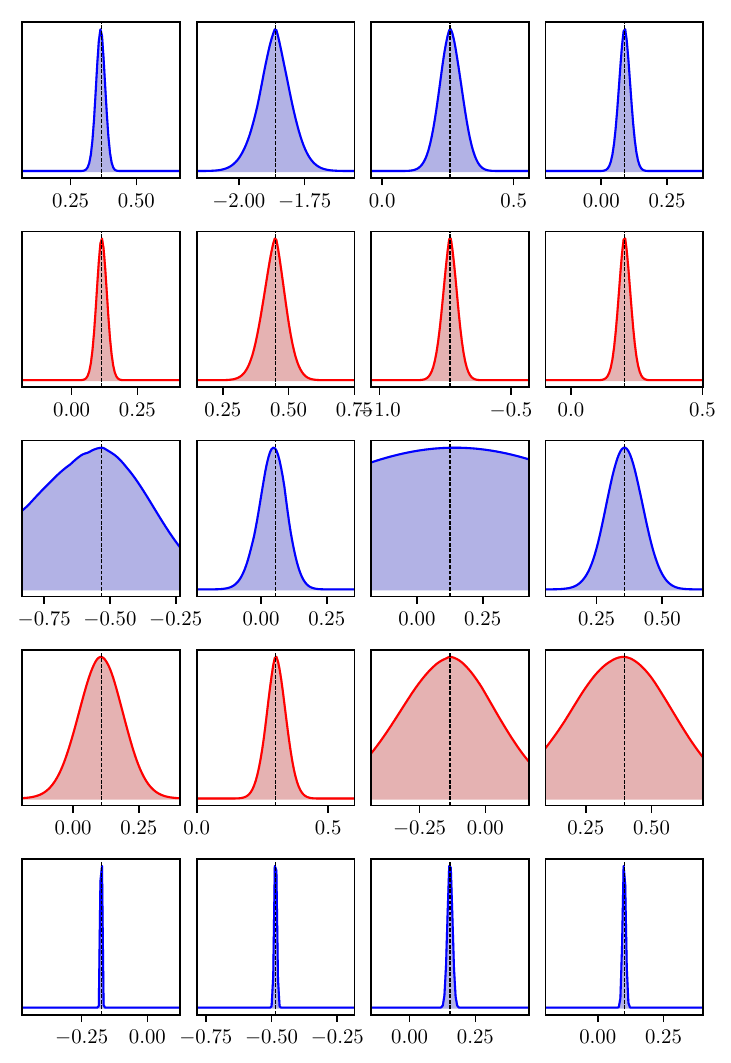}
        \put (4,100) {\scriptsize Regression: Boston Housing}
        \put (-7,88) {\scriptsize $W_1$}
        \put (-7,68) {\scriptsize $b_1$}
        \put (-7,48) {\scriptsize $W_2$}
        \put (-7,28) {\scriptsize $b_2$}
        \put (-7,8) {\scriptsize $W_3$}
        \end{overpic}
    \end{minipage}
    \hskip -0.04in
    \vline
    \begin{minipage}{0.23\textwidth}
        \begin{overpic}[width=1.\textwidth, height=1.\textwidth]{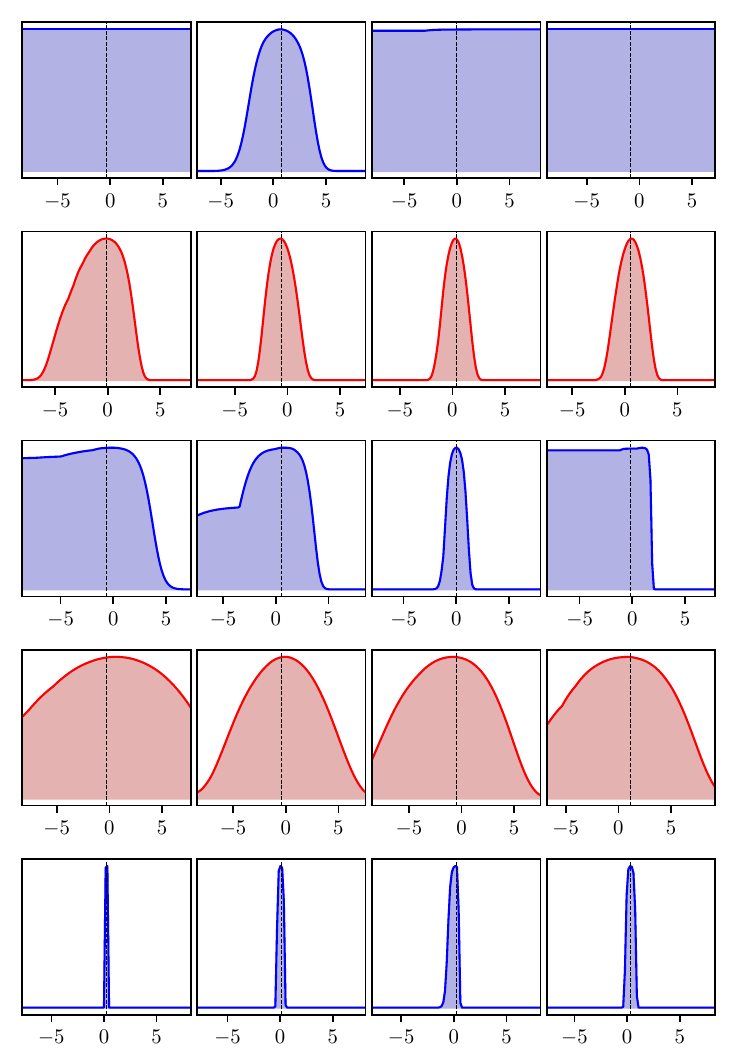}
       \put (10,100) {\scriptsize Classification: MNIST}
        \end{overpic}
    \end{minipage}
    \vskip -0.05in
\caption{Empirical plots of conditional likelihoods for 4 randomly sampled parameters in two-layer NNs.}
\vskip -0.2in
\label{fig_gauss_like}
\end{center}
\end{figure}

\subsubsection{Normal Distribution}

Earlier proofs assumed parameter likelihoods follow a multivariate normal distribution. We provide two justifications for using this assumption in NNs. 

1) Other approximate Bayesian methods incorporate similar assumptions into their methodologies. MFVI commonly fits a factorised normal distribution to the posterior. The Laplace approximation fits a multivariate normal distribution to the mode of a MAP solution. 


2) In figure \ref{fig_gauss_like} we visualise conditional parameter likelihoods for actual NNs trained on regression and classification tasks. After training, a parameter is randomly selected, and all others are frozen. The chosen parameter is varied over a small range and the data likelihood calculated at each point. Hence conditional distributions may be plotted. The plots suggest that thinking of local modes as approximately normally distributed is not unreasonable for the purpose of analysis.

This justifies modelling a single mode of the parameter likelihood as multivariate normal. However, the parameter space of a NN is likely to contain many such modes, with each member of an anchored ensemble ending up at a different one. We believe that many of these modes would be exchangeable, for example arising from parameter symmetries. In this case we believe that MAP solutions would also be exchangeable.


Empirically we did not observe this multimodality being problematic - plots such as figure \ref{fig_convergence_toy1d} show predictive posteriors with low bias compared to the true posterior.

\subsubsection{Presence of Special Cases}
\label{sec_presence_special_case}

Setting the anchor distribution equal to the prior leads to an RMS approximate posterior that, in general, has underestimated variance and overestimated correlation.

\begin{figure}[t]
\begin{center}
    \begin{minipage}{0.24\textwidth}
        \centering
        \begin{overpic}[width=1.\textwidth, height=0.6\textwidth]{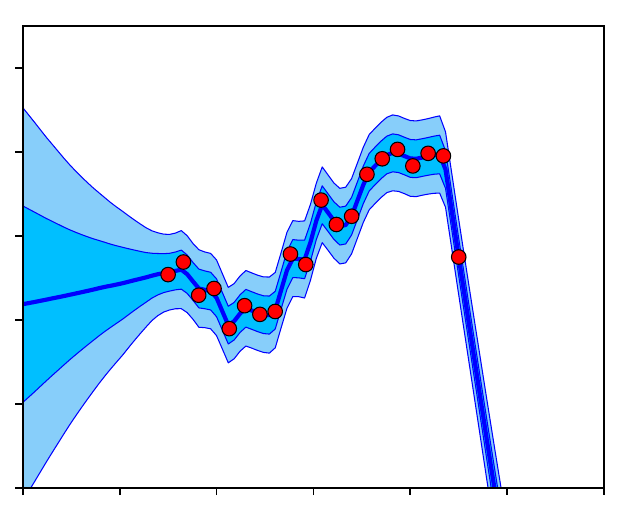}
        \put (4,50) {\small A. 10x Unconstrained NNs}
        \put (17,62) {\small Toy Regression Task}
        \end{overpic}
    \end{minipage}
    \hskip -0.08in
    \begin{minipage}{0.24\textwidth}
        \begin{overpic}[width=1.\textwidth, height=0.6\textwidth]{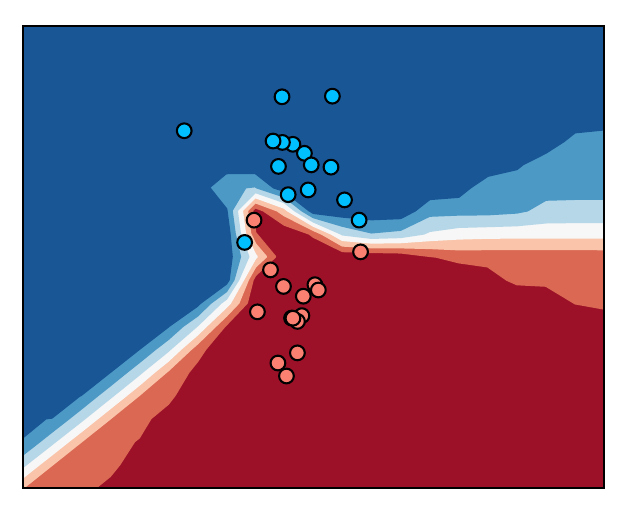}
        \put (11,62) {\small Toy Classification Task}
        \end{overpic}
    \end{minipage}
    \vskip -0.03in
    \begin{minipage}{0.24\textwidth}
        \centering
        \begin{overpic}[width=1.\textwidth, height=0.6\textwidth]{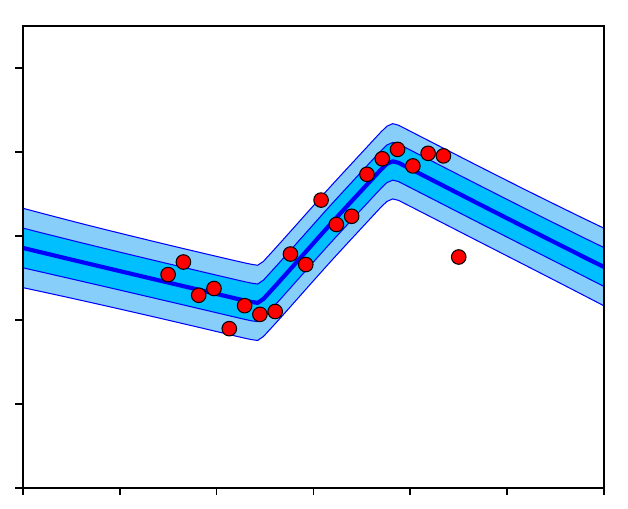}  
        \put (4,50) {\small B. 10x Regularised NNs}
        \end{overpic}
    \end{minipage}
    \hskip -0.08in
    \begin{minipage}{0.24\textwidth}
        \centering
        \begin{overpic}[width=1.\textwidth, height=0.6\textwidth]{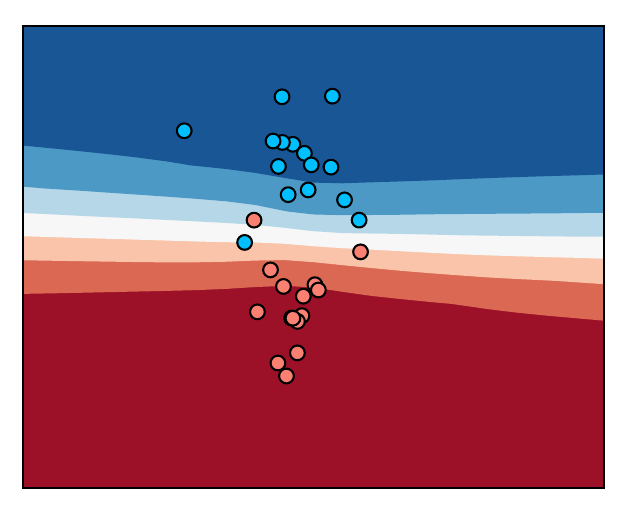}  
        \end{overpic}
    \end{minipage}
    \vskip -0.03in
    \begin{minipage}{0.24\textwidth}
        \centering
        \begin{overpic}[width=1.\textwidth, height=0.6\textwidth]{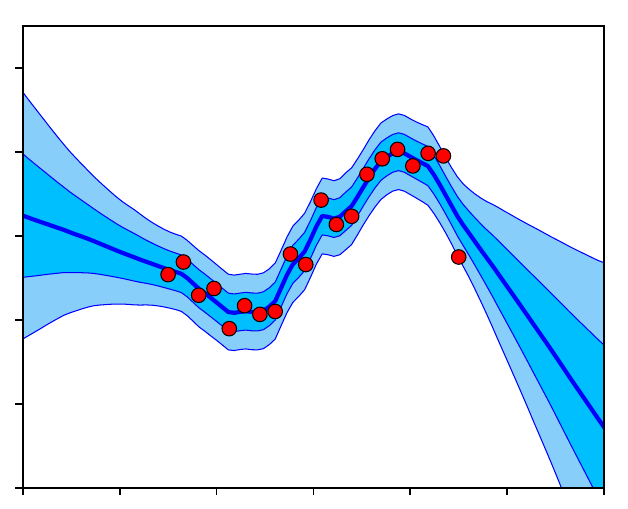}
        \put (4,50) {\small C. 10x Anchored NNs}
        \end{overpic}
    \end{minipage}
    \hskip -0.08in
    \begin{minipage}{0.24\textwidth}
        \centering
        \begin{overpic}[width=1.\textwidth, height=0.6\textwidth]{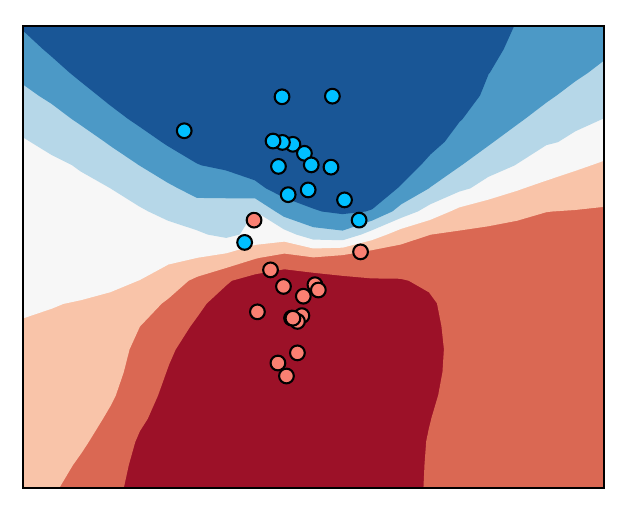}
        \end{overpic}
    \end{minipage}
    \vskip -0.03in
    \hskip 0.009in 
    \begin{minipage}{0.24\textwidth}
        \centering
        \begin{overpic}[width=1.\textwidth, height=0.6\textwidth]{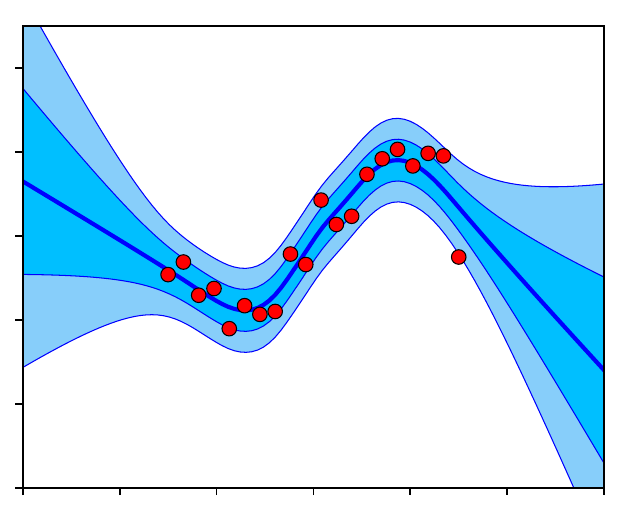} 
        \put (8,50) {\small D. Ground Truth}
        \end{overpic}
    \end{minipage}
    \hskip -0.08in
    \begin{minipage}{0.24\textwidth}
        \centering
        \begin{overpic}[width=1.\textwidth, height=0.6\textwidth]{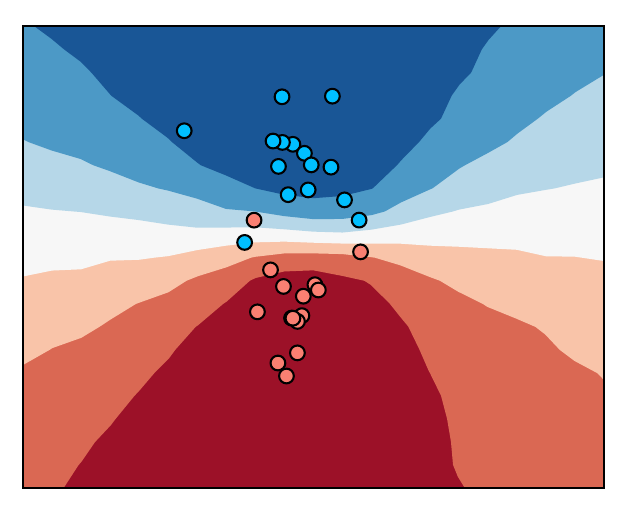}
        \end{overpic}
    \end{minipage}
    \vskip -0.1in
\caption{Comparison of NN ensemble loss choices.}
\vskip -0.2in
\label{fig_reg_or_not}
\end{center}
\end{figure}

Figures \ref{fig_methods}, \ref{fig_reg_or_not} \& \ref{fig_convergence_toy1d} show predictive distributions for anchored ensembles that very closely approximate the true Bayesian posterior, with little sign of bias. This demands an answer to why, rather than if, anchored ensembling performs such accurate inference in these examples. We believe the reason is the presence of the two special conditions that {can} lead to exact recovery.

It should be straightforward to see that extrapolation parameters (definition \ref{def_extrapolation parameters}) exist in the figures. Many hidden nodes will be dead across the range which contains data. Their corresponding final layer weight then has no effect on the data likelihood, but they do affect predictions outside of the training data.

It is more difficult to see that perfect correlations also exist, and we provide a numerical example illustrating this in appendix \ref{app_numerical}. Essentially it relies on two hidden nodes becoming live in between the same two data points. The associated final layer weights are then perfectly correlated.  Whether these special conditions exist beyond fully-connected NNs is something tested indirectly in later experiments with CNNs.




One obvious way to further encourage these conditions is to increase the width of the NN, creating more parameters and an increasing probability of strong correlations. See also a study of multicollinearity in NNs \citep{Cheng2018} [7.1].

\subsection{Implementation Practicalities}

\label{sec_implement}

\begin{figure*}[t!]
\vskip -0.15in
\begin{center}
    \begin{minipage}{0.20\textwidth}
        \centering
        \includegraphics[width=0.99\textwidth, height=0.7\textwidth]{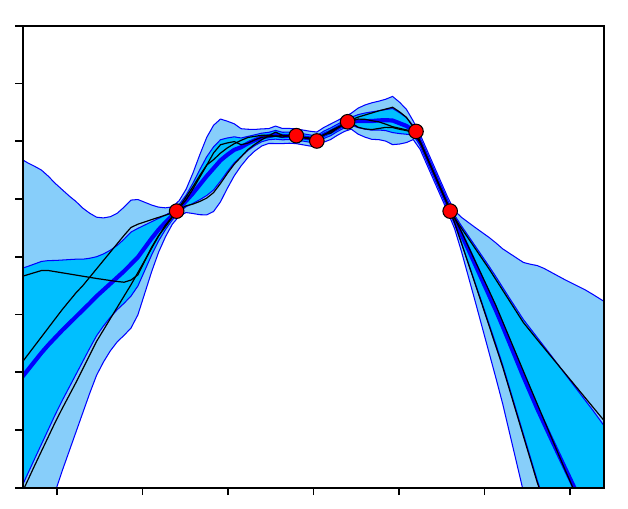}
        \put (-80,8) {\small 3xNNs}
    \end{minipage}
    \hspace{-0.12in}
    \begin{minipage}{0.20\textwidth}
        \centering
        \includegraphics[width=0.99\textwidth, height=0.7\textwidth]{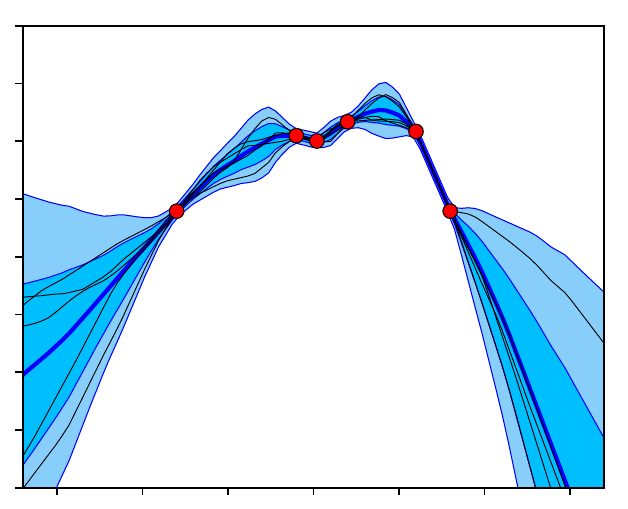}
        \put (-80,8) {\small 5xNNs}
    \end{minipage}
    \hspace{-0.12in}
    \begin{minipage}{0.20\textwidth}
        \centering
        \includegraphics[width=0.99\textwidth, height=0.7\textwidth]{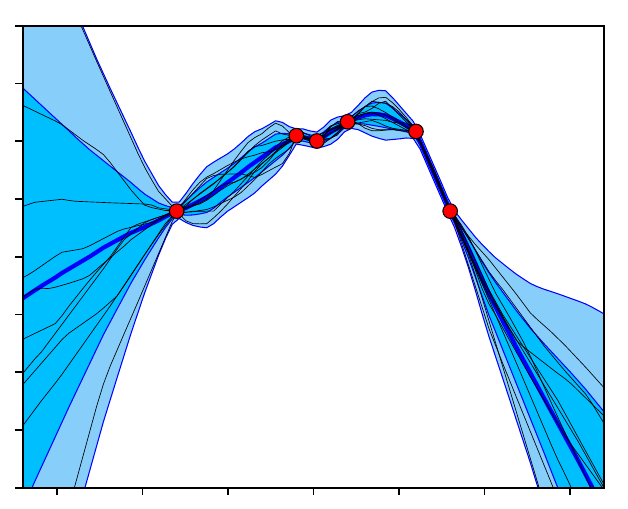}
        \put (-80,8) {\small 10xNNs}
    \end{minipage}
    \hspace{-0.12in}
    \begin{minipage}{0.20\textwidth}
        \centering
        \includegraphics[width=0.99\textwidth, height=0.7\textwidth]{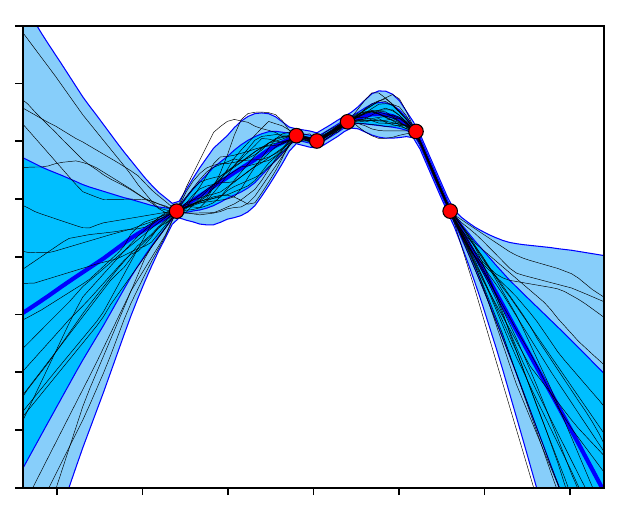}
        \put (-80,8) {\small 20xNNs}
    \end{minipage}
    \hspace{-0.12in}
    \begin{minipage}{0.20\textwidth}
        \centering
        \includegraphics[width=0.99\textwidth, height=0.7\textwidth]{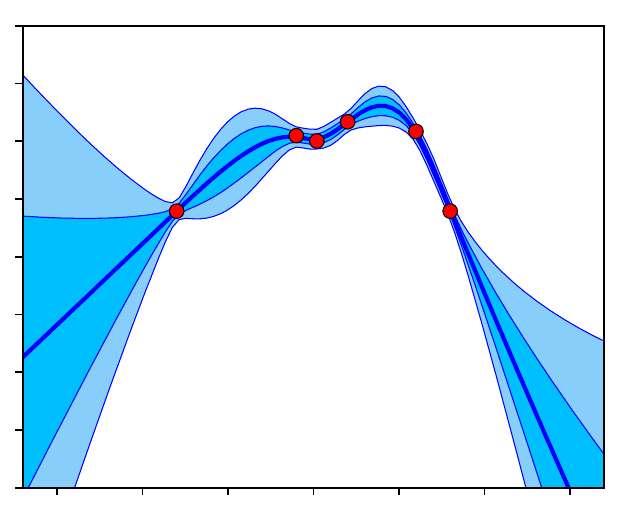}
        \put (-80,18) {\small Ground Truth}
        \put (-80,8) {\small - GP}
    \end{minipage}
\vskip -0.1in
\caption{The predictive distribution of an anchored ensemble approaches that of a ReLU GP.}
\label{fig_convergence_toy1d}
\end{center}
\vskip -0.2in
\end{figure*}

\textit{How many NNs to use in an RMS ensemble?}  A large number of samples (and therefore NNs) would be required to fully capture the posterior parameter distributions. By contrast, if one thinks of each NN as an iid sample from a posterior \textit{predictive} distribution, a much smaller number are required, given output dimensionality is typically small. Note this is unaffected by input dimension. Our experiments in section \ref{sec_results} used 5-10 NNs per ensemble, delivering good performance on tasks ranging from 1-10 outputs. See also figure \ref{fig_convergence_toy1d}. This results in anchored ensembles scaling by $\mathcal{O}(MN)$.


%

\textit{Should the NNs be initialised at anchor points?} It is convenient to draw parameter initialisations from the anchor distribution, and regularise directly around these initialised values, however, we found decoupling initialisations from anchor points benefited experiments. 

%


\begin{figure}[b!]
\vskip -0.2in
\begin{center}
\includegraphics[width=0.5\textwidth, height=0.32\textwidth]{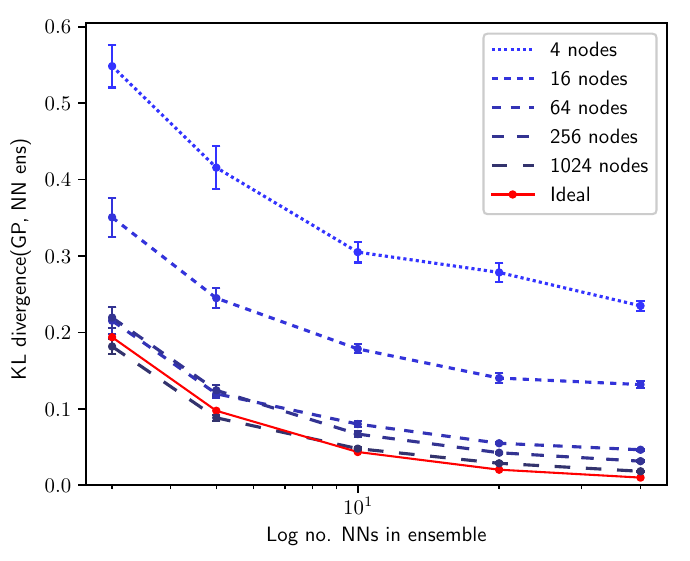}
 \vskip -0.15in
\caption{Difference in predictive distributions of an anchored ensemble and a ReLU GP as a function of width and number of NNs. Mean $\pm$1 standard error.}
\label{fig_KL_converge_plain}
\end{center}
\vskip -0.2in
\end{figure}

\section{Experiments}
\label{sec_results}
This section shares high-level findings from experiments. Further details and hyperparameter settings are given in appendix \ref{sec_exp_deets}. Appendix \ref{sec_app_results} additionally includes two RL experiments; one testing uncertainty-aware agents for model-free RL, and one applying anchored ensembles to noisy environments for model-based RL. Code is available online (\href{http://www.github.com/TeaPearce/Bayesian_NN_Ensembles}{\textcolor{blue}{github/TeaPearce}}). Also see our \href{https://teapearce.github.io/portfolio/github_io_1_ens/}{\textcolor{blue}{interactive demo}}.




\subsection{Qualitative Tests}
\label{sec_panel_results}
We first examine anchored ensembles on toy problems to gain intuition about its behaviour compared to popular approximate inference and ensembling methods.

Figure \ref{fig_methods} compares popular Bayesian inference methods in single-layer NNs for ReLU and sigmoidal non-linearities. GP and HMC produce `gold standard' Bayesian inference, and we judge the remaining methods, which are scalable approximations, to them. Both MFVI (with a factorised normal distribution) and MC dropout do a poor job of capturing interpolated uncertainty. This is a symptom of the posterior approximation ignoring parameter correlations - see also figure \ref{fig_RMS_general_special_egs} which shows MFVI failing to capture correlations in the posterior. This was explored in \cite{Foong2019}.


Figure \ref{fig_reg_or_not}  contrasts anchored ensembles trained on eq. \ref{eqn_anch_loss_matrix} \& \ref{eqn_anch_loss_matrix_class}, with NN ensembles using standard loss functions, either with no regularisation term (`unconstrained', $\pmb{\Gamma}=\mathbf{0}$), or regularised around zero (`regularised', $\pmb{\theta}_{anc,j} = \pmb{0}$). Regularised produces poor results since it encourages all NNs to the same single solution and diversity is reduced. Unconstrained is also inappropriate - although it produces diversity, no notion of prior is maintained and it overfits the data.


Figure \ref{fig_convergence_toy1d} shows the predictive distribution improving with number of NNs compared to a ReLU GP, however it appears a small residual difference remains.


\subsection{Convergence Behaviour}
\label{sec_result_vis_converg}

To assess how precisely anchored ensembling performs Bayesian inference on a real dataset, we compared its predictive distribution with that of an exact method (ReLU GP) on the Boston housing dataset. Figure \ref{fig_KL_converge_plain} quantifies the difference when varying both the width of the NN, and number of NNs in the ensemble. KL divergence between the two predictive distributions was measured and found to decrease as both NN width and number of NNs was increased. As in figure \ref{fig_convergence_toy1d} a small amount of residual difference remains even for 40xNNs of $1,024$ nodes. 


\subsection{UCI Regression Benchmarks}
\label{sec_results_bench}
In order to compare anchored ensembles against popular approximate inference methods, we used a standard BNN benchmark. This assesses uncertainty quality for UCI regression tasks on data drawn from the same distribution as the training data \citep{Hernandez-Lobato2015}. We also implemented the ReLU GP to assess the performance limit on these datasets.

Table \ref{tab_regression} lists our results. We include results reported for Deep Ensembles \citep{Lakshminarayanan2016}, which is considered the state-of-the-art ensemble method. Appendix \ref{app_regression_bench} provides a full comparison with other approximate Bayesian methods including Probabilistic Backpropagation, MC Dropout, and Stochastic Gradient HMC.

Ordering results according to the level of estimated data noise, $\hat{\sigma}^2_{\epsilon}$, shows a clear pattern - anchored ensembles perform best in datasets with low data noise, surpassing both Deep Ensembles and all approximate inference methods listed in appendix \ref{app_regression_bench}. This may be due to an increased importance of interpolation uncertainty when data noise is low, which anchored ensembles models well. On other datasets, the method is also competitive (the Deep Ensemble implementation used additional complexity to capture heteroskedastic uncertainty and has an advantage on higher data noise datasets).








\subsection{Out-of-Distribution Classification}
\label{sec_results_img_class}

We now test on classification tasks, for out-of-distribution (OOD) data, with complex NN architectures, and compare against other ensemble methods.

An uncertainty-aware NN should make predictions of decreasing confidence as it is asked to predict on data further from the distribution seen during training. To test this, we report the proportion of high confidence predictions (defined as a softmax output class being $\geq$ 90\%) made by various ensemble systems - unconstrained, regularised, and anchored (as in section \ref{sec_panel_results}).

We trained on three different datasets, using a NN architecture appropriate to each: 1) Fashion MNIST image classification; 3 fully-connected layers of 100 hidden nodes. 2) IMDb movie review sentiment classification; embedding + 1D convolution + fully-connected layer. 3) CIFAR-10 image classification; convolutional NN (CNN) similar to VGG-13 (9 million parameters). 

The confidence of predictions on novel data categories not seen during training was assessed. Table \ref{tab_fashion_mini} shows example OOD images shown to the NNs trained on CIFAR-10. Edge refers to two CIFAR classes held out during training (ships, dogs). Appendix \ref{sec_exp_deets} provides OOD examples for other datasets.


The three tables show similar patterns. Whilst all methods predict with similar confidence on the training data, confidence differs greatly for other data categories, with anchored ensembles generally producing the most conservative predictions. This gap increases for data drawn further from the training distribution. Encouragingly, we observe similar (though less extreme) behaviour to that in the toy examples of figure \ref{fig_reg_or_not}. 

\begin{table}[t]%
\caption{NLL regression benchmark results. See appendix \ref{sec_app_results} for RMSE and variants of our method. Mean $\pm$1 standard error. }
\begin{center}
\resizebox{1.\columnwidth}{!}{
\begin{tabular}{ l r rrr }
\Xhline{3\arrayrulewidth}
\multicolumn{1}{c}{}  & \multicolumn{1}{c}{}  &  \multicolumn{1}{c}{Deep Ens.} &  \multicolumn{1}{c}{Anch. Ens.} &  \multicolumn{1}{c}{ReLU GP$^1$}  \\ 

\multicolumn{1}{c}{}   & \multicolumn{1}{r}{$\hat{\sigma}^2_{\epsilon}$} &  \multicolumn{1}{c}{\small{\textit{State-Of-Art}}}  &  \multicolumn{1}{c}{\small{\textit{Our Method}}} &  \multicolumn{1}{c}{\small{\textit{Gold Standard}}}   \\ 

\hline 

\multicolumn{1}{c}{}  &\multicolumn{4}{c}{High Epistemic Uncertainty}\\
Energy& 1e-7&  {1.38 $\pm$ 0.22} & \bftab\textcolor{blue}{0.96 $\pm$ 0.13} & {0.86 $\pm$ 0.02}  \\
Naval & 1e-7& {-5.63 $\pm$ 0.05}  & \bftab\textcolor{blue}{-7.17 $\pm$ 0.03} & {-10.05 $\pm$ 0.02}  \\
Yacht & 1e-7 & {1.18 $\pm$ 0.21}  & \bftab\textcolor{blue}{0.37 $\pm$ 0.08} & {0.49 $\pm$ 0.07}  \\
\hline 
\multicolumn{1}{c}{}  & \multicolumn{4}{c}{Equal Epistemic \& Aleatoric Uncertainty}\\
Kin8nm & 0.02&  {-1.20 $\pm$ 0.02}  & \bftab\textcolor{blue}{-1.09 $\pm$ 0.01} & {-1.22 $\pm$ 0.01}  \\
Power  & 0.05&  \bftab\textcolor{blue}{2.79 $\pm$ 0.04} & {2.83 $\pm$ 0.01} & {2.80 $\pm$ 0.01}  \\
Concrete  & 0.05&  {3.06 $\pm$ 0.18} & \bftab\textcolor{blue}{2.97 $\pm$ 0.02} & {2.96 $\pm$ 0.02}  \\
Boston  & 0.08&  \bftab\textcolor{blue}{2.41 $\pm$ 0.25} & {2.52 $\pm$ 0.05} & {2.45 $\pm$ 0.05}  \\
\hline 
\multicolumn{1}{c}{}  & \multicolumn{4}{c}{High Aleatoric Uncertainty}\\
Protein  & 0.5 & \bftab\textcolor{blue}{2.83 $\pm$ 0.02}  & {2.89 $\pm$ 0.01} & {*2.88 $\pm$ 0.00}  \\
Wine  & 0.5 & \bftab\textcolor{blue}{0.94 $\pm$ 0.12}  & \bftab\textcolor{blue}{0.95 $\pm$ 0.01} & {0.92 $\pm$ 0.01}  \\
Song  & 0.7 & \bftab\textcolor{blue}{3.35 $\pm$ NA } & {3.60 $\pm$ NA } & {**3.62 $\pm$ NA }  \\


\Xhline{3\arrayrulewidth} 

\end{tabular}
}
\end{center}
\vskip -0.05in
\tiny{$^1$ For comparison only (not a scalable method).} \tiny{* Trained on $10,000$ rows of data.} \tiny{** Trained on $20,000$ rows of data, tested on $5,000$ data points.}
\vskip -0.2in
\label{tab_regression}
\end{table}

%

\section{Conclusion}

This paper proposed, analysed, and tested a modification to the usual NN ensembling process that results in approximate Bayesian inference - regularising parameters around values drawn from a prior distribution. 

Under simplifying assumptions, we derived an abstracted form of RMS motivating this. We analysed a practical RMS variant to understand the bias of its approximate posterior. Two special conditions were shown to lead to recovery of the true posterior: perfectly correlated parameters and extrapolation parameters. We discussed the validity of applying RMS to NNs, arguing that these two special conditions are partially present in NNs. 

On regression benchmarking experiments, state-of-the-art performance was achieved on 3/10 datasets - outperforming popular approximate inference methods. On image and text classification tasks, anchored ensembles were shown to be more robust than alternative ensemble methods.


\begin{table}[t]

\vspace{-0.1in}

\caption{Proportion of predictions that were high confidence on out-of-distribution data, e.g. a single regularised NN trained on CIFAR-10 made high confidence predictions 54\% of the time when asked to predict on MNIST. Mean over five runs (three for CIFAR).}

\vspace{0.05in}
\begin{center}
\small{CIFAR-10 Image Classification, VGG-13 CNN}
\end{center}

\includegraphics[width=0.49\textwidth]{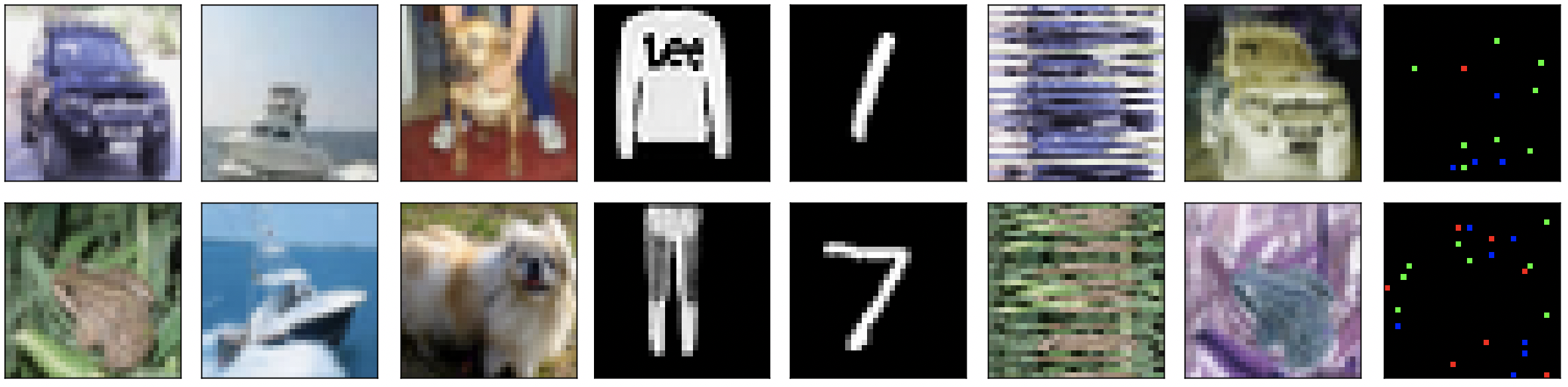}
\put (-234,62) {\scriptsize Train}
\put (-198,62) {\scriptsize --- Edge ---}
\put (-150,62) {\scriptsize Fashion}
\put (-118,62) {\scriptsize MNIST}
\put (-89,62) {\scriptsize Scramble}
\put (-55,62) {\scriptsize Invert}
\put (-25,62) {\scriptsize Noise}
\vspace{0.05in}

\resizebox{1.\columnwidth}{!}{


\begin{tabular}{ l | c | c | cccccc }
\Xhline{3\arrayrulewidth}
\multicolumn{1}{c}{}   & \multicolumn{1}{c}{ \textcolor{white}{--}  Accuracy \textcolor{white}{--}  } &  \multicolumn{1}{c}{ \textcolor{white}{--} Train \textcolor{white}{--}  } &  \multicolumn{1}{c}{Edge} &  \multicolumn{1}{c}{Fashion} &  \multicolumn{1}{c}{MNIST} &  \multicolumn{1}{c}{Scramble} &  \multicolumn{1}{c}{Invert} &  \multicolumn{1}{c}{Noise} \\ 
\hline


1xNNs Reg. & { 81.6\%} & { 0.671} & { 0.466} & { 0.440} & { 0.540} & { 0.459} & { 0.324} & { 0.948} \\
5xNNs Uncons. & { 85.0\%} & { 0.607} & { 0.330} & { 0.208} & { 0.275} & { 0.175} & { 0.209} & { 0.380} \\
5xNNs Reg. & { 86.1\%} & { 0.594} & { 0.296} & { 0.219} & { 0.188} &\textcolor{blue}{ \textbf{ 0.106}} & { 0.153} & { 0.598} \\
5xNNs Anch. & { 85.6\%} & { 0.567} & \textcolor{blue}{ \textbf{ 0.258}} & \textcolor{blue}{ \textbf{ 0.184}} & \textcolor{blue}{ \textbf{ 0.149}} & { 0.134} & \textcolor{blue}{ \textbf{ 0.136}} & \textcolor{blue}{ \textbf{ 0.118}} \\
\Xhline{1\arrayrulewidth}
10xNNs Anch. & { 86.0\%} & { 0.549} & { 0.256} & { 0.119} & { 0.145} & { 0.122} & { 0.124} & { 0.161} \\
\hline
\end{tabular}
}

\vspace{0.05in}
\begin{center}
\small{IMDb Text Sentiment Classification, Embedding+CNN}
\end{center}
\vspace{-0.02in}
\resizebox{1.\columnwidth}{!}{

\begin{tabular}{ l | c| c | cccc }
\Xhline{3\arrayrulewidth}
\multicolumn{1}{c}{} & \multicolumn{1}{c}{ \textcolor{white}{--}  Accuracy \textcolor{white}{--}  }  & \multicolumn{1}{c}{ \textcolor{white}{--}  Train \textcolor{white}{--}  } &  \multicolumn{1}{c}{ \textcolor{white}{--} Reuters \textcolor{white}{--}  } &  \multicolumn{1}{c}{Rand. 1} &  \multicolumn{1}{c}{Rand. 2} &  \multicolumn{1}{c}{Rand. 3} \\ 
\hline
1xNNs Reg. & { 85.3\%} & { 0.637} & { 0.119} & { 0.153} & { 0.211} & { 0.326} \\
5xNNs Uncons. & { 89.1\%} & { 0.670} & { 0.102} & { 0.141} & { 0.100} & { 0.075} \\
5xNNs Reg. & { 87.1\%} & { 0.612} & { 0.051} & { 0.091} & { 0.076} & { 0.055} \\
5xNNs Anch. & { 87.7\%} & { 0.603} &  \textcolor{blue}{ \textbf{ 0.049}} &  \textcolor{blue}{ \textbf{ 0.075}} &  \textcolor{blue}{ \, \textbf{0.061}} & \textcolor{blue}{ \textbf{ 0.009}} \\
\Xhline{3\arrayrulewidth}

%
\end{tabular}
}

\vspace{0.05in}
\begin{center}
\small{Fashion MNIST Image Classification, Fully-Connected NN}
\end{center}
\vspace{-0.02in}
\resizebox{1.\columnwidth}{!}{
\begin{tabular}{ l | c| c| ccccc }

\Xhline{3\arrayrulewidth}
\multicolumn{1}{c}{} & \multicolumn{1}{c}{ Accuracy  } & \multicolumn{1}{c}{ \textcolor{white}{--}  Train \textcolor{white}{--}  }  & \multicolumn{1}{c}{ \textcolor{white}{}  Edge \textcolor{white}{}  } &  \multicolumn{1}{c}{ \textcolor{white}{} CIFAR \textcolor{white}{}  } &  \multicolumn{1}{c}{ MNIST  } &  \multicolumn{1}{c}{Distort} &  \multicolumn{1}{c}{Noise}  \\ 
\hline

1xNN Reg. & { 86.8 \%} & { 0.660} & { 0.584} & { 0.143} &{ 0.160} & { 0.429} & { 0.364}\\
5xNNs Uncons. & { 89.0 \%}& { 0.733} & { 0.581} & { 0.301} &{ 0.104} & { 0.364} & { 0.045}\\
5xNNs Reg. & { 87.8 \%}& { 0.634} & \textcolor{blue}{ \textbf{0.429}} & { 0.115} &{ 0.072} & { 0.342} & { 0.143}\\
5xNNs Anch. & { 88.0 \%}& { 0.631} & { 0.452} & \textcolor{blue}{ \textbf{0.065}} & \textcolor{blue}{ \textbf{0.041}} & \textcolor{blue}{ \textbf{0.246}} & \textcolor{blue}{ \textbf{0.006}}\\

\Xhline{3\arrayrulewidth}
\end{tabular}
}

\vskip -0.1in
\label{tab_fashion_mini}
\end{table}

\FloatBarrier

\section*{Acknowledgements}

Thanks to all anonymous reviewers for their helpful comments  and  suggestions. The lead author was funded through EPSRC (EP/N509620/1) and partially accommodated by the Alan Turing Institute. The model-based RL experiments were run during an internship at PROWLER.io. Thanks to Nicolas Anastassacos for collaborating on an early version of the paper, and Ayman Boustati and Ahmed Al-Ali for helpful discussions.


\bibliography{library}
\bibliographystyle{apalike}

\include{appendices}
\begin{appendices}

\FloatBarrier
\onecolumn

\begin{center}\Large \bfseries
Appendix to \\ Uncertainty in Neural Networks: \\ Approximately Bayesian Ensembling
\end{center}

\section{Proofs}
\label{app_proof}

\begin{definition}{Data likelihood and parameter likelihood}
\label{def_likelihoods}

We take care to define two versions of the likelihood, one in output space, $P_{\mathcal{D} }(\mathcal{D} | \pmb{\theta})$ (data likelihood), and one in parameter space, $P_{\pmb{\theta}}(\mathcal{D} | \pmb{\theta})$ (parameter likelihood). Both return the same values given some data set $\mathcal{D}$ and parameter values $\pmb{\theta}$, and hence are exchangeable, but their forms are subtly different. 

The data likelihood, $ P_{\mathcal{D} }(\mathcal{D} | \pmb{\theta})$, is defined on the output domain. Typically the log of this, $\log( P_{\mathcal{D} }(\mathcal{D} | \pmb{\theta}) )$, might be optimised as the cross entropy loss or (negative) mean squared error. 

In contrast, $P_{\pmb{\theta}}(\mathcal{D} | \pmb{\theta})$ defines a likelihood function in the parameter domain.

\textbf{Illustrative Example}

Consider a linear regression model with dataset, $\mathcal{D}$,  consisting of tuples, $\{ \mathbf{x} , y\}$; a vector of predictor variables $\mathbf{x} \in \mathbb{R}^p$, predicting a single scalar $y \in \mathbb{R}$. If the model is of the form, $\pmb{\theta}^T \mathbf{x}$, a Gaussian data likelihood with variance $\sigma^2_{\epsilon}$ on the output might be assumed. 

This leads to a data likelihood for the target, $y$,
\begin{equation}
P_{\mathcal{D} }(\mathcal{D} | \pmb{\theta}) = \mathcal{N}(y | \pmb{\theta}^T \mathbf{x}, \sigma^2_{\epsilon}).
\end{equation}
For this linear model, the corresponding parameter likelihood is a multivariate normal distribution,
\begin{equation}
P_{\pmb{\theta} }(\mathcal{D} | \pmb{\theta}) \propto \mathcal{N}(\pmb{\theta} | \pmb{\mu}_{like}, \pmb{\Sigma}_{like} ).
\end{equation}
where, $\pmb{\mu}_{like} \in \mathbb{R}^p$ \& $\pmb{\Sigma}_{like} \in \mathbb{R}^{p \times p}$, can be found analytically. They are implicitly functions of the dataset, $\mathcal{D}$, although to lighten notation we do not write this. Subsequently we also drop the explicit referral to $\pmb{\theta}$.

Note that whilst both the data and parameter likelihood follow a normal distribution, they are defined in different domains.

The correspondence between a Gaussian data likelihood and multivariate normal parameter likelihood is only exact for a linear regression model. For non-linear models with Gaussian data likelihoods, and other data likelihoods, the parameter likelihood is not in general multivariate normal. Nevertheless it can be convenient to model it as such.




\end{definition}

\vspace{0.1in}

\textbf{Standard Result 1. } Product of two multivariate Gaussians (\S8.1.8, The Matrix Cookbook,  \citeyear{Pedersen2008}) 
\begin{equation}
\mathcal{N}(\pmb{\mu}_{like},\pmb{\Sigma}_{like}) \mathcal{N}(\pmb{\mu}_{prior},\pmb{\Sigma}_{prior}) \propto \mathcal{N}(\pmb{\mu}_{post},\pmb{\Sigma}_{post})
\end{equation}
\begin{equation}
\label{eq_sig_post}
\pmb{\Sigma}_{post} =  ( \pmb{\Sigma}^{-1}_{prior} + \pmb{\Sigma}^{-1}_{like} )^{-1},
\end{equation}
\begin{equation}
\label{eq_mu_post}
\pmb{\mu}_{post} = \pmb{\Sigma}_{post} \pmb{\Sigma}^{-1}_{prior} \pmb{\mu}_{prior} + \pmb{\Sigma}_{post} \pmb{\Sigma}^{-1}_{like} \pmb{\mu}_{like}.
\end{equation}



\textbf{Standard Result 2. } Affine transform of a normal random variable (\S8.1.4, The Matrix Cookbook, \citeyear{Pedersen2008}) 
\begin{equation}
\mathbf{x} \sim \mathcal{N}( \pmb{\mu}  ,\pmb{\Sigma}),
\end{equation}
\begin{equation}
\label{eq_affine_norm}
\mathbf{y} = \mathbf{A} \mathbf{x} + \mathbf{b},
\end{equation}
\begin{equation}
\mathbf{y} \sim \mathcal{N}( \mathbf{A} \pmb{\mu} +  \mathbf{b} , \mathbf{A} \pmb{\Sigma}  \mathbf{A}^T ).
\end{equation}

\vspace{0.2in}

\begin{theorem}{}
\label{theorem_anchored_cov}
Assume that a model's parameter likelihood follows a multivariate normal distribution, $P_{\pmb{\theta}}(\mathcal{D} | \pmb{\theta}) \propto \mathcal{N}(\pmb{\mu}_{like},\pmb{\Sigma}_{like})$, and the prior also, $P(\pmb{\theta}) = \mathcal{N}(\pmb{\mu}_{prior},\pmb{\Sigma}_{prior})$. The posterior is then also multivariate normal, $P(\pmb{\theta}|\mathcal{D}) = \mathcal{N}(\pmb{\mu}_{post},\pmb{\Sigma}_{post})$. 

Further assume availability of some function which returns MAP parameter estimates taking as input the location of the prior centre, $\pmb{f}_{\text{MAP}}(\pmb{\theta}_{anc})$. In order that, $P(\pmb{f}_{\text{MAP}}(\pmb{\theta}_{anc})) = P(\pmb{\theta}|\mathcal{D})$, then the required distribution of $\pmb{\theta}_{anc}$ is also multivariate normal, $P(\pmb{\theta}_{anc}) =  \mathcal{N}(\pmb{\mu}_{anc},\pmb{\Sigma}_{anc})$, where, $\pmb{\mu}_{anc} = \pmb{\mu}_{prior}$, and, $\pmb{\Sigma}_{anc} = \pmb{\Sigma}_{prior} +  \pmb{\Sigma}_{prior} \pmb{\Sigma}_{like}^{-1} \pmb{\Sigma}_{prior}$.

\end{theorem}

\begin{proof}

Consider a model's parameters having a multivariate normal prior,

\begin{equation}
\label{eq_prior2}
P(\pmb{\theta}) = \mathcal{N}(\pmb{\mu}_{prior},\pmb{\Sigma}_{prior}),
\end{equation}

where, $\pmb{\theta} \in \mathbb{R}^p$, $\pmb{\mu}_{prior} \in \mathbb{R}^p$, $\pmb{\Sigma}_{prior} \in \mathbb{R}^{p \times p}$.

This theorem makes the assumption that the form of the parameter likelihood (def. \ref{def_likelihoods}) is multivariate normal,

\begin{equation}
\label{eq_like}
P_{\pmb{\theta}}(\mathcal{D} | \pmb{\theta})  \propto \mathcal{N}(\pmb{\mu}_{like},\pmb{\Sigma}_{like})
\end{equation}

where, $\pmb{\mu}_{like} \in \mathbb{R}^p$, $\pmb{\Sigma}_{like} \in \mathbb{R}^{p \times p}$. Here $\propto$ is used since it is not a true probability distribution in $\pmb{\theta}$ so need not sum to $1$. 

%

The posterior is calculated by Bayes rule. Recalling that data likelihood and parameter likelihood are exchangeable (def. \ref{def_likelihoods}), and using Standard Result 1,

\begin{equation}
\label{eq:bayes}
P(\pmb{\theta}|\mathcal{D}) 
=  \frac{P_{\mathcal{D} }(\mathcal{D} |\pmb{\theta})  P(\pmb{\theta})}{P(\mathcal{D})} 
=  \frac{P_{\pmb{\theta}}(\mathcal{D} |\pmb{\theta})  P(\pmb{\theta})}{P(\mathcal{D})} 
\propto \mathcal{N}(\pmb{\mu}_{like},\pmb{\Sigma}_{like}) \mathcal{N}(\pmb{\mu}_{prior},\pmb{\Sigma}_{prior}) 
\propto \mathcal{N}(\pmb{\mu}_{post},\pmb{\Sigma}_{post}),
\end{equation}

where, $\pmb{\mu}_{post}$ \& $\pmb{\Sigma}_{post}$ are given by eq.  \ref{eq_mu_post} \& \ref{eq_sig_post}.

We introduce a further distribution, termed `anchor distribution', which we enforce as multivariate normal,

\begin{equation}
P(\pmb{\theta}_{anc}) = \mathcal{N}(\pmb{\mu}_{anc},\pmb{\Sigma}_{anc}).
\end{equation}

It will be used as described in the main text (see figure \ref{fig_randomised_anch_inference} and algorithm \ref{alg_RL_ens}) so that samples are drawn from the anchor distribution, $\pmb{\theta}_{anc} \sim P(\pmb{\theta}_{anc}) $, with a prior then recentred at each sample, denoted $P_{anc}(\pmb{\theta})$,

\begin{equation}
\label{eq_p_0_theta}
P_{anc}(\pmb{\theta}) = \mathcal{N}( \pmb{\theta}_{anc}  ,\pmb{\Sigma}_{prior}).
\end{equation}

Note that this anchor distribution is in the same position as a hyperprior on $\pmb{\mu}_{prior}$, but will have a subtly different role. $\pmb{\Sigma}_{prior}$ is unchanged from eq. \ref{eq_prior2},

Denote $\pmb{f}_{\text{MAP}}(\pmb{\theta}_{anc})$ as the MAP estimates given this recentred prior and the original likelihood from eq. \ref{eq_like}.

\begin{equation}
\pmb{f}_{\text{MAP}}(\pmb{\theta}_{anc}) \coloneqq \text{argmax}_{\pmb{\theta}}  P_{anc}(\pmb{\theta}) P_{\pmb{\theta}}(\mathcal{D} | \pmb{\theta}) 
\end{equation}

In order to prove the theorem, three things regarding $\pmb{f}_{\text{MAP}}(\pmb{\theta}_{anc})$ must be shown: 
\begin{enumerate}
\item Its distribution is multivariate normal - denote mean and covariance $\pmb{\mu}_{post}^{\text{RMS}},\pmb{\Sigma}_{post}^{\text{RMS}}$,
\begin{equation}
P(\pmb{f}_{\text{MAP}}(\pmb{\theta}_{anc})) = \mathcal{N}(\pmb{\mu}_{post}^{\text{RMS}},\pmb{\Sigma}_{post}^{\text{RMS}}),
\end{equation}
\item That $\pmb{\mu}_{anc}$ \& $\pmb{\Sigma}_{anc}$ can be selected in such a way that the mean of the distribution is equal to that of the original posterior
\begin{equation}
\pmb{\mu}_{post}^{\text{RMS}} = \pmb{\mu}_{post},
\end{equation}
\item And also so that the covariance of the distribution is equal to that of the original posterior
\begin{equation}
\pmb{\Sigma}_{post}^{\text{RMS}} = \pmb{\Sigma}_{post}.
\end{equation}

\end{enumerate}


For a multivariate normal distribution, the MAP solution is simply equal to the mean of the posterior, $\pmb{\mu}_{post}$. For the typical case this is given by eq. \ref{eq_mu_post}. In our procedure, the location of the prior mean has been replaced by $\pmb{\theta}_{anc}$, so the MAP solution is given by,
\begin{align}
\pmb{f}_{\text{MAP}}(\pmb{\theta}_{anc}) &= \
\pmb{\Sigma}_{post} \pmb{\Sigma}^{-1}_{prior} \pmb{\theta}_{anc} + \pmb{\Sigma}_{post}  \pmb{\Sigma}^{-1}_{like} \pmb{\mu}_{like} \\
&= \mathbf{A}_1 \pmb{\theta}_{anc}  + \mathbf{b}_1 \label{eq_linear_form}
\end{align}
where two constants have been defined for convenience,
\begin{align}
\mathbf{A}_1 &= \pmb{\Sigma}_{post} \pmb{\Sigma}^{-1}_{prior} \\
\mathbf{b}_1 &= \pmb{\Sigma}_{post} \pmb{\Sigma}^{-1}_{like} \pmb{\mu}_{like},
\end{align}

which is the same form as eq. \ref{eq_affine_norm}. Hence, from Standard Result 2, if $\pmb{\theta}_{anc}$ is normally distributed,  $\pmb{f}_{\text{MAP}}(\pmb{\theta}_{anc}) $ \textbf{will also be normally distributed}. 

Regarding the mean of $\pmb{f}_{\text{MAP}}(\pmb{\theta}_{anc})$, we have,
\begin{align}
\mathbb{E}[\pmb{f}_{\text{MAP}}(\pmb{\theta}_{anc}) ] & = \mathbb{E}[ \mathbf{A}_1 \pmb{\theta}_{anc} + \mathbf{b}_1] \\
& = \mathbf{A}_1 \mathbb{E}[ \pmb{\theta}_{anc} ] + \mathbf{b}_1.
\intertext{By choosing the anchor distribution to be centred about the original prior, $\mathbb{E}[ \pmb{\theta}_{anc}] = \pmb{\mu}_{prior}$, we have,}
&= \mathbf{A}_1  \pmb{\mu}_{prior} + \mathbf{b}_1 \\
&= \pmb{\Sigma}_{post} \pmb{\Sigma}^{-1}_{prior} \pmb{\mu}_{prior} + \pmb{\Sigma}_{post} \pmb{\Sigma}^{-1}_{like} \pmb{\mu}_{like},
\end{align}

This is consistent with eq. \ref{eq_mu_post} and proves that \textbf{the means of the distributions are aligned when }$\pmb{\mu}_{anc} = \pmb{\mu}_{prior}$.

Finally we consider the variance of $\pmb{f}_{\text{MAP}}(\pmb{\theta}_{anc}) $, which we wish to equal $\pmb{\Sigma}_{post}$ by choosing $\pmb{\Sigma}_{anc}$. Using the form from eq. \ref{eq_linear_form} and applying Standard Result 2,
\begin{align}
\mathbb{V}ar[\pmb{f}_{\text{MAP}}(\pmb{\theta}_{anc}) ] &= \mathbb{V}ar[ \mathbf{A}_1 \pmb{\theta}_{anc} + \mathbf{b}_1] \\
&= \mathbf{A}_1 \mathbb{V}ar[ \pmb{\theta}_{anc}] \mathbf{A}^T_1 \\
&= \mathbf{A}_1 \pmb{\Sigma}_{anc}  \mathbf{A}^T_1 \label{eq_anc_MAP_post}
\end{align}
We require $\mathbb{V}ar[\pmb{f}_{\text{MAP}}(\pmb{\theta}_{anc}) ] = \pmb{\Sigma}_{post}$.
\begin{equation}
\label{eq_post_A_anch_A}
\pmb{\Sigma}_{post} = \mathbf{A}_1 \pmb{\Sigma}_{anc} \mathbf{A}^T_1.
\end{equation}
Note that transposes of covariance matrices may be ignored since they are symmetric.
\begin{align}
\pmb{\Sigma}_{anc} &= \mathbf{A_1}^{-1}   \pmb{\Sigma}_{post} \mathbf{A_1}^{-1T} \\
&= (\pmb{\Sigma}_{post} \pmb{\Sigma}_{prior} ^{-1})^{-1} \pmb{\Sigma}_{post}  ( \pmb{\Sigma}_{prior} ^{-1} \pmb{\Sigma}_{post}  )^{-1} \\
&= \pmb{\Sigma}_{prior} \pmb{\Sigma}_{post} ^{-1}\pmb{\Sigma}_{post}  \pmb{\Sigma}_{post}^{-1}  \pmb{\Sigma}_{prior} \\
&= \pmb{\Sigma}_{prior} \pmb{\Sigma}_{post} ^{-1}  \pmb{\Sigma}_{prior} \\
&= \pmb{\Sigma}_{prior} (\pmb{\Sigma}_{prior}^{-1} + \pmb{\Sigma}_{like}^{-1})  \pmb{\Sigma}_{prior} \\
&= \pmb{\Sigma}_{prior} +  \pmb{\Sigma}_{prior} \pmb{\Sigma}_{like}^{-1}  \pmb{\Sigma}_{prior}.
\label{eq_anch_full_ap}
\end{align}
This proves that \textbf{the covariances of the two distributions are aligned when} $\pmb{\Sigma}_{anc} =  \pmb{\Sigma}_{prior} +  \pmb{\Sigma}_{prior} \pmb{\Sigma}_{like}^{-1}  \pmb{\Sigma}_{prior}$.

\end{proof}

\begin{corollary}{}
\label{cor_anc_eq_prior}
Following from theorem \ref{theorem_anchored_cov} (and under the same assumptions), set $\pmb{\mu}_{anc} \coloneqq \pmb{\mu}_{prior}$ and $\pmb{\Sigma}_{anc} \coloneqq \pmb{\Sigma}_{prior}$. The RMS approximate posterior is $P(\pmb{f}_{\text{MAP}}(\pmb{\theta}_{anc})) = \mathcal{N}(\pmb{\mu}_{post}, \pmb{\Sigma}_{post} \pmb{\Sigma}_{prior}^{-1} \pmb{\Sigma}_{post})$.
\end{corollary}

\begin{proof}
Independent of the choice of anchor distribution covariance $\pmb{\Sigma}_{anc}$, theorem \ref{theorem_anchored_cov} demonstrated that the resulting posterior, $P(\pmb{f}_{\text{MAP}}(\pmb{\theta}_{anc}))$ is normally distributed, with mean equal to that of the true posterior $\pmb{\mu}_{post}$.

To discover the covariance of the resulting distribution,  $\mathbb{V}ar[\pmb{f}_{\text{MAP}}(\pmb{\theta}_{anc}) ] $, we take eq. \ref{eq_anc_MAP_post} and simply set, $\pmb{\Sigma}_{anc} \coloneqq \pmb{\Sigma}_{prior}$.
\begin{align}
\mathbb{V}ar[\pmb{f}_{\text{MAP}}(\pmb{\theta}_{anc}) ] &= \mathbf{A}_1 \pmb{\Sigma}_{anc}  \mathbf{A}^T_1 \\
&=  \mathbf{A}_1 \pmb{\Sigma}_{prior}  \mathbf{A}^T_1 \\
&= \pmb{\Sigma}_{post} \pmb{\Sigma}^{-1}_{prior} \pmb{\Sigma}_{prior} \pmb{\Sigma}^{-1}_{prior} \pmb{\Sigma}_{post}  \\
&= \pmb{\Sigma}_{post}  \pmb{\Sigma}^{-1}_{prior} \pmb{\Sigma}_{post} \label{eq_map_cov_anal}
\end{align}

\end{proof}

\begin{lemma}{}
\label{lem_anc_eq_prior_underpredicts_var}
Following from corollary \ref{cor_anc_eq_prior} (and under the same assumptions), when $\pmb{\mu}_{anc} \coloneqq \pmb{\mu}_{prior}$, $\pmb{\Sigma}_{anc} \coloneqq \pmb{\Sigma}_{prior}$, the RMS approximate posterior will in general underestimate the marginal variance compared to the true posterior, $\mathbb{V}ar[\pmb{f}_{\text{MAP}}({\theta}_{anc})] < \mathbb{V}ar[ {\theta} | \mathcal{D}]$.
\end{lemma}

\begin{proof}
We consider the marginal posterior of a single parameter, $\theta \coloneqq \pmb{\theta}_i$, again assuming multivariate normal prior and parameter likelihood.  First consider the following rearrangement of eq. \ref{eq_map_cov_anal}, beginning by noting, $\pmb{\Sigma}_{prior}^{-1} = \pmb{\Sigma}_{post}^{-1} -  \pmb{\Sigma}_{like}^{-1}$.
\begin{align}
 \pmb{\Sigma}_{post} \pmb{\Sigma}_{prior}^{-1} \pmb{\Sigma}_{post}
&=  \pmb{\Sigma}_{post} (\pmb{\Sigma}_{post}^{-1} -  \pmb{\Sigma}_{like}^{-1}) \pmb{\Sigma}_{post} \\
&= (\mathbb{I} -  \pmb{\Sigma}_{post}  \pmb{\Sigma}_{like}^{-1}) \pmb{\Sigma}_{post} \\
&= \pmb{\Sigma}_{post}  -  \pmb{\Sigma}_{post}  \pmb{\Sigma}_{like}^{-1} \pmb{\Sigma}_{post} 
\label{eq_anch_post_eq_rearrang}
\end{align}
To show that RMS generally underestimates the marginal variance, it must hold that diagonal elements of the true posterior covariance matrix are greater than or equal to the same diagonal element of the RMS posterior.
\begin{align}
\mathbb{V}ar[\pmb{f}_{\text{MAP}}({\theta}_{anc})] &< \mathbb{V}ar[ {\theta} | \mathcal{D}] \\
\text{diag}( \pmb{\Sigma}_{post} \pmb{\Sigma}_{prior}^{-1} \pmb{\Sigma}_{post} )_{i} &< \text{diag}( \pmb{\Sigma}_{post})_{i} \\
\intertext{substituting in the diagonal of the rearrangement in eq. \ref{eq_anch_post_eq_rearrang},}
\text{diag}( \pmb{\Sigma}_{post})_{i}  -  \text{diag}(  \pmb{\Sigma}_{post}  \pmb{\Sigma}_{like}^{-1} \pmb{\Sigma}_{post}  )_{i} &< \text{diag}( \pmb{\Sigma}_{post})_{i}
\end{align}
We know that $ABA^T$ is positive definite if $A, B$ are positive definite, and also that the inverse of a positive definite matrix is positive definite (\S9.6.4, \S9.6.10, The Matrix Cookbook, \citeyear{Pedersen2008}). The diagonal of a positive definite matrix is positive. Hence, $\text{diag}( \pmb{\Sigma}_{post}  \pmb{\Sigma}_{like}^{-1} \pmb{\Sigma}_{post} )_{i}  > 0$, and we have shown that $\mathbb{V}ar[\pmb{f}_{\text{MAP}}({\theta}_{anc})] < \mathbb{V}ar[ {\theta} | \mathcal{D}]$. 



\end{proof}

\begin{lemma}{}
\label{lem_anc_eq_prior_isotropic_orientation}
This lemma follows from corollary \ref{cor_anc_eq_prior}. Again parameter likelihood and prior are assumed normally distributed. The prior is additionally assumed isotropic. When $\pmb{\mu}_{anc} \coloneqq \pmb{\mu}_{prior}$, $\pmb{\Sigma}_{anc} \coloneqq \pmb{\Sigma}_{prior}$ the eigenvectors (or `orientation') of the RMS approximate posterior equal those of the true posterior.
\end{lemma}

\begin{proof}
From eq. \ref{eq_map_cov_anal}, $\mathbb{V}ar[\pmb{f}_{\text{MAP}}(\pmb{\theta}_{anc})] = \pmb{\Sigma}_{post}  \pmb{\Sigma}^{-1}_{prior} \pmb{\Sigma}_{post}$. If the prior is isotropic, $\pmb{\Sigma}_{prior} = \sigma^2_{prior}  \mathbb{I}$, then, $\mathbb{V}ar[\pmb{f}_{\text{MAP}}({\theta}_{anc})] = 1/ \sigma^2_{prior} \pmb{\Sigma}_{post}^2 $. Hence the prior only scales the eigenvalues, and doesn't affect the eigenvectors. (Note that this won't be the case for non-isotropic $\pmb{\Sigma}_{prior}$.)

Consider some matrix $A$ and a specific eigenvalue $\lambda_i$ and eigenvector $\mathbf{v}_i$ so that, $A \mathbf{v}_i=\lambda_i \mathbf{v}_i$. It then follows that if $A$ is squared, $A^2 \mathbf{v}_i=A(A\mathbf{v}_i)=\lambda_i A \mathbf{v}_i=\lambda_i^2 \mathbf{v}_i$. Hence eigenvalues are squared but eigenvectors are unaffected. This applies to the transformation $\pmb{\Sigma}_{post}^2$. 

Hence both the square and the multiplication of prior covariance, $1/ \sigma^2_{prior} \pmb{\Sigma}_{post}^2 $, do not modify the original eigenvectors of $\pmb{\Sigma}_{post}$ and its orientation is unaffected.

\end{proof}

\begin{theorem}
\label{theorem_corr_overestimate}
For a two parameter model with normally distributed parameter likelihood and isotropic prior, the RMS approximate posterior will in general overestimate the magnitude of the true posterior parameter correlation coefficient, $\lvert \rho \rvert$. However, if $\lvert \rho \rvert = 1$, then it will recover it precisely. We set $\pmb{\mu}_{anc} \coloneqq \pmb{\mu}_{prior}, \pmb{\Sigma}_{anc} \coloneqq \pmb{\Sigma}_{prior}$.
\end{theorem}{}

\begin{proof}
From corollary \ref{cor_anc_eq_prior}, we have that the RMS approximate posterior is given by $\pmb{\Sigma}_{post}\pmb{\Sigma}_{prior}^{-1}\pmb{\Sigma}_{post}$. Let $\pmb{\Sigma}_{prior} \coloneqq \sigma^2_{prior}\mathbb{I}$, the RMS approximate posterior is then given by $1/\sigma^2_{prior} \pmb{\Sigma}_{post}^2$.
Denote the true posterior covariance as the following 2$\times$2 matrix.
\begin{align}
\pmb{\Sigma}_{post} &= \begin{bmatrix}
a & b \\
b & c \\
\end{bmatrix} 
\end{align}
For general covariance matrices, the correlation coefficient, $\rho$, can be found by solving, $b = \rho \sqrt{a c}$. 

Our RMS approximate posterior is given as follows.
\begin{align}
1/\sigma^2_{prior}\pmb{\Sigma}_{post}^2 &= 
1/\sigma^2_{prior} 
\begin{bmatrix}
a & b \\
b & c \\
\end{bmatrix} 
\begin{bmatrix}
a & b \\
b & c \\
\end{bmatrix} \\
&= 1/\sigma^2_{prior} 
\begin{bmatrix}
a^2+b^2 & ab+bc \\
ab+bc & b^2+c^2 \\
\end{bmatrix} 
\end{align}
The correlation coefficient here, denoted $\rho_\text{\tiny{RMS}}$, is found by solving, $ab+bc = \rho_\text{\tiny{RMS}} \sqrt{(a^2+b^2) (b^2+c^2)}$. 

To prove the correlation coefficient is generally overestimated, we must show that $\rho_\text{\tiny{RMS}}^2 > \rho^2$ when $\rho^2<1$.
\begin{align}
\rho_\text{\tiny{RMS}}^2 &> \rho^2 \\
\frac{(ab+bc)^2}{(a^2+b^2) (b^2+c^2)}  &>  \frac{b^2}{ac}  \\
(ab+bc)^2 ac  &>  b^2 (a^2+b^2) (b^2+c^2)   \\
(a+c)^2 ac  &>  (a^2+b^2) (b^2+c^2)   \\
a^3c + ac^3 + 2a^2c^2  &>  a^2b^2 + a^2c^2 + b^4 + b^2c^2   \\
a^3c + ac^3 + a^2c^2  &>  a^2b^2 + b^4 + b^2c^2   \label{eq_2x2_last}
\end{align}
We now note that from the set up of the proof, $b^2 = \rho^2 a c$. Since $\rho^2<1$, we have, $b^2/ac < 1 \implies b^2<ac$. We can use this to provide an upper bound on the right hand side of eq. \ref{eq_2x2_last}.
\begin{align}
a^2b^2 + b^4 + b^2c^2 &< a^2 (ac) + (ac)^2 + (ac)c^2  \\
    &< a^3c + a^2c^2 +  ac^3
\end{align}
Coincidentally, this is precisely the inequality in eq. \ref{eq_2x2_last}, that we were proving. 

Alternatively, if $\lvert \rho \rvert = 1 \implies b^2=ac$, and $\rho_\text{\tiny{RMS}}^2 = \rho^2$.
\end{proof}

\newpage
\begin{definition}{Extrapolation Parameters}
\label{def_extrapolation parameters}

We define extrapolation parameters as model parameters which have no effect on the data likelihood of a training dataset, but which nevertheless could influence model predictions made on a new data point.

\textbf{Illustrative Example}

Consider a fully-connected NN trained on the MNIST digit dataset. Further consider a preprocessing such that the pixel values by default are set to $0$, and where they contain part of the digit take values $(0,1]$.

Certain pixels may be zero across the entire training dataset, such as those in the corners of the image. First layer weights connected to such pixels will never receive input across the whole training dataset. Hence, the values of these parameter weights have no effect on the data likelihood. However, the weights would still influence predictions for some test image containing values for these pixels. Hence, these are named extrapolation parameters, since they influence extrapolation properties of the model.

The top row of figure \ref{fig_gauss_like} empirically shows examples of flat likelihoods for precisely these types of weights on MNIST.

\end{definition}

\vspace{0.2in}

\begin{theorem}{}
\label{theorem_infinite_likelihood}
For extrapolation parameters (definition \ref{def_extrapolation parameters}) of a model, setting $\pmb{\mu}_{anc} \coloneqq \pmb{\mu}_{prior}$, $\pmb{\Sigma}_{anc} \coloneqq \pmb{\Sigma}_{prior}$, means the marginal RMS approximate posterior equals that of the marginal true posterior. This holds for any distributional form of parameter likelihood.
\end{theorem}

\begin{proof}
Extrapolation parameters (definition \ref{def_extrapolation parameters}) do not have any effect on the data likelihood, therefore their parameter likelihoods are flat. This means that their marginal posterior equals their marginal prior.

This results in a posterior covariance matrix structure as follows, where parameter $i$ is an extrapolation parameter (here shown in the first row for convenience), so has marginal variance equal to the prior variance and is uncorrelated with all other parameters.
\begin{equation*}
\pmb{\Sigma}_{post} 
   = \begin{bmatrix} 
    \sigma^2_{prior,i} & 0 & \dots & 0  \\
    0 &  a_{22} & \dots &  a_{2D}  \\
    \vdots & \vdots & \ddots & \vdots \\
    0 & a_{D2}   &  \dots    & a_{DD} 
    \end{bmatrix} \\
\end{equation*}
From corollary \ref{cor_anc_eq_prior}, $\mathbb{V}ar[\pmb{f}_{\text{MAP}}(\pmb{\theta}_{anc})] = \pmb{\Sigma}_{post}  \pmb{\Sigma}^{-1}_{prior} \pmb{\Sigma}_{post}$.
\begin{align*}
\mathbb{V}ar[\pmb{f}_{\text{MAP}}(\pmb{\theta}_{anc})]   &= \begin{bmatrix} 
    \sigma^2_{prior,i} & 0 & \dots & 0  \\
    0 &  a_{22} & \dots &  a_{2D}  \\
    \vdots & \vdots & \ddots & \vdots \\
    0 & a_{D2}   &  \dots    & a_{DD} 
    \end{bmatrix}
    \begin{bmatrix} 
    1 / \sigma^2_{prior,i} & 0 & \dots & 0  \\
    0 &  b_{22} & \dots &  0 \\
    \vdots & \vdots & \ddots & \vdots \\
    0 & 0  &  \dots    & b_{DD} 
    \end{bmatrix}
    \begin{bmatrix} 
    \sigma^2_{prior,i} & 0 & \dots & 0  \\
    0 &  a_{22} & \dots &  a_{2D}  \\
    \vdots & \vdots & \ddots & \vdots \\
    0 & a_{D2}   &  \dots    & a_{DD} 
    \end{bmatrix} \\
 &=\begin{bmatrix} 
    \sigma^2_{prior,i} & 0 & \dots & 0  \\
    0 &  c_{22} & \dots &  c_{2D}  \\
    \vdots & \vdots & \ddots & \vdots \\
    0 & c_{D2}   &  \dots    & c_{DD} 
    \end{bmatrix}
\end{align*}
This shows that the marginal variance of the RMS approximate posterior equals that of the true posterior, $\mathbb{V}ar[\pmb{f}_{\text{MAP}}(\pmb{\theta}_{anc})]_{i,i} = [\pmb{\Sigma}_{post}]_{i,i}$, for extrapolation parameters. Note that the values of the rest of the covariance matrices ($a$'s, $b$'s, $c$'s) are irrelevant since these have no effect on the marginals of interest.

This proof did not assume any specific distributional form of parameter likelihood, only that it is flat for these extrapolation parameters.

\end{proof}

\newpage
\begin{theorem}{}
\label{theorem_correlations}


Set $\pmb{\mu}_{anc} \coloneqq \pmb{\mu}_{prior}, \pmb{\Sigma}_{anc} \coloneqq \pmb{\Sigma}_{prior}$. The RMS approximate posterior will exactly equal the true posterior, $\pmb{\Sigma}_{post}$, when all eigenvalues of a scaled version of $\pmb{\Sigma}_{post}$ (scaled such that the prior equals the identity matrix) are equal to either $0$ or $1$. This corresponds to posteriors that are a mixture of perfectly correlated and perfectly uncorrelated parameters.

\end{theorem}


\begin{proof} 
We will initially consider a scaled version of the parameter space. This conveniently allows standard results for idempotent matrices to apply to the posterior covariance. A reverse scaling is subsequently applied to show that results hold for the original unscaled version. Finally, we articulate arguments allowing relaxation of the distributional assumptions.

Corollary \ref{cor_anc_eq_prior} showed that the RMS approximate posterior is normally distributed and centered at the true posterior mean, but with modified variance,
$\mathcal{N}(\pmb{\mu}_{post},\pmb{\Sigma}_{post} \pmb{\Sigma}_{prior}^{-1} \pmb{\Sigma}_{post})$. This proof requires specifying conditions that allow $\pmb{\Sigma}_{post} \pmb{\Sigma}_{prior}^{-1} \pmb{\Sigma}_{post} = \pmb{\Sigma}_{post}$ to hold.

One solution is given by $\pmb{\Sigma}_{post} = \pmb{\Sigma}_{prior}$, and is a trivial extension of theorem \ref{theorem_infinite_likelihood}. Here we consider alternative solutions.

The two inputs into the inference process are the likelihood and prior covariances. Consider a scaling $\pmb{\Sigma}_{like}' \coloneqq \pmb{\Sigma}_{prior}^{-1/2} \pmb{\Sigma}_{like} \pmb{\Sigma}_{prior}^{-1/2}$ and $\pmb{\Sigma}_{prior}' \coloneqq \pmb{\Sigma}_{prior}^{-1/2} \pmb{\Sigma}_{prior} \pmb{\Sigma}_{prior}^{-1/2} = \mathbb{I}$. The posterior for this scaled version will be denoted by $\pmb{\Sigma}_{post}'$, and is given as follows.
\begin{align}
    \pmb{\Sigma}_{post}' &= (\pmb{\Sigma}_{like}'^{-1} + \pmb{\Sigma}_{prior}'^{-1} )^{-1} \\
    &= (\pmb{\Sigma}_{prior}^{1/2}\pmb{\Sigma}_{like}^{-1}\pmb{\Sigma}_{prior}^{1/2} + \pmb{\Sigma}_{prior}^{1/2}\pmb{\Sigma}_{prior}^{-1}\pmb{\Sigma}_{prior}^{1/2} )^{-1} \\
    &= \pmb{\Sigma}_{prior}^{-1/2}(\pmb{\Sigma}_{like}^{-1} + \pmb{\Sigma}_{prior}^{-1} )^{-1} \pmb{\Sigma}_{prior}^{-1/2} \\
    &= \pmb{\Sigma}_{prior}^{-1/2} \pmb{\Sigma}_{post} \pmb{\Sigma}_{prior}^{-1/2} \label{eq_post_scaled}
\end{align}
Hence, unsurprisingly the same scaling applies to the posterior covariance, $\pmb{\Sigma}_{post}' = \pmb{\Sigma}_{prior}^{-1/2} \pmb{\Sigma}_{post} \pmb{\Sigma}_{prior}^{-1/2}$.

We now consider conditions under which $\pmb{\Sigma}_{post}' \pmb{\Sigma}_{prior}'^{-1} \pmb{\Sigma}_{post}' = \pmb{\Sigma}_{post}'$ holds. From our choice of rescaling, we have that $\pmb{\Sigma}_{prior}'^{-1} = \mathbb{I}$. So we require that $\pmb{\Sigma}_{post}'^2 = \pmb{\Sigma}_{post}'$. 

This conveniently allows use of results for idempotent matrices - defined as a square matrix, $A$, for which $A^2=A$. Aside from the case when $A = \mathbb{I}$ (which corresponds to $\pmb{\Sigma}_{post}=\pmb{\Sigma}_{prior}$), a matrix is idempotent if and only if it is singular and all eigenvalues are $0$ or $1$.
 
In order that, $\pmb{\Sigma}_{post}' \pmb{\Sigma}_{prior}'^{-1} \pmb{\Sigma}_{post}' = \pmb{\Sigma}_{post}'$, it is therefore sufficient that our scaled posterior, $\pmb{\Sigma}_{post}'$, is singular with all eigenvalues $0$ or $1$. Any possible permutation is allowed.

Naturally, applying a reverse scaling recovers the original parameter space, $\pmb{\Sigma}_{post} = \pmb{\Sigma}_{prior}^{1/2} \pmb{\Sigma}_{post}' \pmb{\Sigma}_{prior}^{1/2}$.

\begin{remark}
To summarise, we have shown that provided the RMS approximate posterior equals the true posterior in the scaled space, it will also be equal in the original unscaled space. In order for this equality to hold, eigenvalues must be $0$ or $1$ in the scaled space. 

See section \ref{sec_theorem_eg_idemp} for numerical examples in a three parameter model when this condition holds.


\end{remark}

\end{proof}




\newpage
\subsection{MAP solution and regularisation interpretation}
\label{app_map_solution_interp}
For completeness, we write out the MAP solution for the case of normally distributed prior, and data likelihoods often used in regression and classification. From this derives our interpretation of the regularisation matrix, $\pmb{\Gamma}$.
\begin{align*} 
\pmb{\theta}_{MAP} &= \text{argmax}_{\pmb{\theta}} P(\pmb{\theta} | \mathcal{D}) \\
 &= \text{argmax}_{\pmb{\theta}} P_{\mathcal{D}}( \mathcal{D} | \pmb{\theta} )  P(\pmb{\theta}) \\
 &  = \text{argmax}_{\pmb{\theta}} \log( P_{\mathcal{D}}( \mathcal{D} | \pmb{\theta} ) ) + \log(  P(\pmb{\theta})) 
 \intertext{If prior is normally distributed, $P(\pmb{\theta}) = \mathcal{N}(\pmb{\mu}, \pmb{\Sigma})$,} 
 & = \text{argmax}_{\pmb{\theta}} \log( P_{\mathcal{D}}( \mathcal{D} | \pmb{\theta} ) ) - 
 \frac{1}{2} 
 (\pmb{\theta} - \pmb{\mu})^T
 \pmb{\Sigma}^{-1}
 (\pmb{\theta} - \pmb{\mu})
 + \text{const.} \\
 & = \text{argmax}_{\pmb{\theta}} \log( P_{\mathcal{D}}( \mathcal{D} | \pmb{\theta} ) ) - 
 \frac{1}{2} 
 (\pmb{\theta} - \pmb{\mu})^T
 \pmb{\Sigma}^{-1}
 (\pmb{\theta} - \pmb{\mu}).
 \intertext{Typically in BNNs the prior covariance is chosen as diagonal. This is sometimes set as isotropic, $\pmb{\Sigma}= \lambda \mathbb{I}$, but here we will keep it in matrix form (but assuming it is diagonal) so that different prior variances can be assigned to different layer weights.}
  & = \text{argmax}_{\pmb{\theta}} \log( P_{\mathcal{D}}( \mathcal{D} | \pmb{\theta} ) ) - 
 \frac{1}{2} 
 \lVert \pmb{\Sigma}^{-1/2} \cdot
 (\pmb{\theta} - \pmb{\mu}) \rVert^2_2
 \intertext{In the case that the prior mean is zero $\pmb{\mu} = \mathbf{0}$,}
 & = \text{argmax}_{\pmb{\theta}} \log( P_{\mathcal{D}}( \mathcal{D} | \pmb{\theta} ) ) - 
 \frac{1}{2}
 \lVert \pmb{\Sigma}^{-1/2} \cdot
 \pmb{\theta} \rVert^2_2 .
\end{align*}
One is free to choose any suitable expression for $\log( P_{\mathcal{D}}( \mathcal{D} | \pmb{\theta} ) )$. Next we describe the resulting forms for common choices of log likelihood in regression and classification tasks.

\textbf{Regression}

For regression, a common choice is that the NN predicts the mean of the function, $\hat{\mathbf{y}}$, and there is additive noise on the true targets $\mathbf{y}$, $P_{\mathcal{D}}( \mathcal{D} | \pmb{\theta} ) = \mathcal{N}(\mathbf{y} | \hat{\mathbf{y}}, \sigma^2_{\epsilon})$
\begin{align*}
\pmb{\theta}_{MAP} & = \text{argmax}_{\pmb{\theta}}  
- \frac{1}{2 \sigma^2_{\epsilon}} \lVert \hat{\mathbf{y}} - \mathbf{y} \rVert^2_2 + \text{const.} - 
 \frac{1}{2} 
 \lVert  \pmb{\Sigma}^{-1/2} \cdot
 \pmb{\theta} \rVert^2_2 \\
   & = \text{argmax}_{\pmb{\theta}}  - \frac{1}{2 \sigma^2_{\epsilon}} \lVert \hat{\mathbf{y}} - \mathbf{y} \rVert^2_2 - 
 \frac{1}{2} 
 \lVert  \pmb{\Sigma}^{-1/2}\cdot
 \pmb{\theta} \rVert^2_2
 \intertext{Generally the mean squared error is minimised,}
    & = \text{argmin}_{\pmb{\theta}}   \frac{1}{N} \lVert \hat{\mathbf{y}} - \mathbf{y} \rVert^2_2 + 
 \frac{1}{N} 
 \lVert \sigma_{\epsilon}  \pmb{\Sigma}^{-1/2} \cdot
 \pmb{\theta} \rVert^2_2
  \intertext{More compactly, we can define $\pmb{\Gamma} \coloneqq \sigma^2_{\epsilon}  \pmb{\Sigma}^{-1}$, as a diagonal matrix with, $\text{diag}(\pmb{\Gamma})_{i} = \sigma^2_{\epsilon} / \sigma^2_{prior, i}$,}
      & = \text{argmin}_{\pmb{\theta}}   \frac{1}{N} \lVert \hat{\mathbf{y}} - \mathbf{y} \rVert^2_2 + 
 \frac{1}{N} 
 \lVert  \pmb{\Gamma}^{1/2} \cdot
 \pmb{\theta} \rVert^2_2
\end{align*}

\textbf{Classification}

The data likelihood is commonly chosen as a multinomial distribution, $P_{\mathcal{D}}( \mathcal{D} | \pmb{\theta} ) \propto \prod_{n=1}^N  \prod_{c=1}^C  \hat{y}_{n,c}^{y_{n,c}}$, for $C$ classes, and $N$ data points, where $\hat{y} \in [0,1]$ denotes predicted probability, and $y_{n,c} \in \{0,1\}$ the true targets.
\begin{align*}
\pmb{\theta}_{MAP} & = \text{argmax}_{\pmb{\theta}}  
 \sum_{n=1}^N  \sum_{c=1}^C y_{n,c} \log(\hat{y}_{n,c}) 
 - \frac{1}{2} 
 \lVert  \pmb{\Sigma}^{-1/2} \cdot
 \pmb{\theta} \rVert^2_2 \\
 \intertext{Cross entropy is typically minimised,}
 & = \text{argmin}_{\pmb{\theta}}  
- \sum_{n=1}^N  \sum_{c=1}^C y_{n,c} \log(\hat{y}_{n,c}) 
 + \frac{1}{2} 
 \lVert  \pmb{\Sigma}^{-1/2} \cdot
 \pmb{\theta} \rVert^2_2 \\
\end{align*}
Here we can simply define $\pmb{\Gamma} \coloneqq \frac{1}{2} \pmb{\Sigma}^{-1}$, with $\text{diag}(\pmb{\Gamma})_{i} = 1/ 2 \sigma^2_{prior, i}$.

\newpage
\section{Numerical Examples}
\subsection{Numerical Examples of Proofs}
\label{app_numerical_egs_of_proofs}
In this section we print covariance matrices that illustrate theoretical results numerically.

\subsubsection{General case: Example of lemma \ref{lem_anc_eq_prior_underpredicts_var}, \ref{lem_anc_eq_prior_isotropic_orientation}, theorem \ref{theorem_corr_overestimate}}
For general $\pmb{\Sigma}_{post}$, RMS will return a posterior with underestimated marginal variancesa and overestimated correlations. Since the prior is isotropic, the orientation (eigenvectors) of the RMS approximate posterior will be unchanged. Here a three parameter is shown. Note $\pmb{\Sigma}_{post}\pmb{\Sigma}_{prior}^{-1}\pmb{\Sigma}_{post} $ represents the RMS approximate posterior.
\begin{align*}
 \pmb{\Sigma}_{prior} &= \begin{bmatrix}
2.0 & 0.0 & 0.0 \\
0.0 & 2.0 & 0.0 \\
0.0 & 0.0 & 2.0
\end{bmatrix} 
 &\pmb{\Sigma}_{like} = \begin{bmatrix}
2.0 & 0.707 & 0.283 \\
0.707 & 1.0 & 0.4 \\
0.283 & 0.4 & 1.0
\end{bmatrix} \\
 \pmb{\Sigma}_{post} &= \begin{bmatrix}
0.953 & 0.238 & 0.067 \\
0.238 & 0.589 & 0.166 \\
0.067 & 0.166 & 0.638
\end{bmatrix} 
&\pmb{\Sigma}_{post}\pmb{\Sigma}_{prior}^{-1}\pmb{\Sigma}_{post} = \begin{bmatrix}
0.485 & 0.189 & 0.073 \\
0.189 & 0.215 & 0.11 \\
0.073 & 0.11 & 0.22
\end{bmatrix} \\
 \intertext{Note, ${\pmb{\Sigma}_{post}}_{i,i} > {\pmb{\Sigma}_{post}\pmb{\Sigma}_{prior}^{-1}\pmb{\Sigma}_{post}}_{i,i}, \forall i$. } 
 \text{Correlation}(\pmb{\Sigma}_{post}) &= \begin{bmatrix}
1.0 & 0.317 & 0.086 \\
0.317 & 1.0 & 0.27 \\
0.086 & 0.27 & 1.0
\end{bmatrix} 
 &\text{Correlation}(\pmb{\Sigma}_{post}\pmb{\Sigma}_{prior}^{-1}\pmb{\Sigma}_{post}) = \begin{bmatrix}
1.0 & 0.585 & 0.224 \\
0.585 & 1.0 & 0.504 \\
0.224 & 0.504 & 1.0
\end{bmatrix} 
 \intertext{Note, ${ \text{Correlation}(\pmb{\Sigma}_{post})}_{i,j} < {\text{Correlation}(\pmb{\Sigma}_{post}\pmb{\Sigma}_{prior}^{-1}\pmb{\Sigma}_{post})}_{i,j}, \forall i \neq j$. } 
\end{align*}
\begin{align*}
\intertext{Eigenvalues and eigenvectors of $\pmb{\Sigma}_{post}$,}
\lambda &= 1.1101 , \mathbf{v} = \begin{bmatrix}-0.8352 &-0.471 & -0.284  \end{bmatrix} \\
\lambda &= 0.6667 , \mathbf{v} = \begin{bmatrix}-0.465  & 0.3288 & 0.822  \end{bmatrix} \\
\lambda &= 0.4032 , \mathbf{v} = \begin{bmatrix} 0.2938 &-0.8186 & 0.4936 \end{bmatrix} \\
\intertext{Eigenvalues and eigenvectors of $\pmb{\Sigma}_{post}\pmb{\Sigma}_{prior}^{-1}\pmb{\Sigma}_{post}$,}
\lambda &= 0.6162 , \mathbf{v} = \begin{bmatrix}-0.8352 &-0.471 & -0.284  \end{bmatrix} \\
\lambda &= 0.2222 , \mathbf{v} = \begin{bmatrix}-0.465  & 0.3288 & 0.822  \end{bmatrix} \\
\lambda &= 0.0813 , \mathbf{v} = \begin{bmatrix} 0.2938 &-0.8186 & 0.4936 \end{bmatrix} \\
\end{align*}


\subsubsection{Special case: Examples of theorem \ref{theorem_infinite_likelihood}, \ref{theorem_correlations}}
\label{sec_theorem_eg_idemp}
We again print out covariance matrices for a three parameter model. Firstly we provide an example where two parameters are perfectly correlated, and one has no effect on the likelihood. We print unscaled and scaled versions. Note that all eigenvalues of the scaled posterior, $\pmb{\Sigma}_{post}'$, are either $0$ or $1$.
\begin{align*}
 \pmb{\Sigma}_{post} &= \begin{bmatrix}
1.0 & 1.0 & 0.0 \\
1.0 & 1.0 & 0.0 \\
0.0 & 0.0 & 2.0
\end{bmatrix} 
\;\;\;\; \pmb{\Sigma}_{prior} = \begin{bmatrix}
2.0 & 0.0 & 0.0 \\
0.0 & 2.0 & 0.0 \\
0.0 & 0.0 & 2.0
\end{bmatrix} 
\;\;\;\; \pmb{\Sigma}_{post}\pmb{\Sigma}_{prior}^{-1}\pmb{\Sigma}_{post} = \begin{bmatrix}
1.0 & 1.0 & 0.0 \\
1.0 & 1.0 & 0.0 \\
0.0 & 0.0 & 2.0
\end{bmatrix}
\end{align*}
\begin{align*}
 \pmb{\Sigma}_{post}' &= \begin{bmatrix}
0.5 & 0.5 & 0.0 \\
0.5 & 0.5 & 0.0 \\
0.0 & 0.0 & 1.0
\end{bmatrix} 
\;\;\;\; \pmb{\Sigma}_{prior}' = \begin{bmatrix}
1.0 & 0.0 & 0.0 \\
0.0 & 1.0 & 0.0 \\
0.0 & 0.0 & 1.0
\end{bmatrix} 
\;\;\;\; \pmb{\Sigma}_{post}'\pmb{\Sigma}_{prior}'^{-1}\pmb{\Sigma}_{post}' = \begin{bmatrix}
0.5 & 0.5 & 0.0 \\
0.5 & 0.5 & 0.0 \\
0.0 & 0.0 & 1.0
\end{bmatrix}
\end{align*}
\begin{align*}
\intertext{Eigenvalues and eigenvectors of $\pmb{\Sigma}_{post} = \pmb{\Sigma}_{post}\pmb{\Sigma}_{prior}^{-1}\pmb{\Sigma}_{post}$,}
\lambda &= 2.0 , \mathbf{v} = \begin{bmatrix} 0.7071 & 0.7071 &  0. \end{bmatrix} \\
\lambda &= 0.0 , \mathbf{v} = \begin{bmatrix} -0.7071 & 0.7071&   0. \end{bmatrix} \\
\lambda &= 2.0 , \mathbf{v} = \begin{bmatrix} 0. & 0. & 1.\end{bmatrix}
\end{align*}
\begin{align*}
\intertext{Eigenvalues and eigenvectors of $\pmb{\Sigma}_{post}' = \pmb{\Sigma}_{post}'\pmb{\Sigma}_{prior}'^{-1}\pmb{\Sigma}_{post}'$,}
\lambda &= 1.0 , \mathbf{v} = \begin{bmatrix} 0.707 & 0.707 &  0.    \end{bmatrix} \\
\lambda &= 0.0 , \mathbf{v} = \begin{bmatrix} -0.707 &  0.707  & 0.    \end{bmatrix} \\
\lambda &= 1.0 , \mathbf{v} = \begin{bmatrix} 0.&  0. & 1. \end{bmatrix} \
\end{align*}

Following is the same set up as the previous case, but now all parameters are perfectly correlated. Eigenvalues of the scaled posterior, $\pmb{\Sigma}_{post}'$, are again either $0$ or $1$.
\begin{align*}
\pmb{\Sigma}_{post} = \begin{bmatrix}
0.667 & 0.667 & 0.667 \\
0.667 & 0.667 & 0.667 \\
0.667 & 0.667 & 0.667
\end{bmatrix} 
 \;\;\;\; \pmb{\Sigma}_{prior} = \begin{bmatrix}
2.0 & 0.0 & 0.0 \\
0.0 & 2.0 & 0.0 \\
0.0 & 0.0 & 2.0
\end{bmatrix} 
\;\;\;\;  \pmb{\Sigma}_{post}\pmb{\Sigma}_{prior}^{-1}\pmb{\Sigma}_{post}  = \begin{bmatrix}
0.667 & 0.667 & 0.667 \\
0.667 & 0.667 & 0.667 \\
0.667 & 0.667 & 0.667
\end{bmatrix} 
 \end{align*}
 \begin{align*}
 \pmb{\Sigma}_{post}' &= \begin{bmatrix}
0.333 & 0.333 & 0.333 \\
0.333 & 0.333 & 0.333 \\
0.333 & 0.333 & 0.333
\end{bmatrix} 
\;\;\;\; \pmb{\Sigma}_{prior}' = \begin{bmatrix}
1.0 & 0.0 & 0.0 \\
0.0 & 1.0 & 0.0 \\
0.0 & 0.0 & 1.0
\end{bmatrix} 
\;\;\;\; \pmb{\Sigma}_{post}'\pmb{\Sigma}_{prior}'^{-1}\pmb{\Sigma}_{post}' = \begin{bmatrix}
0.333 & 0.333 & 0.333 \\
0.333 & 0.333 & 0.333 \\
0.333 & 0.333 & 0.333
\end{bmatrix}
\end{align*}
\begin{align*}
\intertext{Eigenvalues and eigenvectors of $\pmb{\Sigma}_{post} = \pmb{\Sigma}_{post}\pmb{\Sigma}_{prior}^{-1}\pmb{\Sigma}_{post}$,}
\lambda &= 2.0 , \mathbf{v} = \begin{bmatrix} -0.5774 & -0.5774 & -0.5774 \end{bmatrix}\\
\lambda &= 0.0 , \mathbf{v} = \begin{bmatrix} -0.     & -0.7071 &  0.7071 \end{bmatrix}\\
\lambda &= 0.0 , \mathbf{v} = \begin{bmatrix} -0.6667 & -0.0749 &  0.7416 \end{bmatrix}
\end{align*}
\begin{align*}
\intertext{Eigenvalues and eigenvectors of $\pmb{\Sigma}_{post}' = \pmb{\Sigma}_{post}'\pmb{\Sigma}_{prior}'^{-1}\pmb{\Sigma}_{post}'$,}
\lambda &= 1.0 , \mathbf{v} = \begin{bmatrix} 0.577 & 0.577 & 0.577 \end{bmatrix} \\
\lambda &= 0.0 , \mathbf{v} = \begin{bmatrix}  0.    & -0.707  & 0.707 \end{bmatrix} \\
\lambda &= 0.0 , \mathbf{v} = \begin{bmatrix} -0.521 & -0.284  & 0.805 \end{bmatrix} \\ 
\end{align*}

Now we detail an example of a non-isometric prior.
\begin{align*}
\pmb{\Sigma}_{post} = \begin{bmatrix}
1.818 & 0.0 & 1.818 \\
0.0 & 2.0 & 0.0 \\
1.818 & 0.0 & 1.818
\end{bmatrix} 
 \;\;\;\; \pmb{\Sigma}_{prior} = \begin{bmatrix}
20.0 & 0.0 & 0.0 \\
0.0 & 2.0 & 0.0 \\
0.0 & 0.0 & 2.0
\end{bmatrix} 
\;\;\;\;  \pmb{\Sigma}_{post}\pmb{\Sigma}_{prior}^{-1}\pmb{\Sigma}_{post}  = \begin{bmatrix}
1.818 & 0.0 & 1.818 \\
0.0 & 2.0 & 0.0 \\
1.818 & 0.0 & 1.818
\end{bmatrix} 
 \end{align*}
 \begin{align*}
 \pmb{\Sigma}_{post}' &= \begin{bmatrix}
0.091 & 0.0 & 0.287 \\
0.0 & 1.0 & 0.0 \\
0.287 & 0.0 & 0.909
\end{bmatrix} 
\;\;\;\; \pmb{\Sigma}_{prior}' = \begin{bmatrix}
1.0 & 0.0 & 0.0 \\
0.0 & 1.0 & 0.0 \\
0.0 & 0.0 & 1.0
\end{bmatrix} 
\;\;\;\; \pmb{\Sigma}_{post}'\pmb{\Sigma}_{prior}'^{-1}\pmb{\Sigma}_{post}' = \begin{bmatrix}
0.091 & 0.0 & 0.287 \\
0.0 & 1.0 & 0.0 \\
0.287 & 0.0 & 0.909
\end{bmatrix}
\end{align*}
\begin{align*}
\intertext{Eigenvalues and eigenvectors of $\pmb{\Sigma}_{post} = \pmb{\Sigma}_{post}\pmb{\Sigma}_{prior}^{-1}\pmb{\Sigma}_{post}$,}
\lambda &= 3.636 , \mathbf{v} = \begin{bmatrix} 0.707 & 0.  &   0.707 \end{bmatrix} \\
\lambda &= 0.0 , \mathbf{v} = \begin{bmatrix} -0.707 &  0.  &    0.707 \end{bmatrix} \\
\lambda &= 2.0 , \mathbf{v} = \begin{bmatrix} 0.&  1.&  0. \end{bmatrix}
\end{align*}
\begin{align*}
\intertext{Eigenvalues and eigenvectors of $\pmb{\Sigma}_{post}' = \pmb{\Sigma}_{post}'\pmb{\Sigma}_{prior}'^{-1}\pmb{\Sigma}_{post}'$,}
\lambda &= 0.0 , \mathbf{v} = \begin{bmatrix} -0.953 & 0.  &  0.302 \end{bmatrix} \\
\lambda &= 1.0 , \mathbf{v} = \begin{bmatrix} -0.302 & 0.  &  -0.953 \end{bmatrix} \\
\lambda &= 1.0 , \mathbf{v} = \begin{bmatrix} 0. &1.& 0. \end{bmatrix} \\
\end{align*}

\subsection{Mixtures of Parameter Types}
Here, we provide examples of a five parameter model containing a mixture of perfectly correlated, partially correlated, and extrapolation parameters. 

First we consider distinct blocks of perfectly and partially correlated parameters, as well as one extrapolation parameter. In this situation both the perfectly correlated block and the extrapolation parameter posterior are recovered exactly. The RMS approximate posterior of the partially correlated block is biased as per the general case.
 \begin{align*}
\pmb{\Sigma}_{post} = \begin{bmatrix}
1.0 & 1.0 & 0.0 & 0.0 & 0.0 \\
1.0 & 1.0 & 0.0 & 0.0 & 0.0 \\
0.0 & 0.0 & 0.5 & 0.2 & 0.0 \\
0.0 & 0.0 & 0.2 & 0.8 & 0.0 \\
0.0 & 0.0 & 0.0 & 0.0 & 2.0
\end{bmatrix} 
 \; \pmb{\Sigma}_{prior} = \begin{bmatrix}
2 & 0 & 0 & 0 & 0 \\
0 & 2 & 0 & 0 & 0 \\
0 & 0 & 2 & 0 & 0 \\
0 & 0 & 0 & 2 & 0 \\
0 & 0 & 0 & 0 & 2
\end{bmatrix} 
\;  \pmb{\Sigma}_{post}\pmb{\Sigma}_{prior}^{-1}\pmb{\Sigma}_{post}  = \begin{bmatrix}
1.0 & 1.0 & 0.0 & 0.0 & 0.0 \\
1.0 & 1.0 & 0.0 & 0.0 & 0.0 \\
0.0 & 0.0 & 0.145 & 0.13 & 0.0 \\
0.0 & 0.0 & 0.13 & 0.34 & 0.0 \\
0.0 & 0.0 & 0.0 & 0.0 & 2.0
\end{bmatrix} 
 \end{align*} 
 
Secondly the perfectly correlated block overlaps with the partially correlated block. In this scenario, a small amount of bias is introduced on the perfectly correlated block, but not in terms of the correlation. The extrapolation parameter is unaffected.
\begin{align*}
\pmb{\Sigma}_{post} = \begin{bmatrix}
1.0 & 1.0 & 0.1 & 0.2 & 0.0 \\
1.0 & 1.0 & 0.1 & 0.2 & 0.0 \\
0.1 & 0.1 & 0.5 & 0.2 & 0.0 \\
0.2 & 0.2 & 0.2 & 0.8 & 0.0 \\
0.0 & 0.0 & 0.0 & 0.0 & 2.0
\end{bmatrix} 
 \; \pmb{\Sigma}_{prior} = \begin{bmatrix}
2 & 0 & 0 & 0 & 0 \\
0 & 2 & 0 & 0 & 0 \\
0 & 0 & 2 & 0 & 0 \\
0 & 0 & 0 & 2 & 0 \\
0 & 0 & 0 & 0 & 2
\end{bmatrix} 
\;  \pmb{\Sigma}_{post}\pmb{\Sigma}_{prior}^{-1}\pmb{\Sigma}_{post}  = \begin{bmatrix}
1.03 & 1.03 & 0.15 & 0.29 & 0.0 \\
1.03 & 1.03 & 0.15 & 0.29 & 0.0 \\
0.15 & 0.15 & 0.16 & 0.15 & 0.0 \\
0.29 & 0.29 & 0.15 & 0.38 & 0.0 \\
0.0 & 0.0 & 0.0 & 0.0 & 2.0
\end{bmatrix} 
 \end{align*}

\newpage
\subsection{Example of Perfectly Correlated Parameters in a Neural Network}
\label{app_numerical}
\label{app_like_joint}

Section \ref{sec_presence_special_case} stated that perfectly correlated parameters can exist in a NN. Here, we provide a concrete example of such a case, and show that anchored ensembling \textit{does} recover the true posterior for these parameters when $\pmb{\Sigma}_{anc} = \pmb{\Sigma}_{prior}$.

We consider finding the posterior of final layer weights in a small single layer ReLU NN of two hidden nodes for a regression problem with two data points. (Choosing the final layer weights for our analysis allows analytical equations associated with linear regression to be used, simplifying our analysis, though these results would also apply to the first layer weights and biases.)

We design the problem such that the point where both hidden nodes becomes greater than zero (the elbow points of ReLU units) falls in between the two data points, and the active half of the output is also shared, so that the final layer weights are perfectly correlated.

\begin{figure}[h!]
\vskip 0.1in
\begin{center}
    \begin{minipage}{0.3\textwidth}
        \begin{overpic}[width=1.\textwidth]{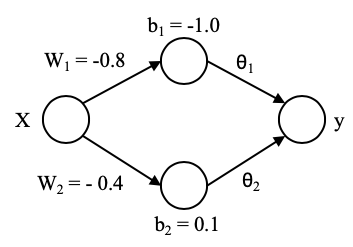}
        \end{overpic}
    \end{minipage}
    \hskip 0.1in
    \begin{minipage}{0.3\textwidth}
        \begin{overpic}[width=1.\textwidth]{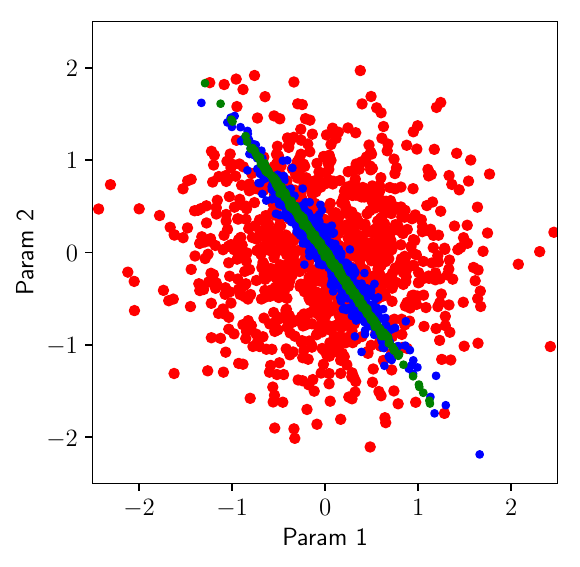}
        \end{overpic}
    \end{minipage}
    \hskip 0.2in
    \begin{minipage}{0.3\textwidth}
        \begin{overpic}[width=1.\textwidth]{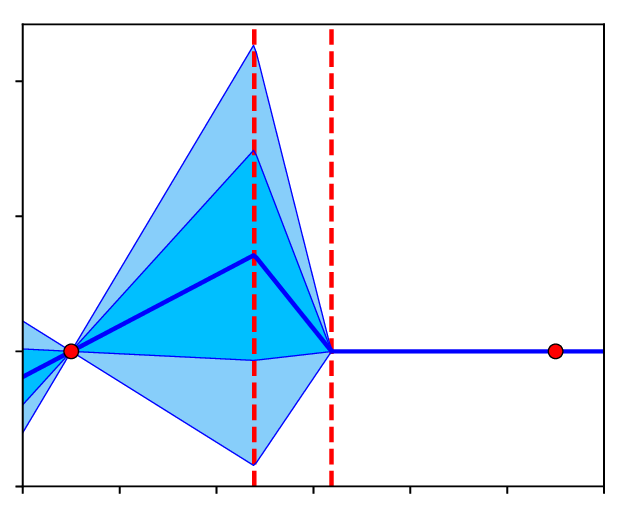}
        \end{overpic}
    \end{minipage}
\caption{Left: Single layer NN of two hidden nodes. Middle: Draws in parameter space for prior (red), analytical posterior (blue) and anchored posterior (green). Right: Posterior predictive distribution - dashed red lines are elbows of ReLU units. }
\vskip -0.2in
\label{fig_numeric_NN_diag}
\end{center}
\end{figure}

Figure \ref{fig_numeric_NN_diag} illustrates our set up. Data points and NN parameters are as follows, 

$\mathcal{D} = \{x_1 = -5, y_1 = 0; x_2 = 5, y_2=0 \}$, $\sigma^2_{\epsilon}=0.01$,

$ \mathbf{W}_1  = [-0.8 ,-0.4],  \mathbf{b}_1 = [-1,  0.1], \hat{y} = \mathbf{W}_2^T  \text{max}(x \mathbf{W}_1 + \mathbf{b}_1,0) $

We set prior means to zero, with isotropic covariance according to $1/H$,
\[
\pmb{\Sigma}_{prior} = 
\begin{bmatrix}
0.5  &  0.0 \\
0.0 &  0.5
\end{bmatrix}
\]

We print out matrices of interest,
\[
\pmb{\Sigma}_{like} = 
\begin{bmatrix}
-6.89e+13  & 9.85e+13 \\
 9.85e+13 &  -1.40e+14
\end{bmatrix}
\] 
\[
\pmb{\Sigma}_{like}^{-1} = 
\begin{bmatrix}
 90.0  &63.0  \\
 63.0 &44.1
\end{bmatrix}
\]
\[
\pmb{\Sigma}_{post} = 
\begin{bmatrix}
  0.169& -0.231\\
 -0.231 & 0.338
\end{bmatrix}
\]
\[
\pmb{\Sigma}_{post} \pmb{\Sigma}_{prior}^{-1} \pmb{\Sigma}_{post} = 
\begin{bmatrix}
  0.166 &-0.235\\
 -0.235  &0.338
\end{bmatrix}
\]
The anchored ensembling posterior, $\pmb{\Sigma}_{post} \pmb{\Sigma}_{prior}^{-1} \pmb{\Sigma}_{post}$, provides a close approximation of the true posterior covariance (and would be exact discounting numerical rounding issues). Note the similarity of the posterior in figure \ref{fig_numeric_NN_diag} (middle) with the perfect correlations example shown in figure \ref{fig_RMS_general_special_egs} (B).

\onecolumn
\section{Further Results}
\label{sec_app_results}

\subsection{Uncertainty-Aware Model-Free Reinforcement Learning}

\begin{wrapfigure}{R}{0.48\textwidth}
\vskip -0.5in
\begin{center}
    \begin{minipage}{0.22\textwidth}
        \centering
        \begin{overpic}[width=1.\textwidth]{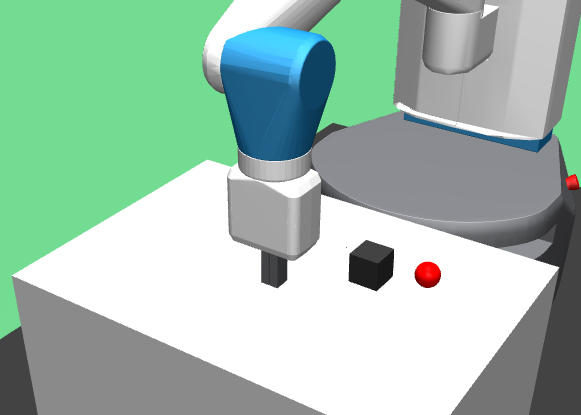}
        \put (6,4) {\scriptsize \textbf{A. Clear optimal action}}
        \end{overpic}
    \end{minipage}
    \hskip -0.08in
    \begin{minipage}{0.26\textwidth}
        \begin{overpic}[width=1.\textwidth, height=0.7\textwidth]{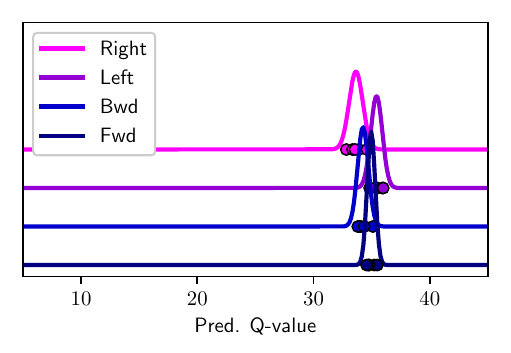}
        \end{overpic}
    \end{minipage}
    \vskip -0.14in
    \begin{minipage}{0.22\textwidth}
        \centering
        \begin{overpic}[width=1.\textwidth]{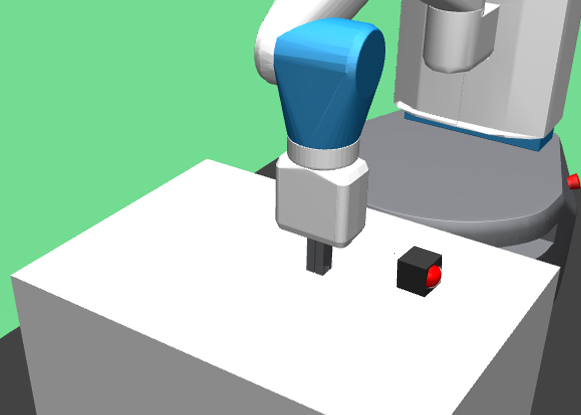}  
        \put (6,4) {\scriptsize \textbf{B. Goal achieved}}
        \end{overpic}
    \end{minipage}
    \hskip -0.08in
    \begin{minipage}{0.26\textwidth}
        \centering
        \begin{overpic}[width=1.\textwidth, height=0.7\textwidth]{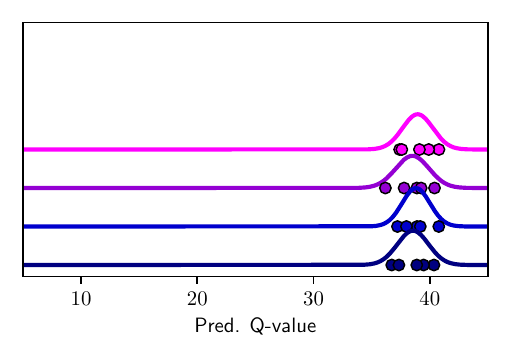}  
        \end{overpic}
    \end{minipage}
    \vskip -0.14in
    \begin{minipage}{0.22\textwidth}
        \centering
        \begin{overpic}[width=0.97\textwidth]{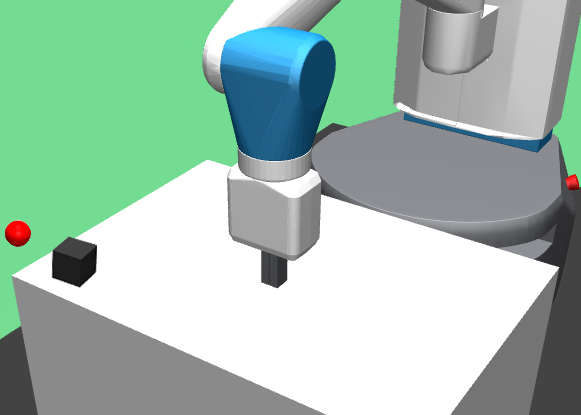}
        \put (6,4) {\scriptsize \textbf{C. Never seen before}}
        \end{overpic}
    \end{minipage}
    \hskip -0.05in
    \begin{minipage}{0.26\textwidth}
        \centering
        \begin{overpic}[width=1.\textwidth, height=0.7\textwidth]{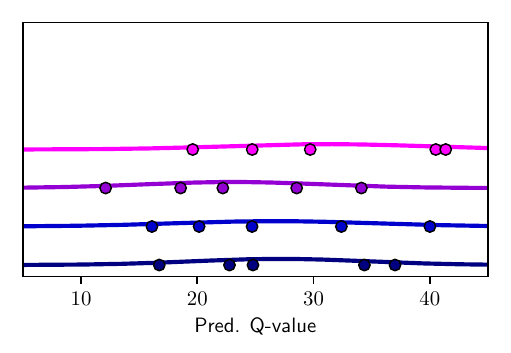}
        \end{overpic}
     \end{minipage}
\caption{Anchored ensembling creates uncertainty-aware agents.}
\vskip -0.5in
\label{fig_RL}
\end{center}
\end{wrapfigure}

An anchored ensemble of 5xNNs, each with two hidden layers, was trained to complete a discretised version of FetchPush - an agent controls a robotic arm, with rewards received when a randomly placed cube is pushed to a goal. 

We used Bayesian Q-learning \citep{Dearden1998}, similar to regular Q-learning, but with Q-values modelled as \textit{distributions} rather than point estimates - the wider the distribution, the less certain the agent. This is beneficial both to drive the exploration/exploitation process via Thompson sampling, and for identifying OOD examples.

Figure \ref{fig_RL} shows the agent's awareness of its uncertainty. After training for $40,000$ episodes, its confidence over actions was plotted for three scenarios: A) Cube and goal are in positions often encountered during training, the agent has learnt that it must move the arm left - the narrow distributions with significantly different means reflect its confidence in this. B) The goal has already been achieved - narrow overlapping distributions with higher means. C) A peculiar goal position that has never been encountered - the broad distributions over all actions reflect its high uncertainty.


\vspace{0.1in}
\subsection{Model-Based Reinforcement: Learning in Noisy Environments}

We tested the benefit of using an anchored ensemble in noisy RL environments. We modified the classic cartpole swingup environment such that different levels of stochastic noise could be added to the future state; consider a state action pair given by, $s, a$, and a noisy state, $\bar{s}$ such that $P(\bar{s}_{t+1} | a_t, s_t) = \mathcal{N}(s_{t+1}, \sigma^2_{\epsilon})$. 

The value of $\sigma^2_{\epsilon}$ was given three settings: low, medium and high, $ \sigma^2_{\epsilon} \in \{0.001, 0.002,0.005\}$. The task was learnt using a model-based RL approach similar to the heteroskedastic ensembles used by \cite{Chua2018a}. Fully-connected three-layer NNs learnt to predict the dynamics of the environment given some state and action. Planning was performed using the cross-entropy method, rolling out for a horizon of 25 steps, with 10 particles.

\begin{figure*}[b!]
\vskip -0.05in
\begin{center}
    \begin{minipage}{0.25\textwidth}
        \centering
        \begin{overpic}[width=1.\textwidth, height=0.6\textwidth]{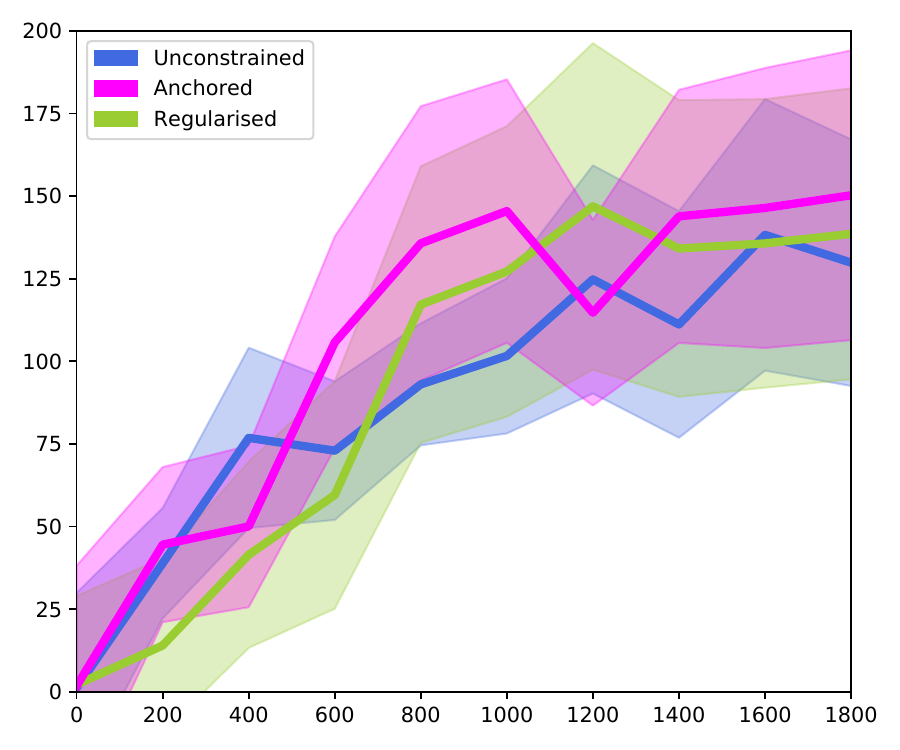}
        \put (20,8) {\small {A. Low Noise}}
        \put(-5,16){\rotatebox{90}{\small Reward}}
         \put (35,-5) {\small Time steps}
        \end{overpic}
    \end{minipage}
    \hskip -0.08in
     \begin{minipage}{0.25\textwidth}
        \centering
        \begin{overpic}[width=1.\textwidth, height=0.6\textwidth]{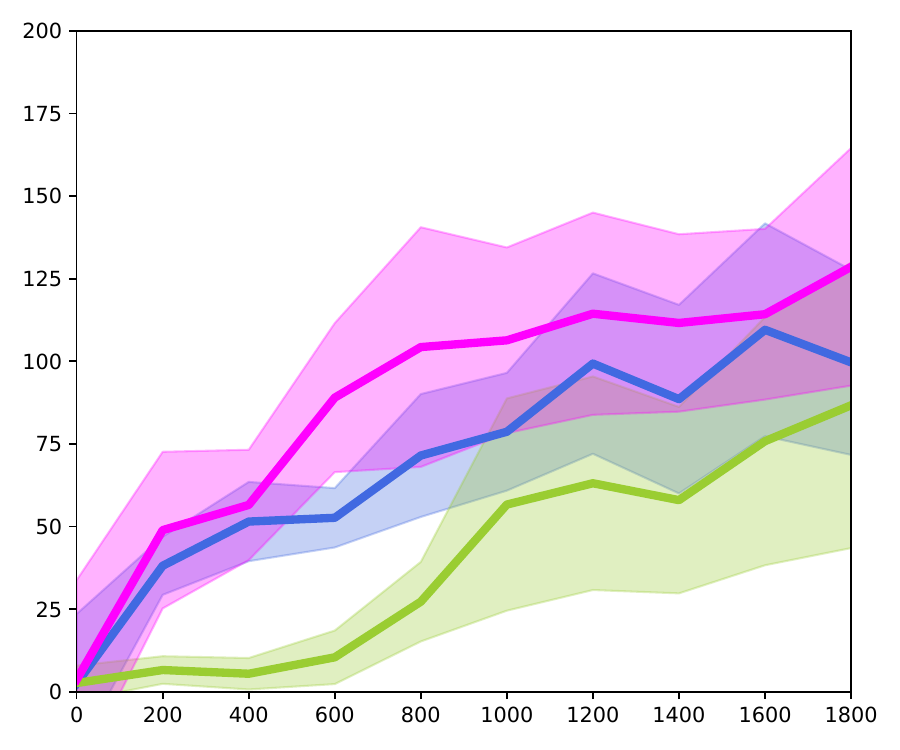}
        \put (20,8) {\small {B. Medium Noise}}
        \end{overpic}
    \end{minipage}
    \hskip -0.08in
     \begin{minipage}{0.25\textwidth}
        \centering
        \begin{overpic}[width=1.\textwidth, height=0.6\textwidth]{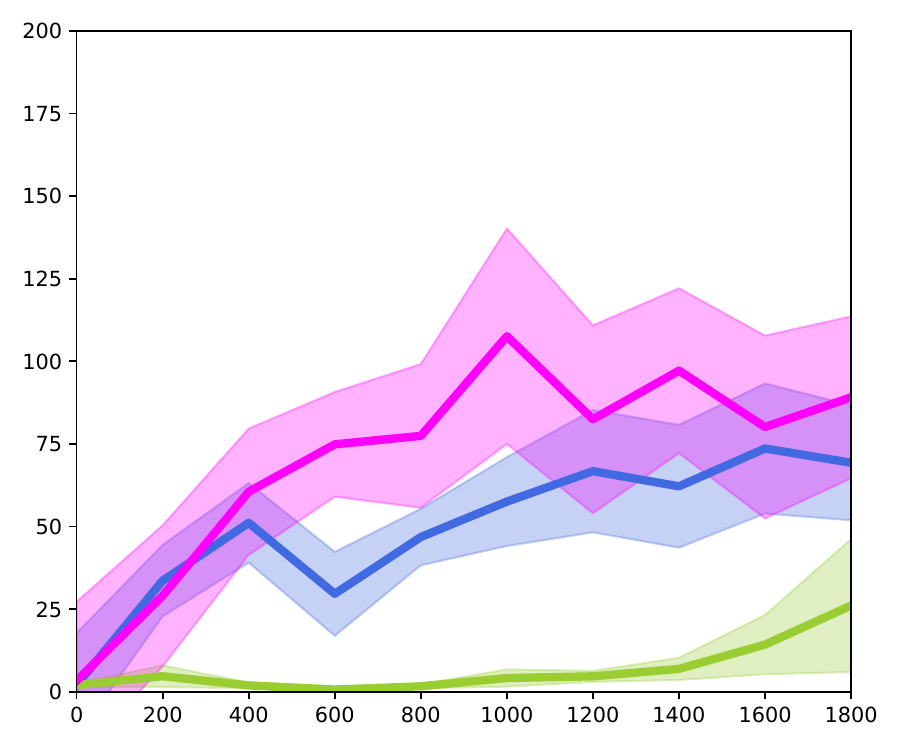}
        \put (20,8) {\small {C. High Noise}}
        \end{overpic}
    \end{minipage}
    \hskip -0.08in
     \begin{minipage}{0.25\textwidth}
        \centering
        \begin{overpic}[width=1.\textwidth, height=0.6\textwidth]{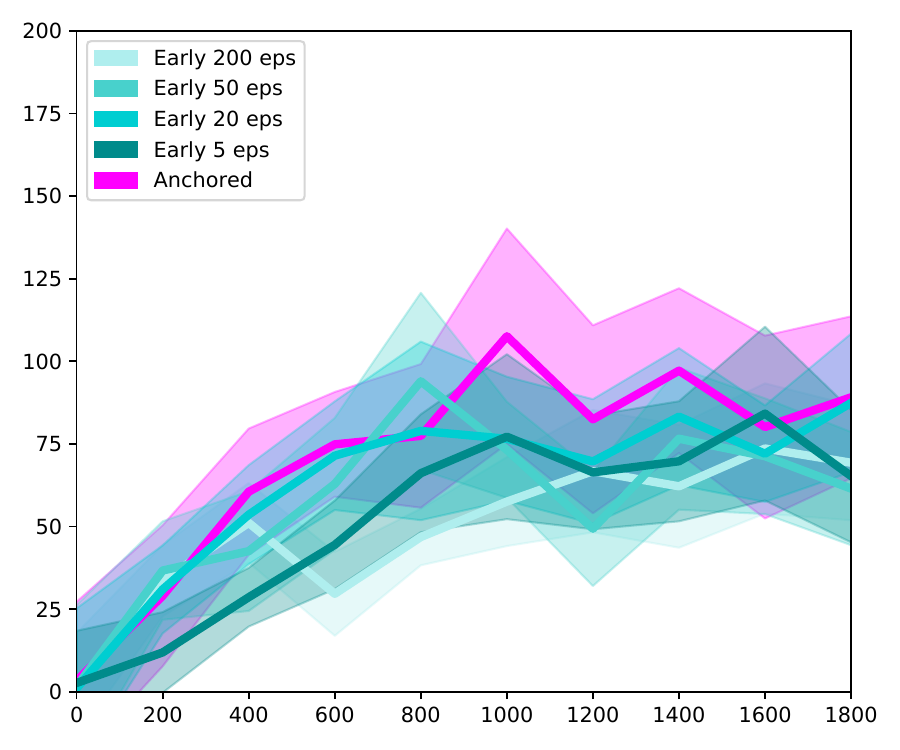}
        \put (20,8) {\small {D. Early Stopping}}
        \end{overpic}
    \end{minipage}
 \caption{Anchored ensembling creates robust MBRL agents in noisy environments. Mean and standard error over five runs.}
\vskip -0.2in
\label{fig_model_based_RL}
\end{center}
\end{figure*}

Figure \ref{fig_model_based_RL} (A, B, C) shows learning curves for the three noise levels. All ensemble techniques perform similarly in the low noise setting. As noise is increased the overall performance of all methods drops. Anchored ensembles are most resiliant, followed by the unconstrained ensemble. In panel D, we compared against an unconstrained ensemble employing early stopping - this corresponds to the proposal that applying early stopping to an ensemble can produce approximate inference \citep{Duvenaud2016}. Whilst careful tuning did offer some improvement over the default setting, a performance gap to anchored ensembling remained.

\newpage
\subsection{Regression Benchmarking}
\label{app_regression_bench}
Tables \ref{tab_rmse} \& \ref{tab_nll} show all experiments run on the regression benchmarking datasets. The below discussion focuses on NLL results in table \ref{tab_nll}.

ERF GP refers to the equivalent GP for an infinite width, single-layer BNN with ERF activations. It was tuned and implemented as for the ReLU GP. We were interested to discover how different activation functions would affect uncertainty estimates. In general the ReLU GP performed better than the ERF GP, with some exceptions, such as for Wine. The target variable for Wine is ordinal, containing five factors, it is therefore understandable that the ReLU GP, which extrapolates linearly, is at a slight disadvantage.

10x 50 NNs refers to an anchored ensemble of ten NNs with 50 hidden nodes. We find that these results fall in between the 5x 50 NNs and the ReLU GP. This agrees with the convergence analysis done in section \ref{sec_result_vis_converg}. 

We also implemented an anchored ensemble of five two-layer NNs, 5x 50-50 NNs. Even with minimal hyperparameter tuning (section \ref{sec_exp_deets}) we found an extra layer gave a performance boost over the 5x 50 NNs. We expect with more careful tuning this margin would increase.

Single 50 NN refers to a single regularised NN, of one hidden layer with 50 hidden nodes, for which we used a constant value of predictive variance. Although this performs poorly in several cases, e.g. Boston and Yacht, the results are surprisingly close to those achieved by both our method and Deep Ensembles, even surpassing them on the Energy dataset. A method outputting constant predictive variance should not perform well in experiments designed to test uncertainty quantification, and this raises questions over the validity of the benchmarks.

Table \ref{tab_uci_comparison} compares anchored ensembles against results reported for other methods.

\begin{table}[h]%
\caption{Variants of our method on benchmark regression datasets, RMSE.}
\vskip 0.15in
\begin{center}
\resizebox{0.8\columnwidth}{!}{
\begin{tabular}{ l r r r r    r r r r }

\Xhline{3\arrayrulewidth}

\multicolumn{3}{c}{}  & \multicolumn{5}{c}{\textbf{RMSE}}  \\

\multicolumn{1}{c}{} 
& \multicolumn{1}{r}{$N$}
& \multicolumn{1}{r}{$D$}
& \multicolumn{1}{c}{ReLU GP}
&  \multicolumn{1}{c}{ERF GP} 
&  \multicolumn{1}{c}{5x 50 NNs} 
&  \multicolumn{1}{c}{10x 50 NNs} 
&  \multicolumn{1}{c}{5x 50-50 NNs} 
&  \multicolumn{1}{c}{Single 50 NN}  \\ 
 
\hline 
&&&&& \\

Boston & 506 & 13 & 2.86 $\pm$ 0.16 & 2.94 $\pm$ 0.18 & 3.09 $\pm$ 0.17 & 3.09 $\pm$ 0.17 & 3.00 $\pm$ 0.18 & 3.40 $\pm$ 0.20 \\
Concrete & 1,030 & 8 & 4.88 $\pm$ 0.13 & 5.21 $\pm$ 0.12 & 4.87 $\pm$ 0.11 & 4.73 $\pm$ 0.11 & 4.75 $\pm$ 0.12 & 5.17 $\pm$ 0.13 \\
Energy & 768 & 8 & 0.60 $\pm$ 0.02 & 0.78 $\pm$ 0.03 & 0.35 $\pm$ 0.01 & 0.34 $\pm$ 0.01 & 0.40 $\pm$ 0.01 & 0.40 $\pm$ 0.01 \\
Kin8nm & 8,192 & 8 & 0.07 $\pm$ 0.00 & 0.08 $\pm$ 0.00 & 0.07 $\pm$ 0.00 & 0.07 $\pm$ 0.00 & 0.06 $\pm$ 0.00 & 0.07 $\pm$ 0.00 \\
Naval & 11,934 & 16 & 0.00 $\pm$ 0.00 & 0.00 $\pm$ 0.00 & 0.00 $\pm$ 0.00 & 0.00 $\pm$ 0.00 & 0.00 $\pm$ 0.00 & 0.00 $\pm$ 0.00 \\
Power & 9,568 & 4 & 3.97 $\pm$ 0.04 & 3.94 $\pm$ 0.04 & 4.07 $\pm$ 0.04 & 4.07 $\pm$ 0.04 & 4.03 $\pm$ 0.04 & 4.23 $\pm$ 0.04 \\
Protein & 45,730 & 9 & 4.34 $\pm$ 0.02 & 4.23 $\pm$ 0.02 & 4.36 $\pm$ 0.02 & 4.34 $\pm$ 0.02 & 4.23 $\pm$ 0.02 & 4.56 $\pm$ 0.02 \\
Wine & 1,599 & 11 & 0.61 $\pm$ 0.01 & 0.60 $\pm$ 0.01 & 0.63 $\pm$ 0.01 & 0.62 $\pm$ 0.01 & 0.62 $\pm$ 0.01 & 0.64 $\pm$ 0.01 \\
Yacht & 308 & 6 & 0.60 $\pm$ 0.08 & 1.48 $\pm$ 0.15 & 0.57 $\pm$ 0.05 & 0.54 $\pm$ 0.05 & 0.85 $\pm$ 0.08 & 0.81 $\pm$ 0.07 \\
Song Year & 515,345 & 90 & 9.01 $\pm$ NA & 8.90 $\pm$ NA & 8.82 $\pm$ NA & 8.82 $\pm$ NA & 8.66 $\pm$ NA & 8.77 $\pm$ NA \\

\Xhline{3\arrayrulewidth} 

\end{tabular}
}
\end{center}
\vskip -0.1in
\label{tab_rmse}
\end{table}

\begin{table}[h]%
\caption{Variants of our method on benchmark regression datasets, NLL.}
\vskip 0.15in
\begin{center}
\resizebox{0.8\columnwidth}{!}{
\begin{tabular}{ l r r r    r r r r }

\Xhline{3\arrayrulewidth}

\multicolumn{2}{c}{}  & \multicolumn{5}{c}{\textbf{NLL}}  \\

\multicolumn{1}{c}{} 
& \multicolumn{1}{r}{$\tphat{\sigma^2_\epsilon$}}  
& \multicolumn{1}{c}{ReLU GP}
&  \multicolumn{1}{c}{ERF GP} 
&  \multicolumn{1}{c}{5x 50 NNs} 
&  \multicolumn{1}{c}{10x 50 NNs} 
&  \multicolumn{1}{c}{5x 50-50 NNs} 
&  \multicolumn{1}{c}{Single 50 NN}  \\ 
 
\hline 
&&&&& \\

Boston & 0.08 & 2.45 $\pm$ 0.05 & 2.46 $\pm$ 0.05 & 2.52 $\pm$ 0.05 & 2.50 $\pm$ 0.05 & 2.50 $\pm$ 0.07 & 2.70 $\pm$ 0.05 \\
Concrete & 0.05 & 2.96 $\pm$ 0.02 & 3.06 $\pm$ 0.02 & 2.97 $\pm$ 0.02 & 2.94 $\pm$ 0.02 & 2.94 $\pm$ 0.02 & 3.08 $\pm$ 0.03 \\
Energy & 1e-7 & 0.86 $\pm$ 0.02 & 1.06 $\pm$ 0.03 & 0.96 $\pm$ 0.13 & 0.52 $\pm$ 0.06 & 0.61 $\pm$ 0.07 & 0.57 $\pm$ 0.03 \\
Kin8nm & 0.02 & -1.22 $\pm$ 0.01 & -1.17 $\pm$ 0.00 & -1.09 $\pm$ 0.01 & -1.16 $\pm$ 0.01 & -1.25 $\pm$ 0.01 & -1.17 $\pm$ 0.01 \\
Naval & 1e-7 & -10.05 $\pm$ 0.02 & -9.66 $\pm$ 0.04 & -7.17 $\pm$ 0.03 & -7.29 $\pm$ 0.02 & -7.08 $\pm$ 0.13 & -6.58 $\pm$ 0.04 \\
Power & 0.05 & 2.80 $\pm$ 0.01 & 2.79 $\pm$ 0.01 & 2.83 $\pm$ 0.01 & 2.83 $\pm$ 0.01 & 2.82 $\pm$ 0.01 & 2.86 $\pm$ 0.01 \\
Protein & 0.5 & 2.88 $\pm$ 0.00 & 2.86 $\pm$ 0.00 & 2.89 $\pm$ 0.01 & 2.88 $\pm$ 0.01 & 2.86 $\pm$ 0.01 & 2.94 $\pm$ 0.00 \\
Wine & 0.5 & 0.92 $\pm$ 0.01 & 0.91 $\pm$ 0.01 & 0.95 $\pm$ 0.01 & 0.94 $\pm$ 0.01 & 0.94 $\pm$ 0.01 & 0.97 $\pm$ 0.01 \\
Yacht & 1e-7 & 0.49 $\pm$ 0.07 & 1.50 $\pm$ 0.13 & 0.37 $\pm$ 0.08 & 0.18 $\pm$ 0.03 & 0.04 $\pm$ 0.08 & 1.50 $\pm$ 0.02 \\
Song Year & 0.7 & 3.62 $\pm$ NA & 3.61 $\pm$ NA & 3.60 $\pm$ NA & 3.60 $\pm$ NA & 3.57 $\pm$ NA & 3.59 $\pm$ NA \\

\Xhline{3\arrayrulewidth} 

\end{tabular}
}
\end{center}
\vskip -0.1in
\label{tab_nll}
\end{table}

\begin{table}[h]%
\caption{Comparison against inference methods on UCI benchmark regression datasets, log likelihood. Adapted from \cite{Mukhoti2018}.}
\vskip 0.15in
\begin{center}
\resizebox{0.8\columnwidth}{!}{
\begin{tabular}{ l r r r r   r r r r}

\Xhline{3\arrayrulewidth}

\multicolumn{2}{c}{}  & \multicolumn{5}{c}{\textbf{Log Likelihood (\textit{not} negative)}}  \\

\multicolumn{1}{c}{} 
& \multicolumn{1}{r}{Anch. Ens.}  
& \multicolumn{1}{c}{Drop conv.}
&  \multicolumn{1}{c}{Drop tune} 
&  \multicolumn{1}{c}{VMG} 
&  \multicolumn{1}{c}{HS-BNN} 
&  \multicolumn{1}{c}{PBP-MV} 
&  \multicolumn{1}{c}{SGHMC tune}  
&  \multicolumn{1}{c}{SGHMC adap.}  \\ 

\hline 
&&&&& \\

    Boston & $-2.52 \pm 0.05$ & $\mathbf{-2.40 \pm 0.04}$ & $\mathbf{-2.40 \pm 0.04}$ & $-2.46 \pm 0.09$ & $-2.54 \pm 0.15$ & $-2.54 \pm 0.08$ & $-2.49 \pm 0.15$ & $-2.54 \pm 0.04$ \\
    Concrete & $-2.97 \pm 0.02$ & $-2.97 \pm 0.02$ & $\mathbf{-2.93 \pm 0.02}$ & $-3.01 \pm 0.03$ & $-3.09 \pm 0.06$ & $-3.04 \pm 0.03$ & $-4.17 \pm 0.72$ & $-3.38 \pm 0.24$ \\
    Energy & $\textcolor{blue}{\mathbf{-0.96 \pm 0.13}}$ & $-1.72 \pm 0.01$ & $-1.21 \pm 0.01$ & $-1.06 \pm 0.03$ & $-2.66 \pm 0.13$ & ${-1.01 \pm 0.01}$ & $--$ & $--$ \\
    Kin8nm & $1.09 \pm 0.01$ & $0.97 \pm 0.00$ & $1.14 \pm 0.01$ & $1.10 \pm 0.01$ & $1.12 \pm 0.03$ & $\mathbf{1.28 \pm 0.01}$ & $--$ & $--$ \\
    Naval & $\textcolor{blue}{\mathbf{7.17 \pm 0.03}}$ & $3.91 \pm 0.01$ & $4.45 \pm 0.00$ & $2.46 \pm 0.00$ & ${5.52 \pm 0.10}$ & $4.85 \pm 0.06$ & $--$ & $--$ \\
    Power & $-2.83 \pm 0.01$ & $-2.79 \pm 0.01$ & $-2.80 \pm 0.01$ & $-2.82 \pm 0.01$ & $-2.81 \pm 0.03$ & $\mathbf{-2.78 \pm 0.01}$ & $--$ & $--$ \\
    Protein & $-2.89 \pm 0.01$ & $-2.87 \pm 0.00$ & $-2.87 \pm 0.00$ & $-2.84 \pm 0.00$ & $-2.89 \pm 0.00$ & $\mathbf{-2.77 \pm 0.01}$ & $--$ & $--$ \\
    Wine & $-0.95 \pm 0.01$ & $\mathbf{-0.92 \pm 0.01}$ & $-0.93 \pm 0.01$ & $-0.95 \pm 0.01$ & $-0.95 \pm 0.05$ & $-0.97 \pm 0.01$ & $-1.29 \pm 0.28$ & $-1.04 \pm 0.17$ \\
    Yacht & $\textcolor{blue}{\mathbf{-0.37 \pm 0.08}}$ & $-1.38 \pm 0.01$ & $-1.25 \pm 0.01$ & $-1.30 \pm 0.02$ & $-2.33 \pm 0.01$ & $-1.64 \pm 0.02$ & $-1.75 \pm 0.19$ & $\mathbf{-1.10 \pm 0.08}$ \\

\Xhline{3\arrayrulewidth} 

\end{tabular}
}
\end{center}
\vskip -0.1in
\label{tab_uci_comparison}
\end{table}

\newpage
\subsection{Fashion MNIST}
\label{app_fashion}

Table \ref{tab_fashion_full} provides a breakdown of results from the OOD classification test in section \ref{sec_results_img_class} for fashion MNIST. Also included are results for entropy, where high entropy represents high uncertainty. These correlated strongly with the proportion metrics, which was true across all three OOD experiments.

\begin{table*}[h]%
\caption{ Fashion MNIST results: proportion of predictions made with $\geq$ 90\% probability, and entropy of predicted categorical distribution. Also shown is relative advantage (percentage change) for each method compared to anchored ensembles. Averaged over five runs/random seeds, mean $\pm$ 1 standard error. Best result in \textcolor{blue}{blue}.}
\begin{center}
\resizebox{1.\columnwidth}{!}{
\begin{tabular}{ l | r | rrrrrrrrr }

\Xhline{3\arrayrulewidth}

\multicolumn{1}{c}{}  & \multicolumn{1}{c}{}  &  \multicolumn{2}{c}{-------Edge Cases-------} &  \multicolumn{2}{c}{---Out-of-distribution---} & \multicolumn{3}{c}{------------Natural Adversarial------------} &  \multicolumn{2}{c}{---Pure Adversarial---}  \\ 

\multicolumn{1}{c}{}  & \multicolumn{1}{c}{Train}  &  \multicolumn{1}{c}{Sneaker} &  \multicolumn{1}{c}{Trouser} &  \multicolumn{1}{c}{CIFAR} &  \multicolumn{1}{c}{MNIST} &  \multicolumn{1}{c}{Rotate} &  \multicolumn{1}{c}{Flip} &  \multicolumn{1}{c}{Invert} &  \multicolumn{1}{c}{Noise} &  \multicolumn{1}{c}{Sparse}  \\ 

\hline
\multicolumn{2}{c}{}  & \multicolumn{8}{c}{Proportion $\geq$ 90\% (smaller better)} \\

reg 1xNN & { 0.660 $\pm$ 0.006} & { 0.739 $\pm$ 0.056} & { 0.429 $\pm$ 0.047} & { 0.143 $\pm$ 0.008} & { 0.160 $\pm$ 0.007} & { 0.609 $\pm$ 0.007} & { 0.330 $\pm$ 0.009} & { 0.349 $\pm$ 0.015} & { 0.271 $\pm$ 0.007} & { 0.456 $\pm$ 0.006} \\
free 5xNN & { 0.733 $\pm$ 0.001} & { 0.781 $\pm$ 0.015} & { 0.380 $\pm$ 0.030} & { 0.301 $\pm$ 0.013} & { 0.104 $\pm$ 0.010} & { 0.571 $\pm$ 0.011} & { 0.300 $\pm$ 0.011} & { 0.222 $\pm$ 0.052} & { 0.042 $\pm$ 0.005} & { 0.048 $\pm$ 0.003} \\
reg 5xNN & { 0.634 $\pm$ 0.002} & { 0.589 $\pm$ 0.054} & \textcolor{blue}{ 0.269 $\pm$ 0.020} & { 0.115 $\pm$ 0.004} & { 0.072 $\pm$ 0.007} & { 0.556 $\pm$ 0.007} & { 0.256 $\pm$ 0.012} & { 0.213 $\pm$ 0.002} & { 0.112 $\pm$ 0.005} & { 0.174 $\pm$ 0.005} \\
anc 5xNN & { 0.631 $\pm$ 0.002} & \textcolor{blue}{ 0.578 $\pm$ 0.049} & { 0.325 $\pm$ 0.037} & \textcolor{blue}{ 0.065 $\pm$ 0.002} & \textcolor{blue}{ 0.041 $\pm$ 0.002} & \textcolor{blue}{ 0.497 $\pm$ 0.003} & \textcolor{blue}{ 0.215 $\pm$ 0.005} & \textcolor{blue}{ 0.025 $\pm$ 0.010} & \textcolor{blue}{ 0.006 $\pm$ 0.001} & \textcolor{blue}{ 0.006 $\pm$ 0.001} \\

\hline
\multicolumn{2}{c}{}  & \multicolumn{8}{c}{Proportion Relative Advantage} \\

1xNN Reg. to 5xNN Anch. & { -4.4\%} & { -21.8\%} & { -24.2\%} & { -54.5\%} & { -74.4\%} & { -18.4\%} & { -34.8\%} & { -92.8\%} & { -97.8\%} & { -98.7\%} \\
5xNN Uncons. to 5xNN Anch. & { -13.9\%} & { -26.0\%} & { -14.5\%} & { -78.4\%} & { -60.6\%} & { -13.0\%} & { -28.3\%} & { -88.7\%} & { -85.7\%} & { -87.5\%} \\
5xNN Reg. to 5xNN Anch. & { -0.5\%} & { -1.9\%} & { 20.8\%} & { -43.5\%} & { -43.1\%} & { -10.6\%} & { -16.0\%} & { -88.3\%} & { -94.6\%} & { -96.6\%} \\


\hline
\multicolumn{2}{c}{}  & \multicolumn{8}{c}{Entropy (larger better)} \\
1xNN Reg. & { 0.328 $\pm$ 0.005} & { 0.253 $\pm$ 0.043} & { 0.575 $\pm$ 0.050} & { 1.176 $\pm$ 0.010} & { 0.984 $\pm$ 0.015} & { 0.484 $\pm$ 0.008} & { 0.713 $\pm$ 0.009} & { 0.836 $\pm$ 0.035} & { 0.808 $\pm$ 0.010} & { 0.580 $\pm$ 0.008} \\
5xNN Uncons. & { 0.230 $\pm$ 0.001} & { 0.161 $\pm$ 0.010} & { 0.535 $\pm$ 0.021} & { 0.688 $\pm$ 0.009} & { 1.016 $\pm$ 0.021} & { 0.453 $\pm$ 0.011} & { 0.685 $\pm$ 0.011} & { 0.573 $\pm$ 0.037} & { 1.036 $\pm$ 0.014} & { 0.992 $\pm$ 0.012} \\
5xNN Reg. & { 0.352 $\pm$ 0.001} & \textcolor{blue}{ 0.365 $\pm$ 0.039} & \textcolor{blue}{ 0.707 $\pm$ 0.019} & { 1.239 $\pm$ 0.009} & { 1.161 $\pm$ 0.012} & { 0.564 $\pm$ 0.008} & { 0.807 $\pm$ 0.017} & { 1.014 $\pm$ 0.014} & { 1.048 $\pm$ 0.008} & { 0.919 $\pm$ 0.009} \\
5xNN Anch. & { 0.349 $\pm$ 0.001} & { 0.327 $\pm$ 0.034} & { 0.623 $\pm$ 0.042} & \textcolor{blue}{ 1.251 $\pm$ 0.011} & \textcolor{blue}{ 1.295 $\pm$ 0.013} & \textcolor{blue}{ 0.624 $\pm$ 0.006} & \textcolor{blue}{ 0.868 $\pm$ 0.002} & \textcolor{blue}{ 1.098 $\pm$ 0.035} & \textcolor{blue}{ 1.238 $\pm$ 0.013} & \textcolor{blue}{ 1.191 $\pm$ 0.014} \\

\hline
\multicolumn{2}{c}{}  & \multicolumn{8}{c}{Entropy Relative Advantage} \\
1xNN Reg. to 5xNN Anch. & { 6.4\%} & { 29.2\%} & { 8.3\%} & { 6.4\%} & { 31.6\%} & { 28.9\%} & { 21.7\%} & { 31.3\%} & { 53.2\%} & { 105.3\%} \\
5xNN Uncons. to 5xNN Anch. & { 51.7\%} & { 103.1\%} & { 16.4\%} & { 81.8\%} & { 27.5\%} & { 37.7\%} & { 26.7\%} & { 91.6\%} & { 19.5\%} & { 20.1\%} \\
5xNN Reg. to 5xNN Anch. & { -0.9\%} & { -10.4\%} & { -11.9\%} & { 1.0\%} & { 11.5\%} & { 10.6\%} & { 7.6\%} & { 8.3\%} & { 18.1\%} & { 29.6\%} \\

\Xhline{3\arrayrulewidth}

\end{tabular}
}
\end{center}

\vskip -0.1in
\label{tab_fashion_full}
\end{table*}

\FloatBarrier
\onecolumn
\section{Additional Material}
\label{sec_exp_result}




\subsection{Algorithms}

\begin{algorithm}[h]
   \caption{Implementing anchored ensembles of NNs}
   \label{alg_RL_ens}
\begin{algorithmic}

\STATE {\bfseries Input:} Training data, $\mathbf{X}$ \& $\mathbf{Y}$, test data point, $\mathbf{x}^*$, prior mean and covariance, $\pmb{\mu}_{prior}$, $\pmb{\Sigma}_{prior}$, ensemble size, $M$, data noise variance estimate, $\hat{\sigma}^2_{\epsilon}$ (regression only).

\STATE {\bfseries Output:} Estimate of mean and variance, $\hat{\mathbf{y}}$, $\hat{\sigma}^2_{y}$ for regression, or class probabilities, $\hat{\mathbf{y}}$ for classification.

\vskip 0.08in
\STATE  \textit{\# Set regularisation matrix}
\STATE $\pmb{\Gamma} \Leftarrow \hat{\sigma}^2_{\epsilon} \pmb{\Sigma}_{prior}^{-1}$  (regression) \hspace{0.05in} OR \hspace{0.05in} $\pmb{\Gamma} \Leftarrow \frac{1}{2}\pmb{\Sigma}_{prior}^{-1}$ (classification)

\vskip 0.08in
\STATE  \textit{\# Create ensemble}
\STATE $\pmb{\mu}_{anc} \Leftarrow \pmb{\mu}_{prior}, \pmb{\Sigma}_{anc} \Leftarrow \pmb{\Sigma}_{prior}$
\STATE \textbf{for} $j = 1$ \textbf{to} $M$
\begin{ALC@g}
	\STATE $\pmb{\theta}_{anc,j} \sim \mathcal{N}(\pmb{\mu}_{anc},\pmb{\Sigma}_{anc})$ \textit{\# Sample anchor points}
	\STATE $NN_j.\text{create}(\pmb{\Gamma}, \pmb{\theta}_{anc, j})$ \textit{\# Create custom regulariser}
	\STATE $NN_j.\text{initialise}()$ \textit{\# Initialisations independent of $ \pmb{\theta}_{anc,j}$}
\end{ALC@g}

\vskip 0.08in
\STATE  \textit{\# Train ensemble}
\STATE \textbf{for} $j = 1$ \textbf{to} $M$
\begin{ALC@g}
	\STATE $NN_j.\text{train}(\mathbf{X}, \mathbf{Y})$, loss in eq. \ref{eqn_anch_loss_matrix} (regression) or eq. \ref{eqn_anch_loss_matrix_class} (classification) or eq. \ref{eq_MAP_loglike_anc} (custom)
\end{ALC@g}

\vskip 0.08in
\STATE  \textit{\# Predict with ensemble}
\STATE \textbf{for} $j = 1$ \textbf{to} $M$
\begin{ALC@g}
	\STATE $\hat{\mathbf{y}}_j \Leftarrow NN_j.\text{predict}(\mathbf{x}^*)$ 
\end{ALC@g}

\vskip 0.08in
\STATE  \textit{\# Regression - combine ensemble estimates}
\STATE $\hat{\mathbf{y}} = \frac{1}{M} \sum_{j=1}^{M}  \hat{\mathbf{y}}_j,$ \textit{\# Mean prediction}
\STATE $\hat{\sigma}^2_{model} = \frac{1}{M-1} \sum_{j=1}^{M} ( \tphat{y_{j}} - \tphat{y} )^2$ \textit{\# Epistemic var.}
\STATE $\hat{\sigma}^2_{y} = \hat{\sigma}^2_{model} + \hat{\sigma}^2_{\epsilon}$ \textit{\# Total var. = epistemic + data noise}

\vskip 0.08in
\STATE  \textit{\# Classification - combine ensemble estimates}
\STATE $\hat{\mathbf{y}} = \frac{1}{M} \sum_{j=1}^{M}  \hat{\mathbf{y}}_j,$ \textit{\# Average softmax output}
\STATE $\hat{\sigma}^2_{y} =$ None \textit{\# N/A for classification}

\vskip 0.08in
\STATE \textbf{return} $\hat{\mathbf{y}}$, $\hat{\sigma}^2_{y}$
\end{algorithmic}
\end{algorithm}

%
%
%
%
%
%
%
%
%
%
%
%
%
%
%
%

\newpage
\FloatBarrier
\onecolumn
\section{Experimental Details}
\label{sec_exp_deets}

\subsection{Introduction to Anchored Ensembles}

Experimental details for figure \ref{fig_anch_action} are as follows.

Six randomly generated data points were used.

Hyperparameters: activation = ERF, $\sigma_{\epsilon}^2$ = 0.003, $b_1$ variance = 1, $W_1$ variance  = 1, $H$ = 100, $M$ = 3 (number of ensembles), optimiser = adam, epochs = 400, learning rate = 0.005.

\subsection{Panel of Inference Methods}
\label{sec_exp_panel}

Experimental details for figure \ref{fig_methods} are as follows.

Same six data points were used for all methods and activation functions, generated by $y=x \sin(5x)$, evaluated at, [-0.8, -0.1, 0.02, 0.2, 0.6, 0.8].

Hyperparameters: $b_1$ variance = 10, $W_1$ variance  = 10, $H$ = 100, $M$ = 10, epochs= 4,000, $\sigma_{\epsilon}^2$ = 0.001, leaky ReLU $\alpha$ = 0.2, optimiser = adam, MC Dropout probability = 0.4, MC Dropout samples = 200, HMC step size = 0.001, HMC no. steps = 150, HMC burn in = 500, HMC total samples = 1000, HMC predict samples = 50, VI predict samples = 50, VI iterations = 2000, VI gradient samples = 200.


\subsection{Ensembling Loss Functions} 

Experimental details for figure \ref{fig_reg_or_not} are as follows.

\subsubsection{Regression}
Generated $\mathbf{X}$ by sampling 20 points linearly spaced from the interval [-1.5, 1.5], $y = sin(2x) + \epsilon$ with $ \epsilon \sim \mathcal{N}(0,0.2^2)$. The $y$ value corresponding to the largest $x$ value was shifted -0.4 to produce a slight outlier. 

Sub-plot A was trained via mean square error, B was regularised, C was anchored. D shows a ReLU GP.

Hyperparameters: activation = ReLU, $\sigma_{\epsilon}^2$ = 0.08, $b_1$ variance = 10, $W_1$ variance = 10, $H$ = 1000, optimiser = adam, epochs = 2,000,  learning rate = 0.003, $M$ = 10, hidden layers = 1.

\subsubsection{Classification}
Generated $\mathbf{X}$ using sklearn's `make blobs' function, n samples = 30.

Sub-plot A was trained via cross entropy, B was regularised, C was anchored. D shows inference with HMC.

Hyperparameters: activation = ReLU, $b_1$ variance = 15/2, $W_1$ variance = 15/2, $b_2$ variance = 1/50, $W_2$ variance = 1/50, $W_3$ variance = 10/50, $H$ = 50, optimiser = adam, epochs = 100,  learning rate = 0.001, $M$ = 10, hidden layers = 2.

\subsection{1-D Convergence Plots} 

Experimental details for figure \ref{fig_convergence_toy1d} are as follows.

Data as in section \ref{sec_exp_panel} was used, with $M$ = [3,5,10,20].

Hyperparameters: activation = ReLU, $\sigma_{\epsilon}^2$ = 0.001, $b_1$ variance = 20, $W_1$ variance  = 20, $H$ = 100, optimiser = adam, epochs = 4,000, learning rate = 0.005.

\subsection{KL Convergence Results} 

Experimental details for figure \ref{fig_KL_converge_plain} are as follows.

Training was done on 50\% of the data, with KL computed over the other 50\%. Results were averaged over ten runs. The `ideal' line shows the metric when posterior samples from the GP itself, rather than anchored NNs, were used.

The Boston Housing dataset was used, with 50\% of data used for training, and testing on the other 50\%. 

Hyperparameters: activation = ReLU, $\sigma_{\epsilon}^2$ = 0.1, $b_1$ variance = 2, $W_1$ variance  = 2, $H$ = [4, 16, 64, 256, 1024], $M$ = [3,5,10,20,40], optimiser = adam, no. runs = 10, epochs = 1,000, learning rate = 0.001 when $H <$ 20  else learning rate = 0.0002.

%
%
%
%

\subsection{Regression Benchmarking Experiments}

We complied with the established protocol \citep{Hernandez-Lobato2015}. Single-layer NNs of 50 nodes were used, experiments repeated 20 times with random train/test splits of 90\%/10\%. The larger Protein and Song datasets allow 100 node NNs, and were repeated five and one time respectively.

The hyperparameter tuning process and final settings for experiments in table \ref{tab_regression}, \ref{tab_rmse} \& \ref{tab_nll} are as follows.

\subsubsection{Hyperparameter Tuning}

Hyperparameter tuning was done on a single train/validation split of 80\%/20\%. We found it convenient to begin by tuning data noise variance and prior variances. We restricted the prior variance search space by enforcing, $\sigma^2_{W_1} = \sigma^2_{b_1}/D$, and $\sigma^2_{W_2} = 1/H$. We therefore had only two hyperparameters to optimise initially: $\sigma^2_{b_1}$ and $\sigma_{\epsilon}^2$. We did this with the GP model, using grid search, maximising marginal log likelihood over the training portion, and minimising NLL of the validation portion. For the larger datasets, when inference over the 80\% training portion was too slow, we reduced the training split to 2,000 data points.

Hyperparameters for priors and data noise estimates were shared between the GP and anchored ensembles. Hyperparameters requiring tuning specifically for anchored ensembles were batch size, learning rate, number of epochs and decay rate. This was done on the same 80\%/20\% split used to select data noise and prior variance. We used random search, directed by our knowledge of the optimisation process (e.g. a lower learning rate requires more epochs to converge), minimising NLL on the validation portion.

We did not retune hyperparameters from scratch for the double layer NN (5x 50-50 NNs). We used settings as for the single-layer NNs (5x 50 NNs), but divided learning rate by 4, and multiplied epochs by 1.5.

For the single regularised NN with constant noise, we again used hyperparameters as for the single-layer ensemble (5x 50 NNs), tuning only the constant amount of variance to be added on the same 80\%/20\% split.

\subsubsection{Hyperparameter Settings}

Table \ref{tab_exp_hypers} provides the key hyperparameters used. The adam optimiser was used for all experiments. ReLU activations were used for all except the ERF GP (prior variance was separately tuned for this, values aren't given in the table).

\begin{table*}[hb]%
\caption{Hyperparameters used for regression benchmark results.}
\vskip 0.15in

\begin{center}
\resizebox{1.\columnwidth}{!}{
\begin{tabular}{ l rrrrrrrrr}

\Xhline{3\arrayrulewidth}

\multicolumn{1}{c}{} 
& \multicolumn{1}{r}{$N$}
& \multicolumn{1}{r}{Batch Size}
& \multicolumn{1}{c}{Learn Rate}
&  \multicolumn{1}{c}{$\tphat{\sigma^2_\epsilon}$} 
&  \multicolumn{1}{c}{$b_1$ variance} 
&  \multicolumn{1}{c}{$W_1$ variance} 
&  \multicolumn{1}{c}{No. Epochs} 
&  \multicolumn{1}{c}{Decay Rate} 
&  \multicolumn{1}{c}{Single NN var.}  \\ 

\hline \\

Boston & 506 & 64 & 0.05 & 0.06 & 10 & 0.77 & 3000 & 0.995 & 0.45 \\
Concrete & 1,030 & 64 & 0.05 & 0.05 & 40 & 5.00 & 2000 & 0.997 & 0.28 \\
Energy & 768 & 64 & 0.05 & 1e-7 & 12 & 1.50 & 2000 & 0.997 & 0.03 \\
Kin8nm & 8,192 & 256 & 0.10 & 0.02 & 40 & 5.00 & 2000 & 0.998 & 0.32 \\
Naval & 11,934 & 256 & 0.10 & 1e-7 & 200 & 12.50 & 1000 & 0.997 & 0.03 \\
Power & 9,568 & 256 & 0.20 & 0.05 & 4 & 1.00 & 1000 & 0.995 & 0.24 \\
Protein & 45,730 & 8192 & 0.10 & 0.5 & 50 & 5.56 & 3000 & 0.995 & 0.71 \\
Wine & 1,599 & 64 & 0.05 & 0.5 & 20 & 1.82 & 500 & 0.997 & 0.77 \\
Yacht & 308 & 64 & 0.05 & 1e-7 & 15 & 2.50 & 3000 & 0.997 & 0.10 \\
Song Year & 515,345 & 32768 & 0.01 & 0.7 & 2 & 0.02 & 500 & 0.996 & 0.84 \\

\Xhline{3\arrayrulewidth} 

\end{tabular}
}
\end{center}
\vskip -0.1in
\label{tab_exp_hypers}
\end{table*}

\subsection{Out-of-Distribution Classification}
\subsubsection{Fashion MNIST}
We trained a three-layer NN on eight of ten classes of Fashion MNIST. We trained on 48,000 examples, tested on 8,000.

Experiments were repeated 5 times with a different random seed for each run.

\begin{figure}[h]
\vskip 0.2in
\begin{center}
\includegraphics[width=0.49\textwidth]{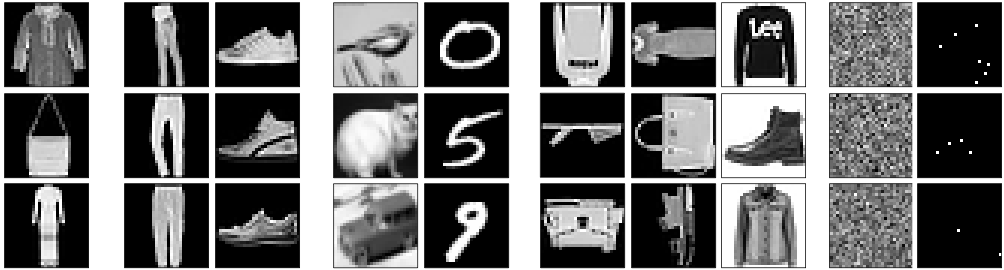}
\put (-240,68) {\small Train}
\put (-198,68) {\small Edge}
\put (-168,68) {\small CIFAR}
\put (-137,68) {\small MNIST}
\put (-95,68) {\small Distort}
\put (-35,68) {\small Noise}
\caption{Fashion MNIST OOD data examples.}
\label{fig_fashion_only_ood_egs}
\end{center}
\vskip -0.2in
\end{figure}
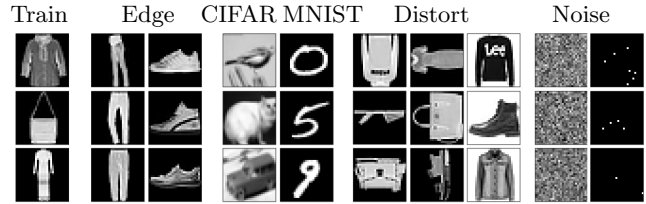

Data categories were created as suggested by their name in table \ref{tab_fashion_full}. Examples are shown in figure \ref{fig_fashion_only_ood_egs}.

\begin{itemize}
\item \textbf{Distort} comprised of rotations, vertical flips, and pixel value inversions.
\item \textbf{Noise} comprised of iid Gaussian noise, mean = $0.0$, standard deviation = $2.0$.
\item \textbf{Sparse} comprised of iid Bernoulli noise, pixles were given a value of $50.0$ with p = $0.005$, else $0.0$.
\end{itemize}

Hyperparameters: activation = ReLU, optimiser = adam, epochs = 30,  learning rate = 0.005, batch size = 256, hidden layers = 3, hidden units = 100

\subsubsection{CIFAR-10}

CIFAR-10 contains 50,000 32x32 color training images, labelled over 10 categories, and 10,000 test images. 

We removed 2 categories during training (ships, dogs) so trained over 40,000 examples.

OOD data classes are as show in the images in table \ref{tab_fashion_mini}.

\begin{itemize}

\item \textbf{Scramble} permuted each row of pixels in a given image. 

\item \textbf{Invert} took the negative of the pixel values. 

\item \textbf{Noise} sampled pixels from bernoulli distribution (p=0.005) of large magnitude (pixel value=50).
\end{itemize}

NN architecture: A convolutional NN was used, with the following structure, 64-64-maxpool-128-128-maxpool-256-256-256-maxpool-512-512-512-maxpool-flatten-2048fc-softmax.

All convolutional kernels were [3 x 3 x number of channels in previous layer]. All maxpooling kernels were [2 x 2]. The total number of parameters was 8,689,472.

Hyperparameters: activation = ReLU, optimiser = adam, learning rate = 0.001 decreasing to 0.0005 after 10 epochs and to 0.0001 after 20 epochs, batch size = 300.

In order to bring test accuracies and confidence on the training dataset roughly in line, it was necessary to train for a different number of training epochs for each method (this effectively applies early stopping to the unconstrained case). Anchored eps = 25, Regularise eps = 30, Unconstrained eps = 15.


Experiments were repeated 3 times with a different random seed for each run.

\subsubsection{IMDb}

Dataset of 25,000 movie reviews, labelled as positive or negative.

{Example}: \textit{ ``this movie is the best horror movie bar none i love how stanley just dumps the women into the lake i have been a fan of judd nelson's work for many years and he blew me away its a blend of horror and ... ''}

OOD data classes were generated as follows.

\begin{itemize}

\item\textbf{Reuters} - taken from the Reuters news dataset.

{Example}: \textit{ ``said it has started talks on the possible ... of the company with various parties that it did not identify the company said the talks began after it ... ''}

\item \textbf{Random 1} - A single integers sampled uniformly at random from $\{1 ... \text{vocabulary size} \}$ and converted to a repeated sequence of words.

{Example}: \textit{ ``member member member member member member member member member member member member ... ''}

\item \textbf{Random 2} - One integer per word sampled uniformly at random from $\{1 ... \text{vocabulary size} \}$ and converted to words.

{Example}: \textit{ ``twists mentally superb finest will dinosaur variety models stands knew refreshing member spock might mode lose leonard resemble began happily names... ''}

\item \textbf{Random 3} - As for Random 2, but now only sample from least commonly used 100 words.

{Example}: \textit{ ``computers towers bondage braveheart threatened rear triangle refuse detectives hangs bondage firmly btw token 1990s mermaid reeves landed dylan remove hum natives insightful demonic... ''}

\end{itemize}

NN architecture: used an embedding layer (outputting 20 dimensions), followed by 1D convolutional layer using 50 filters with kernel size of 3 words. Finally a hidden layer with 200 hidden nodes.

Hyperparameters: activation = ReLU, optimiser = adam, learning rate = 0.001, batch size = 64, max sentence length = 200, vocabulary size = 6000


Experiments were repeated 5 times with a different random seed for each run.

\subsection{Reinforcement Learning}
\subsubsection{Uncertainty-Aware Reinforcement Learning}

The FetchPush environment from OpenAI Gym was used with the sparse rewards setting. We modified the environment slightly. The goal was positioned at a fixed radius from the block (but at varying angle). Actions were discretised and vertical movements removed so the agent had a choice of moving 0.4 units forward/backwards/left/right. Gaussian noise was added to the actions to make the problem stochastic. Inputs were preprocessed so that relative coordinates of gripper to cube and cube to goal were provided directly to the NNs.

We used fixed target NNs which were updated every 500 episodes.

The simulation was run for 40,000 episodes, with final average rewards around $-0.4$. Two-layer NNs of 50 nodes were used. Learning rate = 0.001, batch size = 100, episodes in between training = 100, $\gamma$ = 0.98, buffer size = 100,000.


\end{appendices}
\end{document}